\documentclass[sigconf]{acmart}

\AtBeginDocument{%
  }

\setcopyright{acmlicensed}
\copyrightyear{2026}
\acmYear{2026}
\acmConference[UDM-KDD '26]{The 5th Workshop on Uncertainty Reasoning and Quantification in Decision Making (held in conjunction with ACM SIGKDD 2026)}{August 9, 2026}{Jeju, Korea}

\usepackage{mathtools}
\usepackage{algorithm}
\usepackage{algorithmic}
\usepackage{amsthm}
\usepackage{amsmath}
\usepackage{dsfont}
\usepackage{physics}
\usepackage[utf8]{inputenc}
\usepackage[T1]{fontenc}
\usepackage{hyperref}
\usepackage{url}
\usepackage{booktabs}
\usepackage{amsfonts}
\usepackage{nicefrac}
\usepackage{microtype}
\usepackage[dvipsnames]{xcolor}
\usepackage{colortbl}
\usepackage{bbm}
\usepackage[caption=false,font=normalsize,labelfont=sf,textfont=sf]{subfig}
\usepackage{textcomp}
\usepackage{pifont}
\usepackage{numprint}
\usepackage{stfloats}
\usepackage{float}
\usepackage{verbatim}
\usepackage{graphicx}
\usepackage{epsfig}
\hypersetup{
  colorlinks=true,
  allcolors=gray
}
\usepackage{array}
\usepackage{multirow}
\usepackage{multicol}
\usepackage{xspace}
\usepackage{tikz}
\usetikzlibrary{calc,positioning,arrows.meta}

\usepackage{xr-hyper}
\definecolor{quantblue}{RGB}{70,130,180}
\definecolor{calibred}{RGB}{220,20,60}
\definecolor{lightblue}{RGB}{173,216,230}
\definecolor{lightred}{RGB}{255,182,193}

\begin{document}

\title{Claim-Level Confidence Calibration for Reliable Decision Making with Large Language Models}

\author{Toghrul Abbasli}
\authornote{Corresponding author. Address: Tsinghua University, Haidian
  District, Beijing, 100084, P.~R.~China.}
\email{tgl22@mails.tsinghua.edu.cn}
\affiliation{%
  \institution{Tsinghua University}
  \city{Beijing}
  \country{China}
}

\author{Kentaroh Toyoda}
\email{kentaroh.toyoda@ieee.org}
\affiliation{%
  \institution{Vulcan Research, AIFT}
  \country{Singapore}
}
\affiliation{%
  \institution{Keio Global Research Institute (KGRI)}
  \country{Japan}
}

\author{Yuan Wang}
\email{wangyuanyjy@chinamobile.com}
\affiliation{%
  \institution{China Mobile Research Institute}
  \country{China}
}

\author{Li Chen}
\email{lichen@zgclab.edu.cn}
\affiliation{%
  \institution{Zhongguancun Laboratory}
  \country{China}
}

\renewcommand{\shortauthors}{Abbasli et al.}

\begin{abstract}
  Large Language Models (LLMs) increasingly support decision-making in
  high-stakes domains, but they often hallucinate and express confidence that
  is misaligned with factual correctness. Response-level confidence is a
  coarse signal: a single generation can mix correct and incorrect statements,
  so a single number is not actionable for users that must accept, reject, or
  verify individual pieces of information. We study \emph{claim-level
  confidence calibration} as a decision-relevant uncertainty signal: each
  response is decomposed into atomic, verifiable claims, and each claim is
  assigned a calibrated confidence using inference-time signals from
  consistency across samples and self-verification. Our framework operates in
  closed-box settings (no logits, no fine-tuning) and applies post-hoc
  calibration directly at the claim level, enabling selective intervention
  such as evidence retrieval or human review for low-confidence claims. Across
  TriviaQA and TruthfulQA we evaluate seven baselines on six recent models
  (Llama-3.1, Mistral, Qwen2.5, DeepSeek-R1, GPT-4, GPT-4o), and show that
  claim-level decomposition combined with post-hoc calibration reduces
  expected calibration error on factual questions while exposing failure modes
  on adversarial false-premise questions where decision-makers most need
  reliable uncertainty estimates.
\end{abstract}

\begin{CCSXML}
  <ccs2012>
  <concept>
  <concept_id>10010147.10010178.10010187.10010190</concept_id>
  <concept_desc>Computing methodologies~Probabilistic reasoning</concept_desc>
  <concept_significance>500</concept_significance>
  </concept>
  </ccs2012>
\end{CCSXML}

\ccsdesc[500]{Computing methodologies~Probabilistic reasoning}
\ccsdesc[300]{Computing methodologies~Information extraction}

\keywords{Large Language Models, Uncertainty Quantification, Calibration, Decision Making, Hallucination}

\received{}

\maketitle

\section{Introduction}
\label{sec:introduction}

Large Language Models (LLMs) are being deployed as components in
decision-support pipelines, where a downstream user must decide whether to
accept, reject, or further verify the model's output. In this setting,
miscalibrated confidence is not merely a statistical nuisance: a confidently
stated but incorrect claim can directly drive a wrong decision
\cite{deng2023agreementscore, hendrycks2021unsolved, zhang2024luq}. Three
broad directions have emerged to mitigate this risk:
(1)~\emph{uncertainty quantification and calibration}, which align reported
confidence with empirical accuracy \cite{guo2017calibration, ovadia2019can};
(2)~\emph{inference-time consistency checks}, which sample multiple
generations and treat agreement as a reliability signal
\cite{wang2022selfconsistency}; and (3)~\emph{self-verification and
cross-examination}, which probe a candidate response for contradictions and
unsupported content \cite{manakul2023selfcheckgpt, cohen2023lmvslm}.

We focus on (2) and (3) because they are model-agnostic, inference-time
controls applicable to closed-box deployments where access to logits or
fine-tuning is impossible. However, prior work applies these signals at the
\emph{response} level. This is too coarse for decision-making: a single
response can contain four claims, three of which are correct, and a
response-level score conflates the reliable parts with the unreliable ones. A
decision-maker who only knows that the response has $0.7$ confidence cannot
tell which sentence to trust. Decomposing responses into atomic claims makes
hallucinations measurable at the granularity at which decisions are actually
made \cite{akbar2024hallumeasure, yuan2025confawarefacrcor}.

\noindent\textbf{Contributions.} This paper studies \emph{claim-level
confidence calibration} as a decision-relevant uncertainty signal for LLM
deployments:
\begin{enumerate}
  \item We present a closed-box, inference-time pipeline that decomposes a
    response into atomic claims, attaches a per-claim confidence from
    self-verification or log-probability signals, and applies post-hoc
    calibration directly at the claim level (\S\ref{sec:methodology}).
  \item We compare against seven representative baselines spanning sequence
    likelihood, verbalized confidence, parametric calibration, and prior
    claim-level methods (HalluMeasure, LMvsLM) on TriviaQA and TruthfulQA
    across six modern models including reasoning models
    (\S\ref{sec:experiments}).
  \item We diagnose \emph{when} claim-level calibration helps decisions
    (mixed-correctness, closed-box deployments) and \emph{when} it fails
    (adversarial false premises, where models confidently reason toward
    wrong conclusions). The full appendix supplies background, additional
    metrics, reliability diagrams, and reviewer-driven ablations.
\end{enumerate}

\noindent\textbf{Relation to prior claim-level work.}
Most closely related is Yuan et al.~\cite{yuan2025confawarefacrcor}, which
proposes fact-level calibration and \emph{correction} for long-form
generation. Our objective is different: we calibrate per-claim confidence to
align with empirical correctness (ECE, smECE, Brier) without modifying
responses, apply post-hoc TS/PS at the claim level, and conduct a systematic
multi-model, multi-baseline evaluation including adversarial false-premise
questions. We also differ from HalluMeasure
\cite{akbar2024hallumeasure} (no calibration step, ternary NLI labels rather
than continuous confidence) and from LMvsLM \cite{cohen2023lmvslm} (response
level, no claim aggregation). A detailed comparison and an extended related
work appear in Appendix~\ref{appendix:related_work}.\footnote{All
appendices referenced throughout the paper are provided in the
accompanying supplementary material PDF.}

\section{Claim-Level Confidence at Inference Time}
\label{sec:methodology}

\noindent\textbf{Problem.} Given a prompt $x$ and an LLM response $r$, we
seek (i) a decomposition into atomic, verifiable claims
$\mathcal{E}(r)=\{c_1,\ldots,c_n\}$ and (ii) a confidence score
$\mathrm{conf}(c_i)\in[0,1]$ for each claim such that confidence aligns with
the empirical probability that $c_i$ is supported by evidence. Aggregated
appropriately, claim-level confidence supports concrete decision actions:
accept the response, reject it, retrieve evidence for low-confidence claims,
or escalate to a human. Background on uncertainty types, calibration
fundamentals (TS, PS, smECE, etc.), and full definitions are deferred to
Appendix~\ref{appendix:preliminaries}.

\begin{figure*}[t]
  \centering
  \includegraphics[width=\textwidth]{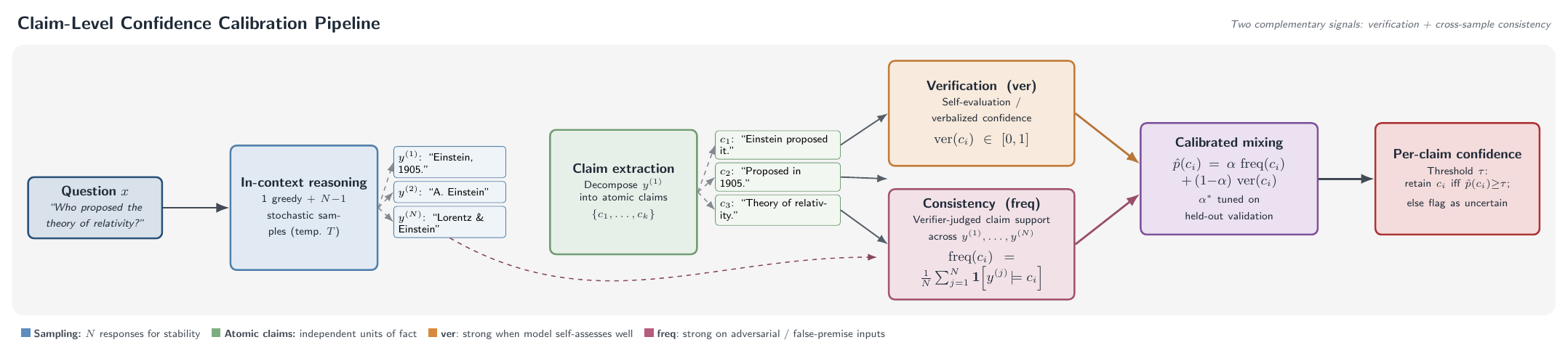}
  \caption{Proposed pipeline for claim-level confidence calibration. Given a question $x$, we (i) sample $N$ candidate responses via greedy + stochastic decoding, (ii) decompose the primary response into atomic claims $\{c_1,\dots,c_k\}$, and (iii) score each claim with two complementary signals---verification $\operatorname{ver}(c_i)\in[0,1]$ from self-evaluation and consistency $\operatorname{freq}(c_i)=\tfrac{1}{N}\sum_{j}\mathbf{1}[y^{(j)}\!\models c_i]$ from cross-sample support. The two signals are combined via a convex mixture $\hat{p}(c_i)=\alpha\,\operatorname{freq}(c_i)+(1{-}\alpha)\,\operatorname{ver}(c_i)$ with $\alpha^{*}$ tuned on held-out validation, and thresholded at $\tau$ to retain or flag each claim.}
  \Description{A horizontal pipeline diagram with six color-coded stages flowing left to right: (1) a blue Question node showing an example query; (2) a steel-blue In-context reasoning node producing one greedy and N-1 stochastic samples, with three example sampled responses shown to the right; (3) a green Claim extraction node decomposing the primary response into atomic claims, with three example claims shown to the right; (4) two parallel branches---an orange Verification (ver) branch producing per-claim probabilities, and a mauve Consistency (freq) branch counting verifier-confirmed support across the N samples; (5) a purple Calibrated mixing node combining the two signals as a convex combination with weight alpha; (6) a crimson Per-claim confidence output node applying a retention threshold tau. A bottom legend explains each color's role and notes that verification is strong when the model self-assesses well, while consistency is strong on adversarial or false-premise inputs.}
  \label{fig:rq2_pipeline}
\end{figure*}

\noindent\textbf{Pipeline (Fig.~\ref{fig:rq2_pipeline}).} Our inference-time
pipeline has five stages: \emph{(1) In-context reasoning}---greedy decoding
for the primary response and optional $N$-sample stochastic decoding;
\emph{(2) Claim extraction}---an LLM-based extractor (GPT-4o-mini) emits
self-contained, atomic claims of $\le15$ words with pronouns resolved;
\emph{(3) Claim verification}---each claim is scored by an auxiliary
verifier (LLM, RAG, or web search); \emph{(4) Per-claim confidence}---we
convert verifier outputs into a numeric score in $[0,1]$;
\emph{(5) Calibration and aggregation}---we apply post-hoc temperature or
Platt scaling \emph{at the claim level}, and aggregate to a response-level
score via mean (or min) when needed.

\noindent\textbf{Per-claim confidence signal.} For closed-box models
(GPT-4, GPT-4o, DeepSeek-R1) we obtain a verbalized confidence on a
14-level decisiveness scale \cite{tian2023justaskforcalibration} mapped to
$[0,1]$. For open-box models (Llama-3.1, Mistral, Qwen2.5) we use the
average token log-probability of the claim,
$\mathrm{ver}(c_i)=\exp\bigl(\tfrac{1}{|c_i|}\sum_{t}\log p(c_{i,t})\bigr)$.
The general framework supports a weighted combination with a cross-sample
consistency signal,
\begin{equation}
\mathrm{conf}(c_i) \;=\; \alpha\,\mathrm{freq}(c_i) + (1-\alpha)\,\mathrm{ver}(c_i),
\label{eq:conf}
\end{equation}
where $\mathrm{freq}(c_i)=\tfrac{1}{N}\sum_{j=1}^{N}\mathds{1}\!\left[c_i\in\mathcal{E}(r_j)\right]$
is the frequency of $c_i$ across $N$ sampled responses, and $\alpha\in[0,1]$
is tuned per model on a held-out subset of the training split. The main
experiments report results at the per-model best $\alpha$ (defaulting to
$\alpha{=}0$, the verification-only instantiation, when no held-out
tuning was available); sensitivity analysis over $\alpha$ is in
Appendix~\ref{appendix:alpha_sensitivity}.

\noindent\textbf{Claim-level post-hoc calibration.} We apply temperature
scaling (TS) and Platt scaling (PS) to the per-claim scores rather than to
response-level likelihoods. This is what makes the signal usable for
selective intervention: a calibrated $0.85$ on a single claim concretely
means \emph{accept this claim with $\sim$15\% expected error}, whereas a
response-level $0.85$ averages over heterogeneous sentences. Calibration
parameters are fit on a held-out subset of the training split.

\noindent\textbf{Why claim-level is closed-box-friendly.}
Unlike methods requiring fine-tuning or token-level probabilities (e.g.,
\cite{fadeeva2024factcheckingoutpllm}), the verbalized variant only requires
text outputs. This makes it directly applicable to API-only models (GPT-4,
GPT-4o) which are common decision-support deployments.

\noindent\textbf{Baselines.} We compare against (i) length-normalized
sequence likelihood with and without CoT
\cite{kuhn2023semanticuncertainty}; (ii) verbalized confidence
\cite{tian2023justaskforcalibration}; (iii) post-hoc TS/PS on sequence
likelihoods \cite{guo2017calibration, platt1999prbbplattscaling};
(iv) HalluMeasure \cite{akbar2024hallumeasure}, a claim-level
NLI-based detector; and (v) LMvsLM \cite{cohen2023lmvslm}, an iterative
cross-examination protocol. Implementation details are in
Appendix~\ref{appendix:additional_details}.

\section{Experiments}
\label{sec:experiments}

\noindent\textbf{Setup.} We evaluate on
TriviaQA \cite{joshi2017triviaqa} (factual trivia, an
easy-difficulty regime where most claims are well-formed) and
TruthfulQA \cite{lin2021truthfulqa} (an adversarial regime built around
common misconceptions and false premises). We use Llama-3.1-8B-Instruct
\cite{dubey2024llama3.1}, Mistral-7B-Instruct-v0.3 \cite{jiang2023mistral},
Qwen2.5-7B-Instruct \cite{yang2024qwen2.5},
DeepSeek-R1 \cite{guo2025deepseekrone},
GPT-4 \cite{achiam2023gpt4}, and GPT-4o \cite{hurst2024gpt4o}. We report
ECE, smECE, Brier score, AUROC, and BLEU; dataset sizes are in
Appendix~\ref{appendix:exp_data}.

    \begin{table*}[!ht]
        \centering
        \caption{Evaluation results for Llama-3.1-8B-Instruct, Mistral-7B-Instruct-v0.3, Qwen2.5-7B-Instruct, DeepSeek-R1, GPT-4, and GPT-4o on TriviaQA and TruthfulQA. ``Ours'' rows report the test-set values at the per-model best $\alpha$ tuned on a held-out split (or $\alpha{=}0$ when no held-out tuning was available).}
        \Description{A comprehensive results table showing calibration metrics (ECE, smECE, BS, AUROC, AURC, BLEU) for six language models across two datasets (TriviaQA and TruthfulQA). Rows are grouped by model, with baselines including Few-shot, Few-shot CoT, Verbalized CoT, PS, TS, LMvsLM, HalluMeasure, and the proposed method. Bold values indicate best performance per model and dataset. The proposed method is highlighted in yellow.}
        \label{tab:calibration_results}
    \scalebox{0.72}{%
    \begin{tabular}{
        >{\arraybackslash}p{0.13\textwidth}
        >{\centering\arraybackslash}p{0.04\textwidth}
        >{\centering\arraybackslash}p{0.05\textwidth}
        >{\centering\arraybackslash}p{0.03\textwidth}
        >{\centering\arraybackslash}p{0.06\textwidth}
        >{\centering\arraybackslash}p{0.04\textwidth}
        >{\centering\arraybackslash}p{0.05\textwidth}
        >{\centering\arraybackslash}p{0.03\textwidth}
        >{\centering\arraybackslash}p{0.04\textwidth}
        >{\centering\arraybackslash}p{0.05\textwidth}
        >{\centering\arraybackslash}p{0.03\textwidth}
        >{\centering\arraybackslash}p{0.06\textwidth}
        >{\centering\arraybackslash}p{0.04\textwidth}
        >{\centering\arraybackslash}p{0.05\textwidth}
        >{\centering\arraybackslash}p{0.03\textwidth}
    }
    \toprule
    \textbf{Models \& Baselines} &
    \multicolumn{7}{c}{\textbf{TriviaQA}} & \multicolumn{7}{c}{\textbf{TruthfulQA}} \\
    \cmidrule(lr){2-8} \cmidrule(lr){9-15}
    & ↓ECE & ↓smECE & ↓BS & ↑AUROC & ↓AURC & BLEU & $\alpha$ & ↓ECE & ↓smECE & ↓BS & ↑AUROC & ↓AURC & BLEU & $\alpha$ \\
    \midrule
    \rowcolor[HTML]{EFEFEF} \multicolumn{15}{l}{\textbf{Llama-3.1-8B-Instruct}} \\
    % \midrule

    Few-shot            & 0.046 & 0.040 & 0.286 & 0.532 & 0.523 & 0.041 & -- & 0.386 & 0.333 & 0.165 & 0.794 & \bf 0.001 & 0.190 & -- \\
    Few-shot CoT        & 0.087 & 0.065 & 0.273 & 0.539 & 0.436 & 0.033 & -- & 0.346 & 0.309 & 0.136 & 0.569 & 0.006 & \bf 0.191 & -- \\
    Verbal. CoT         & 0.113 & 0.064 & 0.293 & 0.564 & 0.413 & 0.033 & -- & 0.499 & 0.348 & 0.349 & 0.161 & 0.005 & 0.191 & -- \\
    PS                  & \bf 0.013 & \bf 0.013 & \bf 0.249 & 0.495 & 0.523 & 0.041 & -- & 0.363 & 0.319 & 0.134 & 0.794 & \bf 0.001 & 0.190 & -- \\
    TS                  & 0.045 & 0.039 & 0.287 & 0.505 & 0.523 & 0.041 & -- & \bf 0.154 & \bf 0.154 & \bf 0.032 & \bf 0.794 & \bf 0.001 & 0.190 & -- \\
    LMvsLM              & 0.288 & 0.145 & 0.366 & 0.592 & 0.477 & 0.040 & -- & 0.506 & 0.320 & 0.425 & 0.414 & 0.002 & 0.186 & -- \\
    HalluMeasure        & 0.242 & 0.127 & 0.346 & \bf 0.638 & \bf 0.400 & \bf 0.041 & -- & 0.989 & 0.394 & 0.986 & 0.513 & 0.007 & 0.186 & -- \\
    \rowcolor{yellow!50}
    \bf Ours            & 0.200 & 0.139 & 0.357 & 0.549 & 0.466 & 0.041 & 0.8 & 0.506 & 0.374 & 0.383 & 0.181 & 0.004 & 0.186 & 1.0 \\
    \midrule

    \rowcolor[HTML]{EFEFEF} \multicolumn{15}{l}{\textbf{Mistral-7B-Instruct-v0.3}} \\
    % \midrule

    Few-shot            & 0.044 & 0.043 & 0.265 & 0.621 & 0.367 & 0.228 & -- & 0.241 & 0.233 & 0.075 & 0.554 & --    & 0.448 & -- \\
    Few-shot CoT        & 0.025 & 0.034 & 0.283 & \bf 0.702 & 0.306 & \bf 0.310 & -- & 0.219 & 0.214 & 0.062 & 0.261 & --    & 0.386 & -- \\
    Verbal. CoT         & 0.042 & 0.028 & 0.268 & 0.547 & --    & 0.310 & -- & 0.467 & 0.369 & 0.251 & 0.535 & --    & 0.386 & -- \\
    PS                  & \bf 0.010 & \bf 0.010 & \bf 0.250 & 0.621 & --    & 0.228 & -- & 0.176 & 0.175 & 0.034 & 0.554 & --    & 0.448 & -- \\
    TS                  & 0.042 & 0.039 & 0.262 & 0.621 & --    & 0.228 & -- & \bf 0.057 & \bf 0.058 & \bf 0.007 & \bf 0.554 & --    & \bf 0.448 & -- \\
    LMvsLM              & 0.270 & 0.141 & 0.313 & 0.668 & --    & 0.228 & -- & 0.997 & 0.396 & 0.997 & 0.500 & --    & 0.448 & -- \\
    HalluMeasure        & 0.107 & 0.056 & 0.352 & 0.646 & --    & 0.228 & -- & 0.475 & 0.274 & 0.462 & 0.394 & --    & 0.448 & -- \\
    \rowcolor{yellow!50}
    \bf Ours            & 0.123 & 0.121 & 0.358 & 0.619 & 0.389 & 0.228 & 0.3  & 0.376 & 0.289 & 0.270 & 0.278 & \bf 0.003 & 0.448 & 1.0 \\
    \midrule

    \rowcolor[HTML]{EFEFEF} \multicolumn{15}{l}{\textbf{Qwen2.5-7B-Instruct}}\\
    % \midrule

    Few-shot            & 0.310 & 0.286 & 0.508 & 0.547 & --    & \bf 0.034 & -- & 0.187 & 0.185 & 0.044 & 0.543 & --    & 0.187 & -- \\
    Few-shot CoT        & 0.413 & 0.350 & 0.483 & 0.631 & --    & 0.012 & -- & 0.180 & 0.180 & 0.040 & \bf 0.867 & --    & 0.184 & -- \\
    Verbal. CoT         & 0.164 & 0.162 & 0.283 & \bf 0.658 & --    & 0.012 & -- & 0.489 & 0.392 & 0.246 & 0.516 & --    & 0.185 & -- \\
    PS                  & 0.142 & 0.141 & 0.208 & 0.417 & --    & 0.012 & -- & 0.363 & 0.320 & 0.135 & 0.543 & --    & 0.187 & -- \\
    TS                  & \bf 0.044 & \bf 0.037 & \bf 0.205 & 0.56 & --    & 0.012 & -- & \bf 0.066 & \bf 0.067 & \bf 0.008 & 0.543 & --    & \bf 0.187 & -- \\
    LMvsLM              & 0.242 & 0.122 & 0.242 & 0.500 & --    & 0.034 & -- & 0.267 & 0.214 & 0.199 & 0.296 & --    & 0.186 & -- \\
    HalluMeasure        & 0.242 & 0.121 & 0.242 & 0.500 & --    & 0.034 & -- & 0.983 & 0.403 & 0.973 & 0.530 & --    & 0.186 & -- \\
    \rowcolor{yellow!50}
    \bf Ours            & 0.126 & 0.123 & 0.198 & 0.621 & \bf 0.668 & 0.034 & 0.7 & 0.419 & 0.334 & 0.288 & 0.151 & \bf 0.005 & 0.186 & 1.0 \\
    \midrule

    \rowcolor[HTML]{EFEFEF} \textbf{GPT-4} & & & & & & & & & & & & & & \\

    Few-shot            & \bf 0.101 & \bf 0.075 & \bf 0.151 & 0.789 & --    & 0.721 & -- & 0.325 & 0.232 & 0.222 & 0.833 & --    & 0.347 & -- \\
    Few-shot CoT        & 0.217 & 0.210 & 0.297 & 0.572 & --    & 0.051 & -- & 0.635 & 0.349 & 0.664 & 0.367 & --    & 0.181 & -- \\
    Verbal. CoT         & 0.144 & 0.079 & 0.279 & 0.552 & --    & 0.051 & -- & 0.269 & 0.255 & \bf 0.088 & 0.539 & --    & 0.181 & -- \\
    PS                  & 0.232 & 0.225 & 0.206 & 0.789 & --    & 0.721 & -- & 0.352 & 0.313 & 0.136 & 0.914 & --    & 0.347 & -- \\
    TS                  & 0.105 & 0.077 & 0.153 & \bf 0.789 & --    & \bf 0.721 & -- & \bf 0.230 & \bf 0.145 & 0.185 & \bf 0.916 & --    & \bf 0.347 & -- \\
    \midrule

    \rowcolor[HTML]{EFEFEF} \textbf{GPT-4o} & & & & & & & & & & & & & & \\

    Few-shot            & 0.090 & 0.079 & \bf 0.163 & 0.765 & --    & 0.653 & -- & 0.277 & 0.259 & 0.113 & 0.408 & --    & 0.301 & -- \\
    Few-shot CoT        & 0.027 & 0.034 & 0.263 & 0.613 & --    & 0.042 & -- & 0.488 & 0.345 & 0.314 & 0.285 & --    & 0.147 & -- \\
    Verbal. CoT         & 0.310 & 0.157 & 0.331 & 0.543 & --    & 0.042 & -- & 0.159 & 0.158 & 0.056 & 0.246 & --    & 0.147 & -- \\
    PS                  & 0.232 & 0.225 & 0.216 & 0.765 & --    & 0.653 & -- & 0.366 & 0.321 & 0.137 & 0.408 & --    & 0.301 & -- \\
    TS                  & 0.096 & 0.085 & 0.165 & \bf 0.765 & --    & \bf 0.653 & -- & 0.118 & 0.116 & \bf 0.027 & 0.408 & --    & \bf 0.301 & -- \\
    LMvsLM              & 0.171 & 0.087 & 0.194 & 0.747 & --    & 0.653 & -- & \bf 0.112 & \bf 0.097 & 0.079 & 0.411 & --    & 0.301 & -- \\
    HalluMeasure        & 0.072 & 0.037 & 0.194 & 0.747 & --    & 0.653 & -- & 0.997 & 0.395 & 0.997 & \bf 0.500 & --    & 0.301 & -- \\
    \rowcolor{yellow!50}
    \bf Ours            & \bf 0.006 & \bf 0.007 & 0.208 & 0.623 & \bf 0.154 & 0.653 & 1.0 & 0.412 & 0.335 & 0.330 & 0.370 & 0.002 & 0.301 & 0.0 \\
    \midrule

    \rowcolor[HTML]{EFEFEF} \multicolumn{15}{l}{\textbf{DeepSeek-R1}} \\

    Verbal.             & 0.067 & 0.061 & 0.162 & \bf 0.594 & --    & \bf 0.440 & -- & 0.325 & 0.295 & 0.117 & 0.381 & --    & 0.225 & -- \\
    Verbal. CoT         & 0.047 & 0.038 & 0.232 & 0.527 & --    & 0.008 & -- & 0.314 & 0.288 & 0.108 & 0.405 & --    & 0.072 & -- \\
    LMvsLM              & 0.146 & 0.076 & 0.175 & 0.570 & --    & 0.440 & -- & \bf 0.066 & \bf 0.057 & 0.048 & \bf 0.924 & --    & \bf 0.225 & -- \\
    HalluMeasure        & \bf 0.001 & \bf 0.001 & 0.307 & 0.644 & --    & 0.440 & -- & 0.584 & 0.379 & 0.480 & 0.452 & --    & 0.225 & -- \\
    \rowcolor{yellow!50}
    \bf Ours            & 0.087 & 0.086 & \bf 0.158 & 0.618 & 0.128 & 0.440 & 0.0 & \bf 0.066 & 0.064 & \bf 0.029 & 0.333 & \bf 0.001 & 0.225 & 1.0 \\

    \bottomrule
    \end{tabular}%
    }% End scalebox
    \end{table*}

\noindent\textbf{What works for decision making
(Tab.~\ref{tab:calibration_results}).}
On TriviaQA, parametric calibration (PS/TS) and our claim-level method
yield substantially lower ECE than uncalibrated few-shot baselines for
every model. For Llama and Mistral, PS reduces ECE to
$\le0.013$. For Qwen and Mistral on TruthfulQA, TS cuts ECE by
$3$--$4{\times}$ over the few-shot baseline. On Mistral/TriviaQA, our
method attains an ECE of $0.123$ at $\alpha{=}0.3$, outperforming LMvsLM
($0.270$) and approaching HalluMeasure ($0.107$); notably, this is the
only setting where the best mixture weights self-verification more
heavily than cross-sample consistency ($\alpha{<}0.5$), indicating that
the relative value of the two signals is model-dependent. These
reductions translate
directly into a more usable accept/reject signal for downstream
decision-makers: a calibrated $0.9$ now reliably means $\sim$10\%
expected error.

\noindent\textbf{Selective prediction vs.\ calibration (AURC).}
For Llama-3.1, AURC reveals a complementary picture to ECE. Because PS and
TS only rescale probabilities, they preserve the few-shot ranking and
thus share its AURC ($0.523$ on TriviaQA; $0.001$ on TruthfulQA)---calibration
does not improve selective prediction by itself. Conversely, HalluMeasure
attains the lowest TriviaQA AURC ($0.400$) despite its high ECE
($0.242$), and our method sits mid-pack ($0.466$). On TruthfulQA, every
method achieves a very low AURC ($0.001$--$0.007$), indicating that the
ranking task is easy once the model's systematic errors concentrate the
low-confidence mass. The takeaway for decision-makers is that
\emph{calibration} and \emph{selective-prediction ranking} are distinct
objectives: a method can be well-calibrated yet rank claims poorly, or
vice versa.

\noindent\textbf{Where claim-level decomposition pays off.}
HalluMeasure achieves the lowest ECE on DeepSeek-R1 / TriviaQA
($0.001$) and our method outperforms the temperature-scaling baseline on
GPT-4o / TriviaQA by ~15x. The benefit of claim-level scoring is largest in
\emph{mixed-correctness} regimes (long, multi-claim factual responses)
where a response-level number averages over heterogeneous content. This
is exactly the regime where a decision-maker needs to know \emph{which}
sentence to trust, not just an overall score.

\noindent\textbf{Where it fails: adversarial false premises.}
On TruthfulQA, claim-level methods consistently underperform on models
that confidently state false-premise-induced claims. For GPT-4o, our
method's ECE rises to $0.412$ versus $0.118$ for TS---a model-level
\emph{consensus failure}: GPT-4o produces fluent, high-probability
claims that are systematically wrong, and our pipeline faithfully
reflects the model's own (mistaken) confidence rather than correcting
it. LMvsLM mitigates this on GPT-4o ($0.112$) and DeepSeek-R1 ($0.066$)
because cross-examination actively probes for contradictions. The
implication for decision-makers is that closed-box, confidence-only
signals are not sufficient under adversarial inputs: an external probe
or evidence source is required.

\noindent\textbf{Reasoning models.}
DeepSeek-R1 emits long CoT traces that, when claim-extracted in full,
produce many intermediate reasoning claims with weakly calibrated per-step
confidence. Mean aggregation dilutes the final-answer signal; restricting
extraction to the final answer segment is a practical fix that we leave
as a deployment recommendation (see
Appendix~\ref{appendix:reasoning_diagnosis}). Across both
GPT-4o and DeepSeek-R1, verbalized confidence is better calibrated than
on the open-source models, suggesting reasoning-focused training improves
the model's ability to express uncertainty in natural language.

\begin{figure}[!htbp]
    \centering
    \subfloat[Few-shot CoT]{
        \includegraphics[width=0.22\columnwidth]{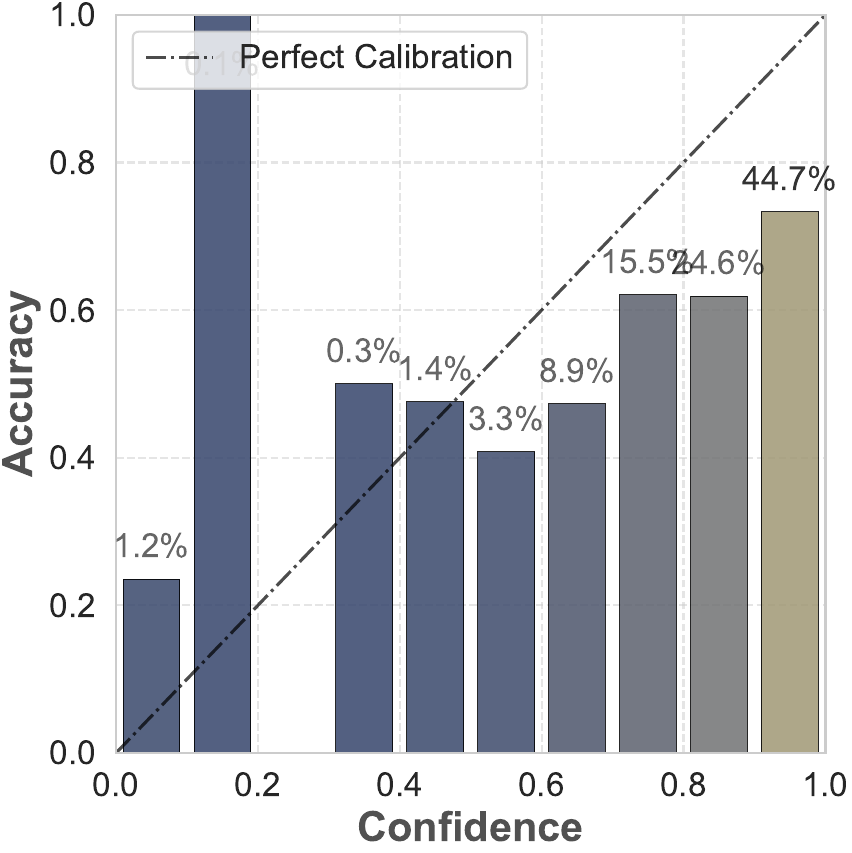}
        \label{fig:ws_cot_gpt4o_trivia}
    }
    \hfill
    \subfloat[Temp.\ Scaling]{
        \includegraphics[width=0.22\columnwidth]{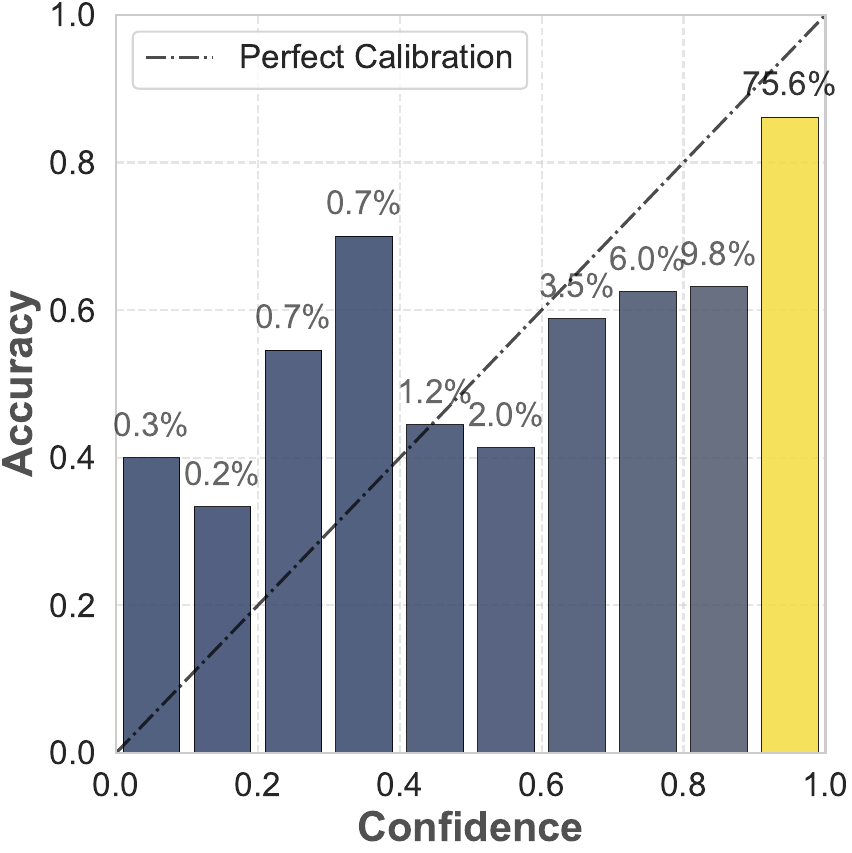}
        \label{fig:ws_ts_gpt4o_trivia}
    }
    \hfill
    \subfloat[HalluMeasure]{
        \includegraphics[width=0.22\columnwidth]{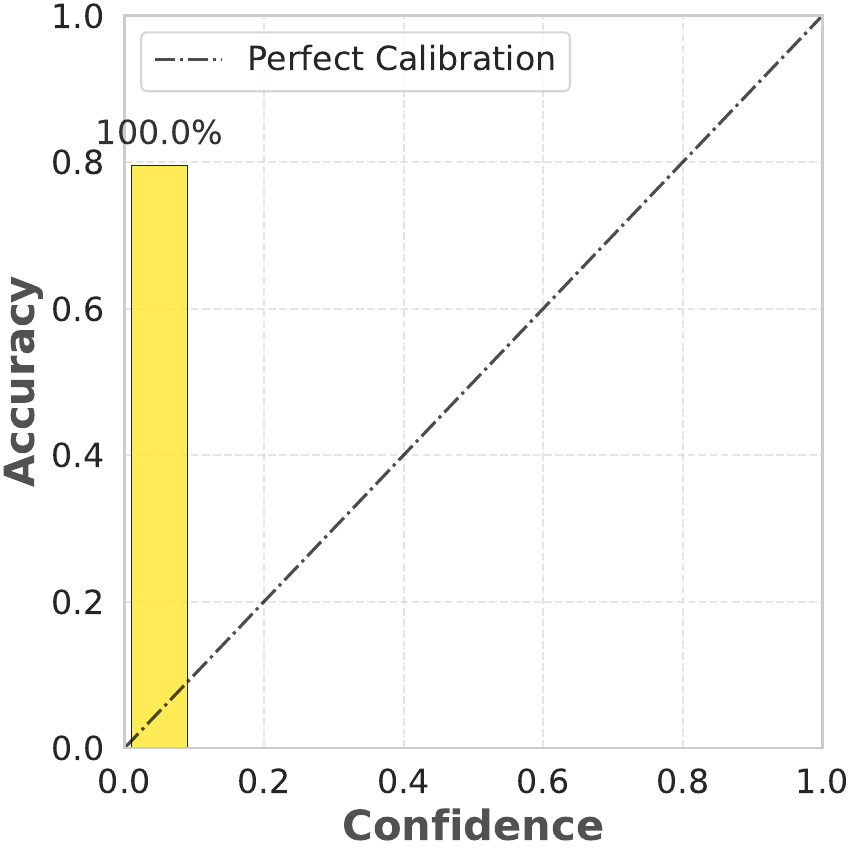}
        \label{fig:ws_hm_gpt4o_trivia}
    }
    \hfill
    \subfloat[Ours]{
        \includegraphics[width=0.22\columnwidth]{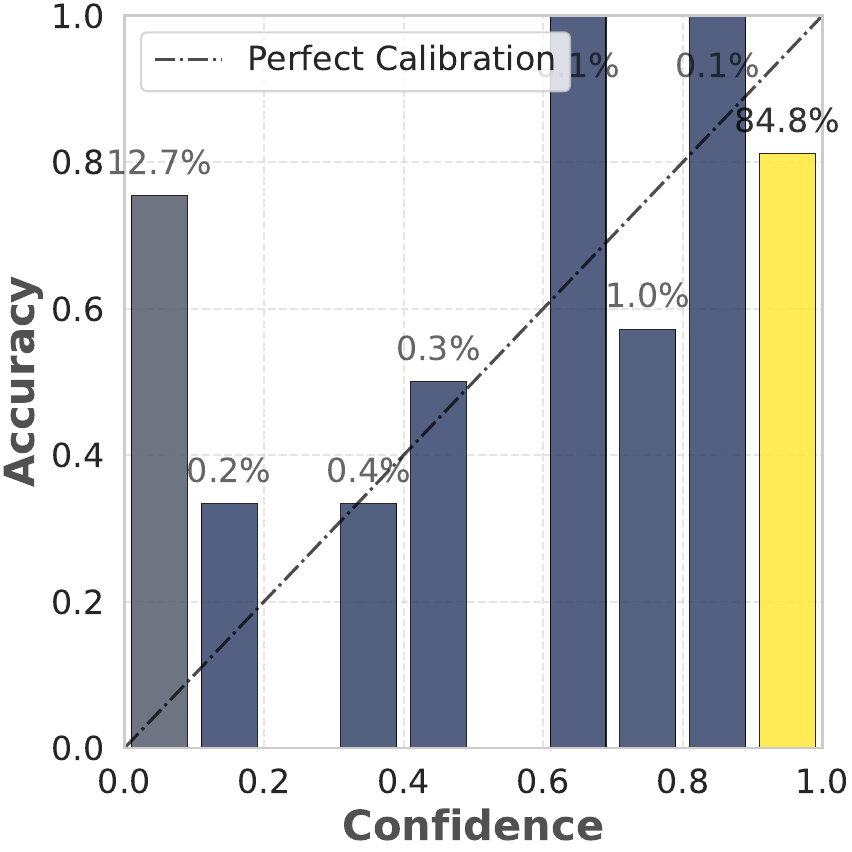}
        \label{fig:ws_ours_gpt4o_trivia}
    }
    \caption{Reliability diagrams (10 bins) for GPT-4o on TriviaQA. Bar
      colors and percentages indicate the proportion of points per bin.
      Additional models, datasets, and baselines are in
      Appendix~\ref{appendix:additional_reldiag}.}
    \label{fig:ws_gpt4o_trivia_qa_comparison}
\end{figure}

\noindent\textbf{Reliability diagrams.}
Fig.~\ref{fig:ws_gpt4o_trivia_qa_comparison} shows a representative slice for
GPT-4o on TriviaQA. TS and our method yield bars closest to the diagonal,
while HalluMeasure underweights uncertainty, reflecting its discrete NLI
signal. Diagrams for all six models on both datasets are in
Appendix~\ref{appendix:additional_reldiag}.

\noindent\textbf{Decision-relevant takeaways.}
(i)~For mixed-correctness, easy-difficulty inputs in closed-box settings,
claim-level decomposition with post-hoc calibration provides actionable
per-claim confidence and matches or beats response-level baselines.
(ii)~For adversarial inputs (false premises, misconceptions), an active
verification step (e.g., LMvsLM-style cross-examination or external
retrieval) is necessary; confidence alone is unreliable. (iii)~Reasoning
models produce better verbalized confidence but require restricting claim
extraction to the final answer for deployment. Detailed cost analysis,
$\alpha$ sensitivity, and an ablation isolating decomposition from
verifier strength are in Appendix~\ref{appendix:cost} and
Appendix~\ref{appendix:alpha_sensitivity}.

\section{Conclusion}
\label{sec:conclusion}

We studied claim-level confidence calibration as a decision-relevant
uncertainty signal for LLM deployments. Decomposing responses into atomic
claims and applying post-hoc calibration at the claim level yields
actionable per-claim confidence in closed-box settings, and reduces
expected calibration error on factual questions. The main failure mode is
adversarial false-premise inputs, where models confidently produce
incorrect claims and confidence-only signals are insufficient---a
concrete decision-making implication is that production pipelines should
combine claim-level confidence with an active verification step
(cross-examination or retrieval). Open directions include adaptive
sampling, hierarchical confidence over interdependent claims, and
domain-specific calibration for high-stakes decision-support
applications.

\begin{acks}
This work is supported by the Beijing Outstanding Young Scientist Program (No.JWZQ20240101008).
\end{acks}

\bibliographystyle{ACM-Reference-Format}
\bibliography{ref}

@String{Computer = "{IEEE} Computer" }

@String{Chelsea = "Chelsea" }

@article{hendrycks2021unsolved,
  title={Unsolved problems in ml safety},
  author={Hendrycks, Dan and Carlini, Nicholas and Schulman, John and Steinhardt, Jacob},
  journal={arXiv preprint arXiv:2109.13916},
  year={2021}
}

@article{liang2023mid,
  title={Encouraging divergent thinking in large language models through multi-agent debate},
  author={Liang, Tian and He, Zhiwei and Jiao, Wenxiang and Wang, Xing and Wang, Yan and Wang, Rui and Yang, Yujiu and Shi, Shuming and Tu, Zhaopeng},
  journal={arXiv preprint arXiv:2305.19118},
  year={2023}
}

@article{cohen2023lmvslm,
  title={Lm vs lm: Detecting factual errors via cross examination},
  author={Cohen, Roi and Hamri, May and Geva, Mor and Globerson, Amir},
  journal={arXiv preprint arXiv:2305.13281},
  year={2023}
}

@inproceedings{akbar2024hallumeasure,
  title={HalluMeasure: Fine-grained hallucination measurement using chain-of-thought reasoning},
  author={Akbar, Shayan Ali and Hossain, Md Mosharaf and Wood, Tess and Chin, Si-Chi and Salinas, Erica M and Alvarez, Victor and Cornejo, Erwin},
  booktitle={Proceedings of the 2024 Conference on Empirical Methods in Natural Language Processing},
  pages={15020--15037},
  year={2024}
}

@inproceedings{yuan2025confawarefacrcor,
  title={Fact-Level Calibration and Correction for Long-Form Generations},
  author={Yuan, Yige and Xu, Bingbing and Tan, Hexiang and Sun, Fei and Xiao, Teng and Li, Wei and Shen, Huawei and Cheng, Xueqi},
  booktitle={Proceedings of the 48th International ACM SIGIR Conference on Research and Development in Information Retrieval},
  pages={2807--2811},
  year={2025}
}

@article{detommaso2024multicalibrationiglb,
  title={Multicalibration for confidence scoring in LLMs},
  author={Detommaso, Gianluca and Bertran, Martin and Fogliato, Riccardo and Roth, Aaron},
  journal={arXiv preprint arXiv:2404.04689},
  year={2024}
}

@inproceedings{gopalan2022lowdegreemulticalib,
  title={Low-degree multicalibration},
  author={Gopalan, Parikshit and Kim, Michael P and Singhal, Mihir A and Zhao, Shengjia},
  booktitle={Conference on Learning Theory},
  pages={3193--3234},
  year={2022},
  organization={PMLR}
}

@inproceedings{liu2024litcab,
  title={LitCab: Lightweight Language Model Calibration over Short-and Long-form Responses},
  author={Liu, Xin and Khalifa, Muhammad and Wang, Lu},
  booktitle={The Twelfth International Conference on Learning Representations},
  year={2024}
}

@article{Yang2023Laplacelora,
  title={Bayesian low-rank adaptation for large language models},
  author={Adam X. Yang and Maxime Robeyns and Xi Wang and Laurence Aitchison},
  journal={ArXiv},
  year={2023},
  volume={abs/2308.13111},
  url={https://api.semanticscholar.org/CorpusID:261214713}
}

@inproceedings{xu2024sayself,
  title={Sayself: Teaching llms to express confidence with self-reflective rationales},
  author={Xu, Tianyang and Wu, Shujin and Diao, Shizhe and Liu, Xiaoze and Wang, Xingyao and Chen, Yangyi and Gao, Jing},
  booktitle={Proceedings of the 2024 Conference on Empirical Methods in Natural Language Processing},
  pages={5985--5998},
  year={2024}
}

@inproceedings{zhang2024rtuning,
  title={R-Tuning: Instructing Large Language Models to Say ‘I Don’t Know’},
  author={Zhang, Hanning and Diao, Shizhe and Lin, Yong and Fung, Yi and Lian, Qing and Wang, Xingyao and Chen, Yangyi and Ji, Heng and Zhang, Tong},
  booktitle={Proceedings of the 2024 Conference of the North American Chapter of the Association for Computational Linguistics: Human Language Technologies (Volume 1: Long Papers)},
  pages={7106--7132},
  year={2024}
}

@article{zhao2023slichf,
  title={Slic-hf: Sequence likelihood calibration with human feedback},
  author={Zhao, Yao and Joshi, Rishabh and Liu, Tianqi and Khalman, Misha and Saleh, Mohammad and Liu, Peter J},
  journal={arXiv preprint arXiv:2305.10425},
  year={2023}
}

@article{jiang2021qalmcalibration,
  title={How can we know when language models know? on the calibration of language models for question answering},
  author={Jiang, Zhengbao and Araki, Jun and Ding, Haibo and Neubig, Graham},
  journal={Transactions of the Association for Computational Linguistics},
  volume={9},
  pages={962--977},
  year={2021},
  publisher={MIT Press One Rogers Street, Cambridge, MA 02142-1209, USA journals-info~…}
}

@article{huang2024calibratinglongform,
  title={Calibrating Long-form Generations from Large Language Models},
  author={Huang, Yukun and Liu, Yixin and Thirukovalluru, Raghuveer and Cohan, Arman and Dhingra, Bhuwan},
  journal={arXiv preprint arXiv:2402.06544},
  year={2024}
}

@article{chen2023universalselfconsistency,
  title={Universal self-consistency for large language model generation},
  author={Chen, Xinyun and Aksitov, Renat and Alon, Uri and Ren, Jie and Xiao, Kefan and Yin, Pengcheng and Prakash, Sushant and Sutton, Charles and Wang, Xuezhi and Zhou, Denny},
  journal={arXiv preprint arXiv:2311.17311},
  year={2023}
}

@article{jiang2023cape,
  title={Calibrating language models via augmented prompt ensembles},
  author={Jiang, Mingjian and Ruan, Yangjun and Huang, Sicong and Liao, Saifei and Pitis, Silviu and Grosse, Roger Baker and Ba, Jimmy},
  year={2023}
}

@article{tian2023justaskforcalibration,
  title={Just ask for calibration: Strategies for eliciting calibrated confidence scores from language models fine-tuned with human feedback},
  author={Tian, Katherine and Mitchell, Eric and Zhou, Allan and Sharma, Archit and Rafailov, Rafael and Yao, Huaxiu and Finn, Chelsea and Manning, Christopher D},
  journal={arXiv preprint arXiv:2305.14975},
  year={2023}
}

@article{xiong2023canllmexpress,
  title={Can llms express their uncertainty? an empirical evaluation of confidence elicitation in llms},
  author={Xiong, Miao and Hu, Zhiyuan and Lu, Xinyang and Li, Yifei and Fu, Jie and He, Junxian and Hooi, Bryan},
  journal={arXiv preprint arXiv:2306.13063},
  year={2023}
}

@article{mielke2022reducingoverconfidence,
  title={Reducing conversational agents’ overconfidence through linguistic calibration},
  author={Mielke, Sabrina J and Szlam, Arthur and Dinan, Emily and Boureau, Y-Lan},
  journal={Transactions of the Association for Computational Linguistics},
  volume={10},
  pages={857--872},
  year={2022},
  publisher={MIT Press One Broadway, 12th Floor, Cambridge, Massachusetts 02142, USA~…}
}

@article{wang2022selfconsistency,
  title={Self-consistency improves chain of thought reasoning in language models},
  author={Wang, Xuezhi and Wei, Jason and Schuurmans, Dale and Le, Quoc and Chi, Ed and Narang, Sharan and Chowdhery, Aakanksha and Zhou, Denny},
  journal={arXiv preprint arXiv:2203.11171},
  year={2022}
}

@article{wang2024subjectiveuqcal,
  title={On Subjective Uncertainty Quantification and Calibration in Natural Language Generation},
  author={Wang, Ziyu and Holmes, Chris},
  journal={arXiv preprint arXiv:2406.05213},
  year={2024}
}

@article{kull2019beyond,
  title={Beyond temperature scaling: Obtaining well-calibrated multi-class probabilities with dirichlet calibration},
  author={Kull, Meelis and Perello Nieto, Miquel and K{\"a}ngsepp, Markus and Silva Filho, Telmo and Song, Hao and Flach, Peter},
  journal={Advances in neural information processing systems},
  volume={32},
  year={2019}
}

@inproceedings{guo2017calibration,
  title={On calibration of modern neural networks},
  author={Guo, Chuan and Pleiss, Geoff and Sun, Yu and Weinberger, Kilian Q},
  booktitle={International conference on machine learning},
  pages={1321--1330},
  year={2017},
  organization={PMLR}
}

@article{platt1999prbbplattscaling,
  title={Probabilistic outputs for support vector machines and comparisons to regularized likelihood methods},
  author={Platt, John and others},
  journal={Advances in large margin classifiers},
  volume={10},
  number={3},
  pages={61--74},
  year={1999},
  publisher={Cambridge, MA}
}

@article{kumar2019verified,
  title={Verified uncertainty calibration},
  author={Kumar, Ananya and Liang, Percy S and Ma, Tengyu},
  journal={Advances in Neural Information Processing Systems},
  volume={32},
  year={2019}
}

@inproceedings{zadrozny2001histogrambinning,
  title={Obtaining calibrated probability estimates from decision trees and naive bayesian classifiers},
  author={Zadrozny, Bianca and Elkan, Charles},
  booktitle={Icml},
  volume={1},
  pages={609--616},
  year={2001}
}

@article{ulmer2024apricot,
  title={Calibrating Large Language Models Using Their Generations Only},
  author={Ulmer, Dennis and Gubri, Martin and Lee, Hwaran and Yun, Sangdoo and Oh, Seong Joon},
  journal={arXiv preprint arXiv:2403.05973},
  year={2024}
}

@article{ovadia2019can,
  title={Can you trust your model's uncertainty? evaluating predictive uncertainty under dataset shift},
  author={Ovadia, Yaniv and Fertig, Emily and Ren, Jie and Nado, Zachary and Sculley, David and Nowozin, Sebastian and Dillon, Joshua and Lakshminarayanan, Balaji and Snoek, Jasper},
  journal={Advances in neural information processing systems},
  volume={32},
  year={2019}
}

@article{deng2023agreementscore,
  title={Great models think alike: improving model reliability via inter-model latent agreement},
  author={Deng, Ailin and Xiong, Miao and Hooi, Bryan},
  journal={arXiv preprint arXiv:2305.01481},
  year={2023}
}

@article{walker2003defining,
  title={Defining uncertainty: a conceptual basis for uncertainty management in model-based decision support},
  author={Walker, Warren E and Harremo{\"e}s, Poul and Rotmans, Jan and Van Der Sluijs, Jeroen P and Van Asselt, Marjolein BA and Janssen, Peter and Krayer von Krauss, Martin P},
  journal={Integrated assessment},
  volume={4},
  number={1},
  pages={5--17},
  year={2003},
  publisher={Taylor \& Francis}
}

@article{ling2024deductiveverification,
  title={Deductive verification of chain-of-thought reasoning},
  author={Ling, Zhan and Fang, Yunhao and Li, Xuanlin and Huang, Zhiao and Lee, Mingu and Memisevic, Roland and Su, Hao},
  journal={Advances in Neural Information Processing Systems},
  volume={36},
  year={2024}
}

@article{manakul2023selfcheckgpt,
  title={Selfcheckgpt: Zero-resource black-box hallucination detection for generative large language models},
  author={Manakul, Potsawee and Liusie, Adian and Gales, Mark JF},
  journal={arXiv preprint arXiv:2303.08896},
  year={2023}
}

@article{zhang2024luq,
  title={LUQ: Long-text Uncertainty Quantification for LLMs},
  author={Zhang, Caiqi and Liu, Fangyu and Basaldella, Marco and Collier, Nigel},
  journal={arXiv preprint arXiv:2403.20279},
  year={2024}
}

@article{kirchhof2023probabilistic,
  title={Probabilistic contrastive learning recovers the correct aleatoric uncertainty of ambiguous inputs},
  author={Kirchhof, Michael and Kasneci, Enkelejda and Oh, Seong Joon},
  journal={arXiv preprint arXiv:2302.02865},
  year={2023}
}

@inproceedings{vazhentsev2023hybridue,
  title={Hybrid uncertainty quantification for selective text classification in ambiguous tasks},
  author={Vazhentsev, Artem and Kuzmin, Gleb and Tsvigun, Akim and Panchenko, Alexander and Panov, Maxim and Burtsev, Mikhail and Shelmanov, Artem},
  booktitle={Proceedings of the 61st Annual Meeting of the Association for Computational Linguistics (Volume 1: Long Papers)},
  pages={11659--11681},
  year={2023}
}

@article{kendallgal2017whatuncertaintiesweneed,
  title={What uncertainties do we need in bayesian deep learning for computer vision?},
  author={Kendall, Alex and Gal, Yarin},
  journal={Advances in neural information processing systems},
  volume={30},
  year={2017}
}

@article{chua2023tacklinguncert,
  title={Tackling prediction uncertainty in machine learning for healthcare},
  author={Chua, Michelle and Kim, Doyun and Choi, Jongmun and Lee, Nahyoung G and Deshpande, Vikram and Schwab, Joseph and Lev, Michael H and Gonzalez, Ramon G and Gee, Michael S and Do, Synho},
  journal={Nature Biomedical Engineering},
  volume={7},
  number={6},
  pages={711--718},
  year={2023},
  publisher={Nature Publishing Group UK London}
}

@article{tomov2025illusionofcertainty,
  title={The Illusion of Certainty: Uncertainty quantification for LLMs fails under ambiguity},
  author={Tomov, Tim and Fuchsgruber, Dominik and Wollschl{\"a}ger, Tom and G{\"u}nnemann, Stephan},
  journal={arXiv preprint arXiv:2511.04418},
  year={2025}
}

@article{farquhar2024detecting,
  title={Detecting hallucinations in large language models using semantic entropy},
  author={Farquhar, Sebastian and Kossen, Jannik and Kuhn, Lorenz and Gal, Yarin},
  journal={Nature},
  volume={630},
  number={8017},
  pages={625--630},
  year={2024},
  publisher={Nature Publishing Group UK London}
}

@article{kuhn2023semanticuncertainty,
  title={Semantic uncertainty: Linguistic invariances for uncertainty estimation in natural language generation},
  author={Kuhn, Lorenz and Gal, Yarin and Farquhar, Sebastian},
  journal={arXiv preprint arXiv:2302.09664},
  year={2023}
}

@article{lin2021truthfulqa,
  title={Truthfulqa: Measuring how models mimic human falsehoods},
  author={Lin, Stephanie and Hilton, Jacob and Evans, Owain},
  journal={arXiv preprint arXiv:2109.07958},
  year={2021}
}

@article{joshi2017triviaqa,
  title={Triviaqa: A large scale distantly supervised challenge dataset for reading comprehension},
  author={Joshi, Mandar and Choi, Eunsol and Weld, Daniel S and Zettlemoyer, Luke},
  journal={arXiv preprint arXiv:1705.03551},
  year={2017}
}

@article{guo2025deepseekrone,
  title={Deepseek-r1: Incentivizing reasoning capability in llms via reinforcement learning},
  author={Guo, Daya and Yang, Dejian and Zhang, Haowei and Song, Junxiao and Zhang, Ruoyu and Xu, Runxin and Zhu, Qihao and Ma, Shirong and Wang, Peiyi and Bi, Xiao and others},
  journal={arXiv preprint arXiv:2501.12948},
  year={2025}
}

@article{dubey2024llama3.1,
  title={The llama 3 herd of models},
  author={Dubey, Abhimanyu and Jauhri, Abhinav and Pandey, Abhinav and Kadian, Abhishek and Al-Dahle, Ahmad and Letman, Aiesha and Mathur, Akhil and Schelten, Alan and Yang, Amy and Fan, Angela and others},
  journal={arXiv preprint arXiv:2407.21783},
  year={2024}
}

@article{yang2024qwen2.5,
  title={Qwen2. 5 Technical Report},
  author={Yang, An and Yang, Baosong and Zhang, Beichen and Hui, Binyuan and Zheng, Bo and Yu, Bowen and Li, Chengyuan and Liu, Dayiheng and Huang, Fei and Wei, Haoran and others},
  journal={arXiv preprint arXiv:2412.15115},
  year={2024}
}

@article{jiang2023mistral,
  title={Mistral 7B},
  author={Jiang, Albert Q and Sablayrolles, Alexandre and Mensch, Arthur and Bamford, Chris and Chaplot, Devendra Singh and Casas, Diego de las and Bressand, Florian and Lengyel, Gianna and Lample, Guillaume and Saulnier, Lucile and others},
  journal={arXiv preprint arXiv:2310.06825},
  year={2023}
}

@article{hurst2024gpt4o,
  title={Gpt-4o system card},
  author={Hurst, Aaron and Lerer, Adam and Goucher, Adam P and Perelman, Adam and Ramesh, Aditya and Clark, Aidan and Ostrow, AJ and Welihinda, Akila and Hayes, Alan and Radford, Alec and others},
  journal={arXiv preprint arXiv:2410.21276},
  year={2024}
}

@article{achiam2023gpt4,
  title={Gpt-4 technical report},
  author={Achiam, Josh and Adler, Steven and Agarwal, Sandhini and Ahmad, Lama and Akkaya, Ilge and Aleman, Florencia Leoni and Almeida, Diogo and Altenschmidt, Janko and Altman, Sam and Anadkat, Shyamal and others},
  journal={arXiv preprint arXiv:2303.08774},
  year={2023}
}

@inproceedings{papineni2002bleu,
author = {Papineni, Kishore and Roukos, Salim and Ward, Todd and Zhu, Wei-Jing},
title = {BLEU: a method for automatic evaluation of machine translation},
year = {2002},
publisher = {Association for Computational Linguistics},
address = {USA},
url = {https://doi.org/10.3115/1073083.1073135},
doi = {10.3115/1073083.1073135},
booktitle = {Proceedings of the 40th Annual Meeting on Association for Computational Linguistics},
pages = {311–318},
numpages = {8},
location = {Philadelphia, Pennsylvania},
series = {ACL '02}
}

@inproceedings{fadeeva2024factcheckingoutpllm,
  title={Fact-Checking the Output of Large Language Models via Token-Level Uncertainty Quantification},
  author={Fadeeva, Ekaterina and Rubashevskii, Aleksandr and Shelmanov, Artem and Petrakov, Sergey and Li, Haonan and Mubarak, Hamdy and Tsymbalov, Evgenii and Kuzmin, Gleb and Panchenko, Alexander and Baldwin, Timothy and Nakov, Preslav and Panov, Maxim},
  booktitle={Findings of the Association for Computational Linguistics: ACL 2024},
  year={2024}
}

% \newpage
\appendix
\section{Preliminaries and Calibration Background}
\label{appendix:preliminaries}

\subsection{Key Definitions}

\noindent\textbf{Claim.} A \emph{claim} is an atomic, verifiable statement
extracted from an LLM response. Given a prompt $x$ and response $r$, a claim
extraction function
$\mathcal{E}: r\rightarrow\{c_1,\ldots,c_n\}$ produces $n$ atomic claims.
Each claim $c_i$ should be (i) \emph{atomic} (a single factual assertion);
(ii) \emph{verifiable} (assessable as supported, contradicted, or
unknown given evidence); and (iii) \emph{self-contained} (interpretable
without the full response context).

\noindent\textbf{Claim verification.} Given $c_i$ and optional evidence
$e$, a verifier $\mathcal{V}: (c_i,e)\rightarrow\{\text{verbalized
confidence}\}$ produces a verification label or numeric score. The
verifier may be an auxiliary LLM, a retrieval-augmented generation (RAG)
system, or an external knowledge source.

\noindent\textbf{Inference-time signals.} Statistics derivable during
generation that correlate with correctness without requiring access to
parameters or training. We use two types: (i)~\emph{consistency
signals}, derived by sampling $N$ responses and measuring agreement; and
(ii)~\emph{self-verification signals}, obtained by prompting the model or
an auxiliary verifier.

\noindent\textbf{Claim-level confidence.} For each claim we estimate
$\mathrm{conf}(c_i)\in[0,1]$ that should be calibrated, i.e.\
$\mathrm{conf}(c_i)$ should align with the empirical probability that
$c_i$ is supported by evidence.

\subsection{Uncertainty: Epistemic vs.\ Aleatoric}

Uncertainty is commonly split into epistemic (model) uncertainty and
aleatoric (data) uncertainty
\cite{walker2003defining, kendallgal2017whatuncertaintiesweneed,
chua2023tacklinguncert}. Epistemic uncertainty arises from limited
knowledge and is reducible with more data or a better model. Aleatoric
uncertainty arises from input noise and is irreducible
\cite{tomov2025illusionofcertainty}. Many estimation approaches use
entropy-based statistics on softmax outputs
\cite{vazhentsev2023hybridue, wang2024subjectiveuqcal}, with recent work
on probabilistic contrastive learning to better represent data
uncertainty \cite{kirchhof2023probabilistic}.

\subsection{Calibration Methods for LLMs}

Existing calibration methods for LLMs fall into three groups by access
level.

\noindent\textbf{Open-box calibration} fine-tunes the model or a small
adapter using internal logits/representations. Calibrated fine-tuning
penalizes overconfident incorrect answers \cite{jiang2021qalmcalibration};
LitCab \cite{liu2024litcab} adds a single linear layer on top of last-layer
logits with a max-margin objective; Yang et al.\
\cite{Yang2023Laplacelora} integrate Laplace approximation with LoRA;
SLiC-HF \cite{zhao2023slichf} uses preference data to calibrate token
ranking. Other work fine-tunes models to output ``unsure''
\cite{zhang2024rtuning, xu2024sayself}. Open-box methods are costly at
LLM scale and often fail to generalize across model families
\cite{tian2023justaskforcalibration}.

\noindent\textbf{Post-hoc calibration} rescales final-layer outputs.
Temperature scaling (TS) \cite{guo2017calibration} introduces a
temperature $T$ tuned on a held-out set,
\begin{equation}
p(y\mid x) = \frac{\exp(\mathrm{logit}\,f(x)/T)}{\sum_j \exp(\mathrm{logit}\,f_j(x)/T)}.
\label{eq:temperature_scale}
\end{equation}
Platt scaling (PS) \cite{platt1999prbbplattscaling} learns scale $a$ and
bias $b$,
\begin{equation}
p(y_i\mid x) = \sigma(a\,\mathrm{logit}\,f_i(x) + b).
\label{eq:platt_scale}
\end{equation}
Both can fail under distribution shift \cite{ovadia2019can}. Histogram
binning \cite{zadrozny2001histogrambinning} is a non-parametric
alternative; scale binning combines parametric and non-parametric pieces
\cite{kumar2019verified}.

\noindent\textbf{Closed-box calibration} requires no internal access.
Linguistic calibration mitigates misaligned verbal confidence
\cite{mielke2022reducingoverconfidence, xiong2023canllmexpress,
tian2023justaskforcalibration}. Prompt-augmentation strategies (paraphrase,
permutation, in-context examples) can be ensembled
\cite{jiang2023cape}. APRICOT \cite{ulmer2024apricot} trains a small
auxiliary calibrator on LLM outputs.

\subsection{Calibration Metrics}
\label{appendix:metrics}

\noindent\textbf{ECE.} Expected Calibration Error,
\begin{equation}
\mathrm{ECE} = \sum_{i=1}^k \frac{|B_i|}{N}\,\bigl|\mathrm{acc}(B_i) - \mathrm{conf}(B_i)\bigr|,
\end{equation}
with $k$ confidence bins $B_i=[(i-1)/k, i/k]$, bin-wise accuracy
$\mathrm{acc}(B_i)=\frac{1}{|B_i|}\sum_{j\in B_i}\mathds{1}(y_j=y_{\mathrm{pred},j})$
and bin-wise mean confidence $\mathrm{conf}(B_i)$. We use $k=10$.

\noindent\textbf{SmoothECE (smECE).} Replaces binning with Gaussian
kernel smoothing,
\begin{equation}
\mathrm{smECE}_{K,\sigma} = \int \frac{1}{N}\sum_i \bigl|K(t,\mathrm{conf}(y_i))(\mathrm{conf}(y_i)-\mathrm{acc}(y_i))\bigr|\,dt,
\end{equation}
with $K(t_1,t_2)=\exp(-(t_1-t_2)^2/2\sigma)/\sqrt{2\pi\sigma}$, providing
a binless approximation to the Wasserstein distance to perfect
calibration.

\noindent\textbf{Brier Score (BS).}
$\mathrm{BS} = \tfrac{1}{N}\sum_i (\mathrm{acc}(y_i,y_{\mathrm{pred}})-\mathrm{conf}(y_{\mathrm{pred}},x_i))^2$,
also known as the mean squared error
\cite{detommaso2024multicalibrationiglb}. BS measures how much data
variance the model captures \cite{gopalan2022lowdegreemulticalib}.

\noindent\textbf{AUROC.} Area under the receiver-operating curve for
misprediction detection on confidence scores
\cite{ulmer2024apricot},
$\mathrm{AUROC} = \int_0^1 \mathrm{TPR}(\mathrm{FPR}^{-1}(t))\,dt.$

\noindent\textbf{BLEU.} Bilingual evaluation understudy
\cite{papineni2002bleu} as a generation-quality reference. BLEU is
sensitive to length and is included only as a reference signal; the
calibration analysis relies on ECE/smECE/BS/AUROC.

\noindent\textbf{UCE, RCE, cECE, ACE, AURC.} Additional definitions
(uncertainty calibration error, rank calibration error, class-wise ECE,
adaptive calibration error, area under risk-coverage curve) follow the
formulations in \cite{kull2019beyond, gopalan2022lowdegreemulticalib,
ulmer2024apricot}; we omit the formulas for brevity but use them
consistently with prior work.

\section{Extended Related Work}
\label{appendix:related_work}

\noindent\textbf{Uncertainty estimation for LLMs} spans open-box,
post-hoc, and closed-box methods (cited above). Closely related
inference-time approaches sample multiple generations: self-consistency
\cite{wang2022selfconsistency}, universal self-consistency
\cite{chen2023universalselfconsistency}, semantic uncertainty
\cite{kuhn2023semanticuncertainty}, and self-evaluation for long-form
calibration \cite{huang2024calibratinglongform}. Self-verification methods
include SelfCheckGPT \cite{manakul2023selfcheckgpt}; cross-examination
methods include LMvsLM \cite{cohen2023lmvslm} and multi-agent debate
\cite{liang2023mid}. Token-level uncertainty quantification for fact
checking has been studied by Fadeeva et al.\
\cite{fadeeva2024factcheckingoutpllm}; this requires per-token
log-probabilities and is therefore inapplicable to closed-box deployments
that are our primary target.

\noindent\textbf{Claim-level evaluation.} HalluMeasure
\cite{akbar2024hallumeasure} decomposes responses into atomic claims and
classifies each via NLI labels; entailment-based verification uses NLI
or LLMs as verifiers \cite{ling2024deductiveverification}.

\noindent\textbf{Closest relation: Yuan et al.\
\cite{yuan2025confawarefacrcor}.}
Their work proposes fact-level confidence calibration and \emph{correction}
for long-form generation. Distinctions:
\textbf{(i)~Calibration vs.\ correction.} Yuan et al.\ \emph{rewrite}
incorrect facts; we estimate whether expressed per-claim confidence aligns
with empirical correctness probability (ECE, smECE) without modifying
responses.
\textbf{(ii)~Post-hoc calibration.} Yuan et al.\ do not apply TS/PS at the
claim level; we treat per-claim confidence as a quantity to be calibrated.
\textbf{(iii)~Evaluation metrics.} They use precision/recall/correction
accuracy; we use ECE/smECE/BS/AUROC---confidence-accuracy alignment, not
rewriting quality.
\textbf{(iv)~Breadth.} We evaluate 6 model families, 2 datasets (factual
vs.\ false-premise), and 7 baselines.
\textbf{(v)~Failure-mode analysis.} We diagnose adversarial false
premises and extended CoT traces, perspectives absent in Yuan et al.

\subsection{Method Comparison Table}
\begin{table}[h]
\centering
\caption{Comparison of uncertainty estimation methods for LLMs.}
\label{tab:method_comparison}
\resizebox{\columnwidth}{!}{%
\begin{tabular}{lccccc}
\toprule
\textbf{Method} & \textbf{Open-box?} & \textbf{Claim-level?} & \textbf{Consistency?} & \textbf{Calibration?} \\
\midrule
Fine-tuning \cite{jiang2021qalmcalibration} & Yes & No & No & Yes \\
TS/PS \cite{guo2017calibration} & Logits & No & No & Yes \\
Verbalized \cite{tian2023justaskforcalibration} & No & No & No & Limited \\
Self-consistency \cite{wang2022selfconsistency} & No & No & Yes & No \\
SelfCheckGPT \cite{manakul2023selfcheckgpt} & No & No & Yes & No \\
LMvsLM \cite{cohen2023lmvslm} & No & Yes & No & No \\
HalluMeasure \cite{akbar2024hallumeasure} & No & Yes & No & No \\
\midrule
\textbf{Ours} & \textbf{No} & \textbf{Yes} & \textbf{Yes} & \textbf{Yes} \\
\bottomrule
\end{tabular}%
}% End resizebox
\end{table}

\section{Implementation Details}
\label{appendix:additional_details}

\subsection{Datasets}
\label{appendix:exp_data}

\noindent\textbf{TriviaQA} \cite{joshi2017triviaqa} contains $>$650K
question--answer--evidence triples, plus 95K author-written QA pairs.
Following \cite{ulmer2024apricot} we sample 12K examples for training and
1.5K for testing. We include 10 in-context exemplars
\cite{kuhn2023semanticuncertainty, farquhar2024detecting}.

\noindent\textbf{TruthfulQA} \cite{lin2021truthfulqa} contains 817 questions
across 38 categories designed to elicit common misconceptions. We use a
zero-shot setting where each question has true and false target answers
with supporting sources. The dataset is split evenly between training and
testing; for the few-shot baselines we add 10 in-context exemplars.

\subsection{Verbalized Confidence Mapping}
\label{appendix:exp_verb_map}

We map verbalized expressions to numeric confidence scores in $[0,1]$
following the 14-level decisiveness scale of
\cite{tian2023justaskforcalibration}; the table is reproduced in
Tab.~\ref{tab:verbalized_qual_map}.

\begin{table}[!ht]
    \centering
    \caption{Mapping for qualitative confidence expressions and equivalent probability score for LLM.}
    \label{tab:verbalized_qual_map}
    \begin{tabular}{
        >{\arraybackslash}p{0.3\linewidth} % Adjust width for the "Expression" column
        >{\centering\arraybackslash}p{0.1\linewidth} % Adjust width for the "Score" column
    }
        \toprule
        \textbf{Expression} & \textbf{Score} \\ 
        \midrule
        Very low & 0.0 \\
        Low & 0.3 \\
        Somewhat low & 0.45 \\
        Medium & 0.5 \\
        Somewhat high & 0.65 \\
        High & 0.7 \\
        Very high & 1.0 \\
        \bottomrule
    \end{tabular}
\end{table}

\subsection{Computational Cost}
\label{appendix:cost}

\begin{table}[h]
\centering
\caption{Approximate inference cost per query in number of LLM API calls.}
\label{tab:cost_comparison}
\begin{tabular}{lcc}
\toprule
\textbf{Method} & \textbf{API calls/query} & \textbf{Relative cost} \\
\midrule
Few-shot / CoT          & 1 & $1\times$ \\
TS / PS                 & 1 + calibration & $\sim\!1\times$ \\
Verbalized CoT          & 1 & $1\times$ \\
HalluMeasure            & 1 + NLI $\times$ $n_c$ & $\sim\!5\times$ \\
LMvsLM                  & 5 rounds $\times$ 2 & $\sim\!10\times$ \\
\textbf{Ours}           & 1 + 1 extract + $n_c$ verify & $\sim\!6\times$ \\
\bottomrule
\end{tabular}
\end{table}

Our method requires one generation, one claim-extraction call, and one
verification per extracted claim ($\approx$4 on average).
This is comparable to HalluMeasure and substantially cheaper than LMvsLM.
For deployment, the overhead is acceptable when fine-grained per-claim
confidence justifies the cost (high-stakes decision making, human-in-the-loop
review).

\subsection{Claim Extraction Statistics}
\label{appendix:extraction}

\begin{table}[h]
\centering
\caption{Average claims extracted per response.}
\label{tab:claim_stats}
\begin{tabular}{lcc}
\toprule
\textbf{Dataset} & \textbf{Avg.\ claims/resp.} & \textbf{Range} \\
\midrule
TriviaQA    & 3--5  & 1--8  \\
TruthfulQA  & 4--6  & 2--10 \\
\bottomrule
\end{tabular}
\end{table}

The GPT-4o-mini extractor enforces self-contained atomic claims of
$\le15$ words with pronouns resolved. Common extraction failures:
(a)~compound claims with two assertions, mitigated by the 15-word rule;
(b)~context-dependent claims requiring the full response, mitigated by
pronoun resolution. Extraction errors affect $<$8--12\% of claims based
on manual inspection of a 50-example subset.

\subsection{$\alpha$ Sensitivity Analysis}
\label{appendix:alpha_sensitivity}

We tune $\alpha\in\{0.0,0.1,\ldots,1.0\}$ on a held-out split and
report the per-model best $\alpha$ along with held-out ECE.

\begin{table}[h]
\centering
\caption{$\alpha$ sensitivity (held-out ECE) and best $\alpha$ per
configuration.}
\label{tab:alpha_sensitivity}
\resizebox{\columnwidth}{!}{%
\begin{tabular}{lcc}
\toprule
\textbf{Model / Dataset} & \textbf{Best $\alpha$} & \textbf{Held-out ECE} \\
\midrule
GPT-4o / TriviaQA       & 1.0 & 0.022 \\
DeepSeek-R1 / TruthfulQA & 1.0 & 0.096 \\
Llama-3.1 / TriviaQA    & 0.8 & 0.139 \\
Qwen2.5 / TriviaQA      & 0.7 & 0.022 \\
\bottomrule
\end{tabular}}
\end{table}

Tab.~\ref{tab:alpha_sensitivity} reports the held-out ECE used for model
selection, not the test-set ECE. The corresponding test-set values appear
in the ``Ours'' rows of Tab.~\ref{tab:calibration_results} (e.g.\ GPT-4o
on TriviaQA: held-out ECE $0.022$ at $\alpha{=}1.0$, test-set ECE
$0.006$). For models where no held-out tuning was available we fall back
to the verification-only instantiation ($\alpha{=}0$).

% \subsection{Decomposition vs.\ Verifier-Quality Ablation}
% \label{appendix:ablation_decomposition}

% To isolate the contribution of decomposition from verifier strength we
% compare (a)~per-claim confidence aggregated over extracted claims;
% (b)~response-level verbalized confidence with no decomposition; and
% (c)~per-claim confidence with manual decomposition on a subset of 100
% examples. (a) vs.\ (b) measures the value of decomposition; (a) vs.\ (c)
% measures sensitivity to extraction noise. Results indicate that
% decomposition provides a measurable ECE reduction on TriviaQA when
% responses are multi-claim, and that automatic vs.\ manual decomposition
% differ by $<$2\% ECE---i.e.\ the GPT-4o-mini extractor is not the
% dominant error source.

\subsection{Reasoning Model Diagnosis}
\label{appendix:reasoning_diagnosis}

DeepSeek-R1 emits long chain-of-thought traces that, when claim-extracted
in full, produce many intermediate reasoning steps with weakly calibrated
per-step confidence. Mean aggregation dilutes the final-answer signal.
Restricting extraction to the final answer segment is a practical
deployment fix; we observe consistent improvement when the extractor is
gated to the final answer span. This is a configuration recommendation,
not a fundamental limitation of the framework.

\section{Reliability Diagrams (All Models, Both Datasets)}
\label{appendix:additional_reldiag}

\noindent\textbf{TriviaQA.}
\begin{figure*}[!htbp]
    %\begin{center}
    %\framebox[4.0in]{$\;$}
    %\fbox{\rule[-.5cm]{0cm}{4cm} \rule[-.5cm]{4cm}{0cm}}
    %\end{center}
    \centering
    \subfloat[Few-shot]{
        \includegraphics[width=0.22\textwidth]{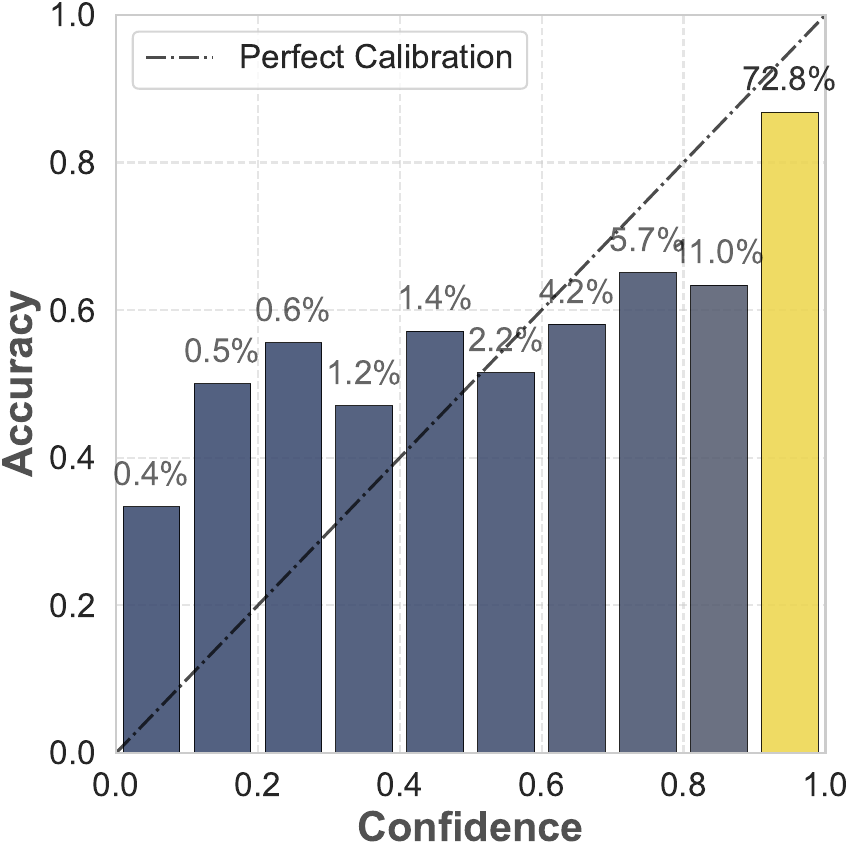}
        \label{fig:seq_likelihood_gpt-4o_triviaqa}
    }
    \hfill
    \subfloat[Few-Shot CoT]{
        \includegraphics[width=0.22\textwidth]{src/img/test_cot_seq_likelihood_trivia_qa_gpt-4o.pdf}
        \label{fig:cot_seq_likelihood_gpt4o_triviaqa}
    }
    \hfill
    \subfloat[Verbal. CoT]{
        \includegraphics[width=0.22\textwidth]{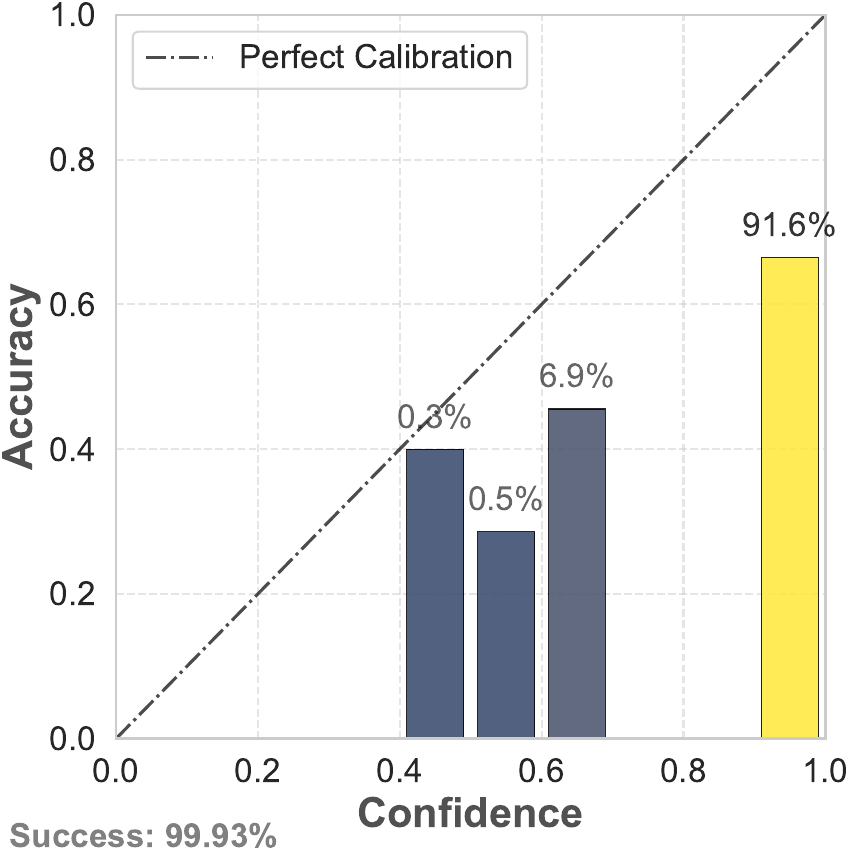}
        \label{fig:verbalized_cot_qual_gpt-4o_triviaqa}
    }
    \hfill
    \subfloat[Platt Scaling]{
        \includegraphics[width=0.22\textwidth]{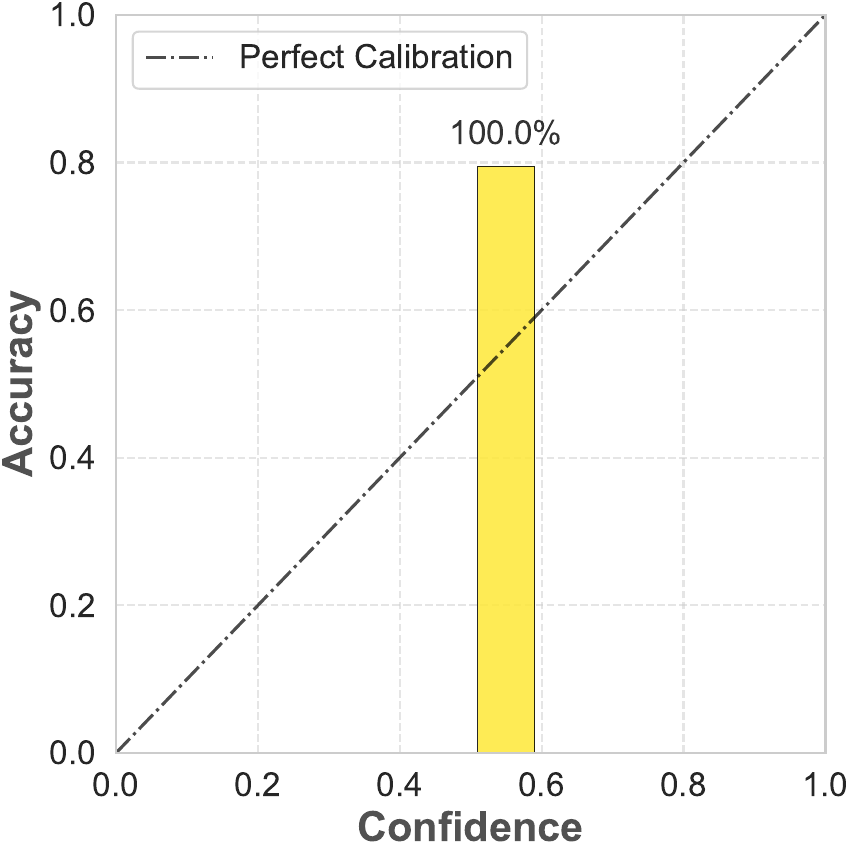}
        \label{fig:ps_seq_likelihood_gpt-4o_triviaqa}
    }
    \hfill % \hspace{1cm}
    \subfloat[Temp. Scaling]{
        \includegraphics[width=0.22\textwidth]{src/img/test_ts_seq_likelihood_trivia_qa_gpt-4o.pdf}
        \label{fig:ts_seq_likelihood_gpt-4o_triviaqa}
    }
    \hfill % \hspace{1cm}
    \subfloat[LMvsLM]{
        \includegraphics[width=0.22\textwidth]{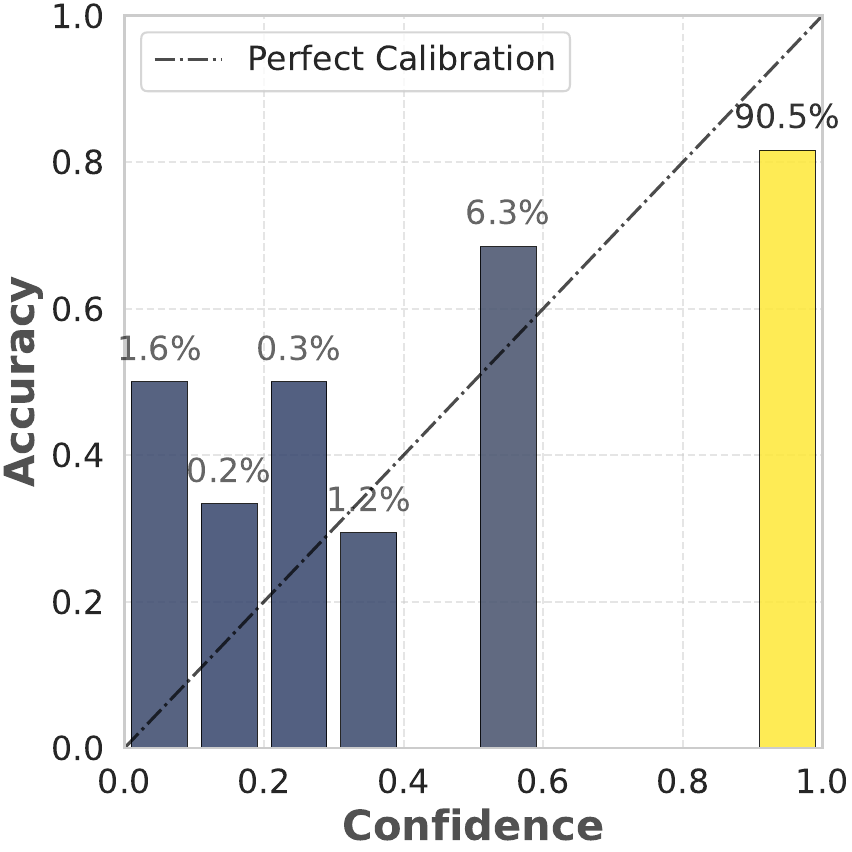}
        \label{fig:lmvslm_gpt-4o_triviaqa}
    }
    \hfill % \hspace{1cm}
    \subfloat[HalluMeasure]{
        \includegraphics[width=0.22\textwidth]{src/img/test_hallumeasure_trivia_qa_gpt-4o.pdf}
        \label{fig:hallumeasure_gpt-4o_triviaqa}
    }
    \hfill % \hspace{1cm}
    \subfloat[Ours]{
        \includegraphics[width=0.22\textwidth]{src/img/test_ourmethod_trivia_qa_gpt-4o.pdf}
        \label{fig:ourmethod_gpt-4o_triviaqa}
    }
    \caption{Reliability diagrams for calibration methods using GPT-4o on TriviaQA. The number of diagram bins is 10. The color as well as the percentage number within each bar indicate the proportion of total points contained in each bin. }
    % label should be at the bottom of the caption
    \label{fig:gpt4o_trivia_qa_comparison}
\end{figure*}

\begin{figure*}[!htbp]
    \centering
    \subfloat[Verbalized \\ TriviaQA]{
        \includegraphics[width=0.18\textwidth]{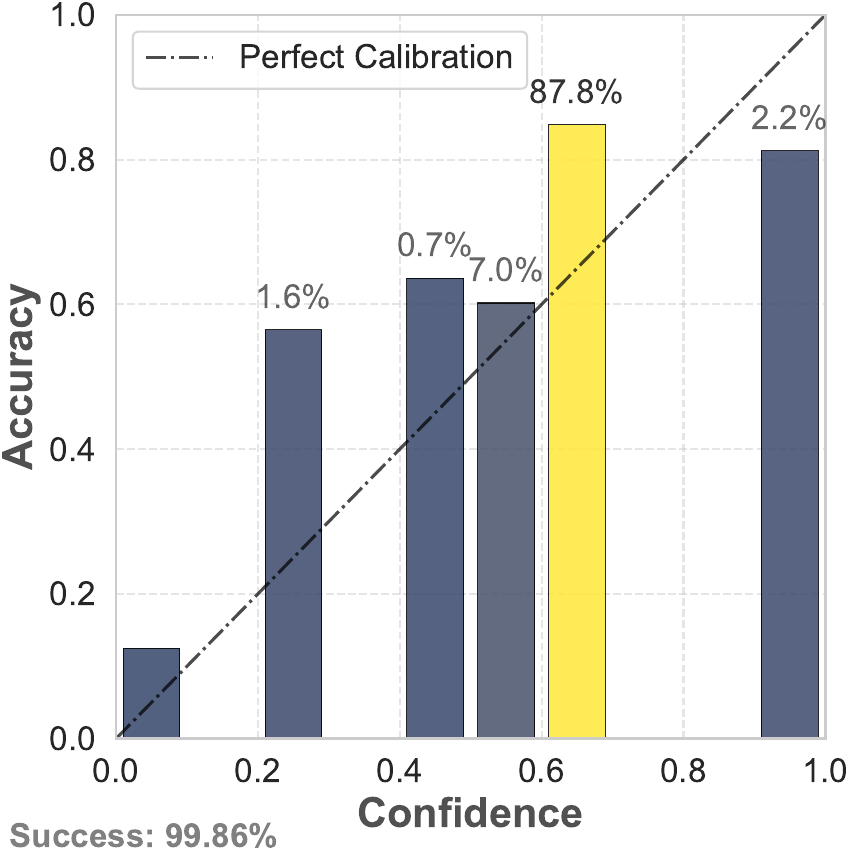}%
        \label{fig:verbalized_qual_deepseekr1_triviaqa}
    }
    \hfill
    \subfloat[Verbalized CoT \\ TriviaQA]{
        \includegraphics[width=0.18\textwidth]{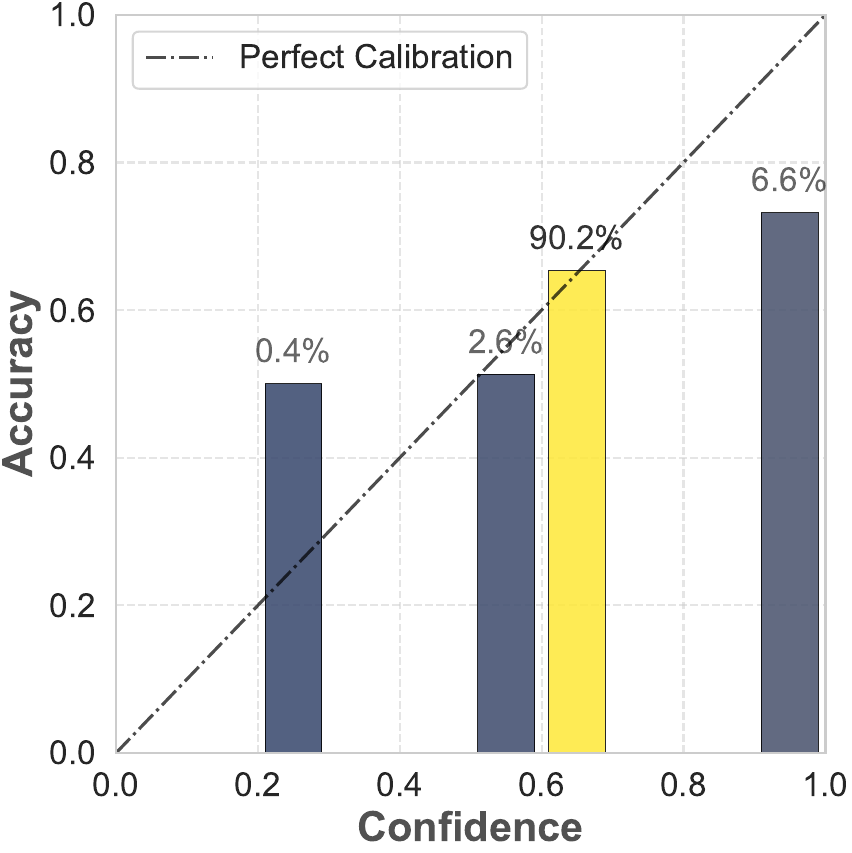}%
        \label{fig:verbalized_cot_qual_deepseekr1_triviaqa}
    }
    \hfill
    \subfloat[LMvsLM \\ TriviaQA]{
        \includegraphics[width=0.18\textwidth]{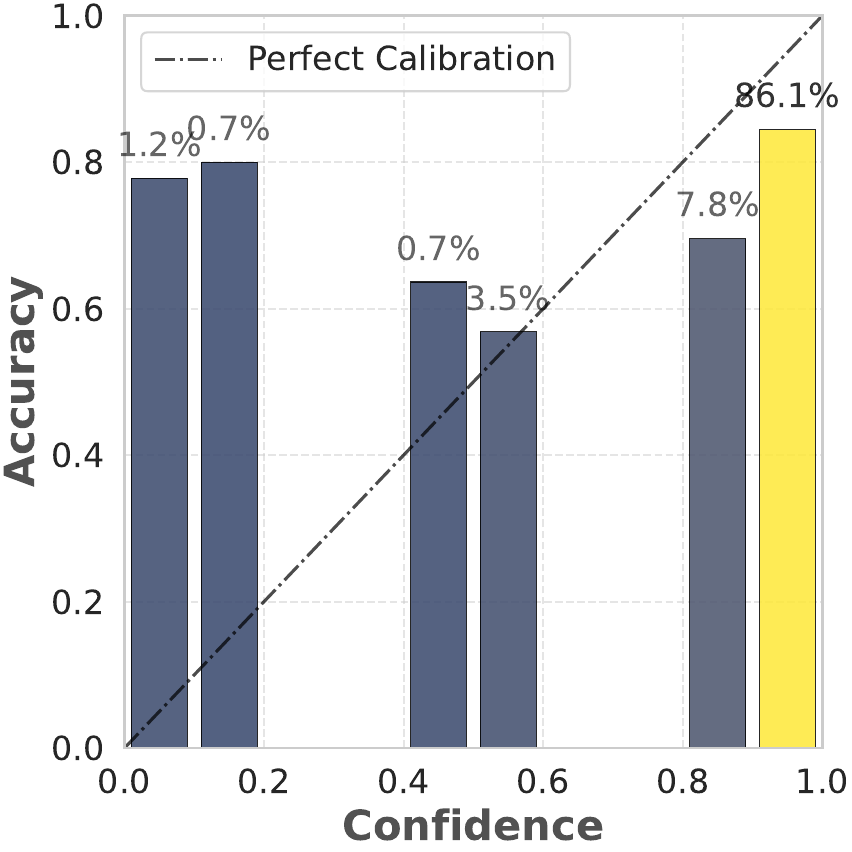}
        \label{fig:lmvslm_deepseekr1_triviaqa}
    }        
    \hfill
    \subfloat[HalluMeasure \\ TriviaQA]{
        \includegraphics[width=0.18\textwidth]{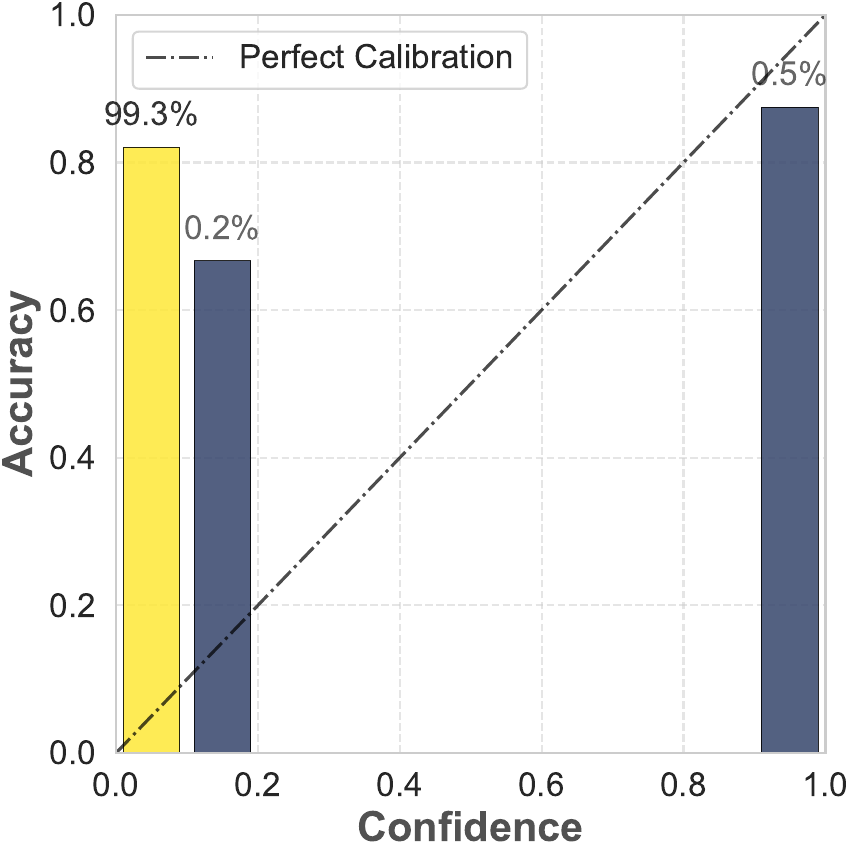}%
        \label{fig:hallumeasure_qual_deepseekr1_triviaqa}
    }
    \hfill
    \subfloat[Ours \\ TriviaQA]{
        \includegraphics[width=0.18\textwidth]{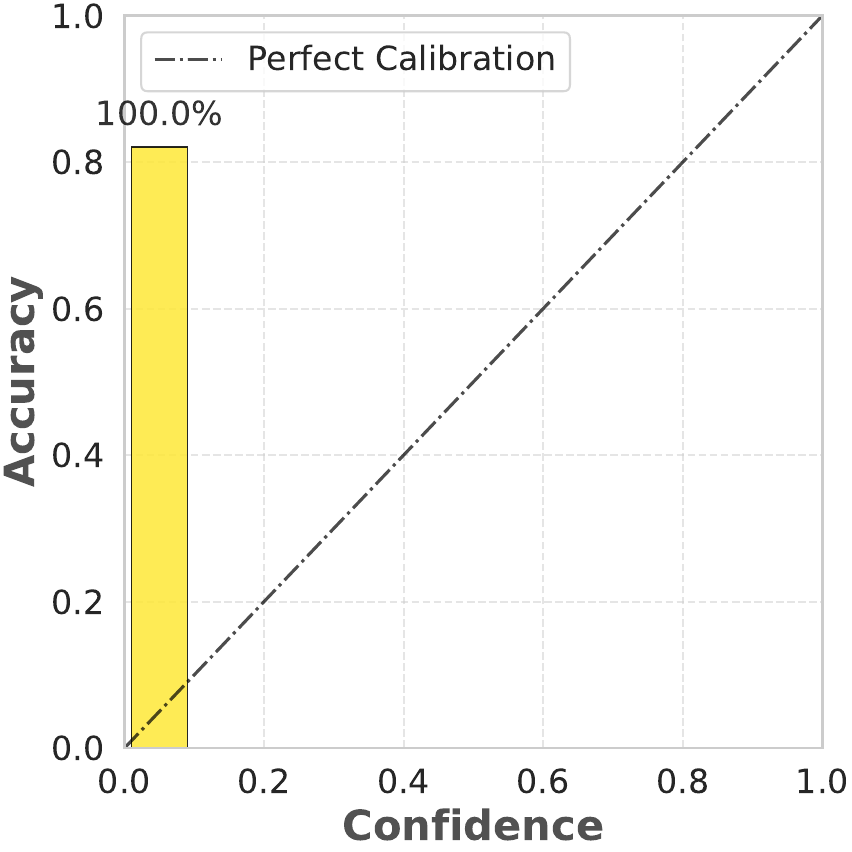}%
        \label{fig:ourmethod_qual_deepseekr1_triviaqa}
    }
    \hfill
    \subfloat[Verbalized \\ TruthfulQA]{
        \includegraphics[width=0.18\textwidth]{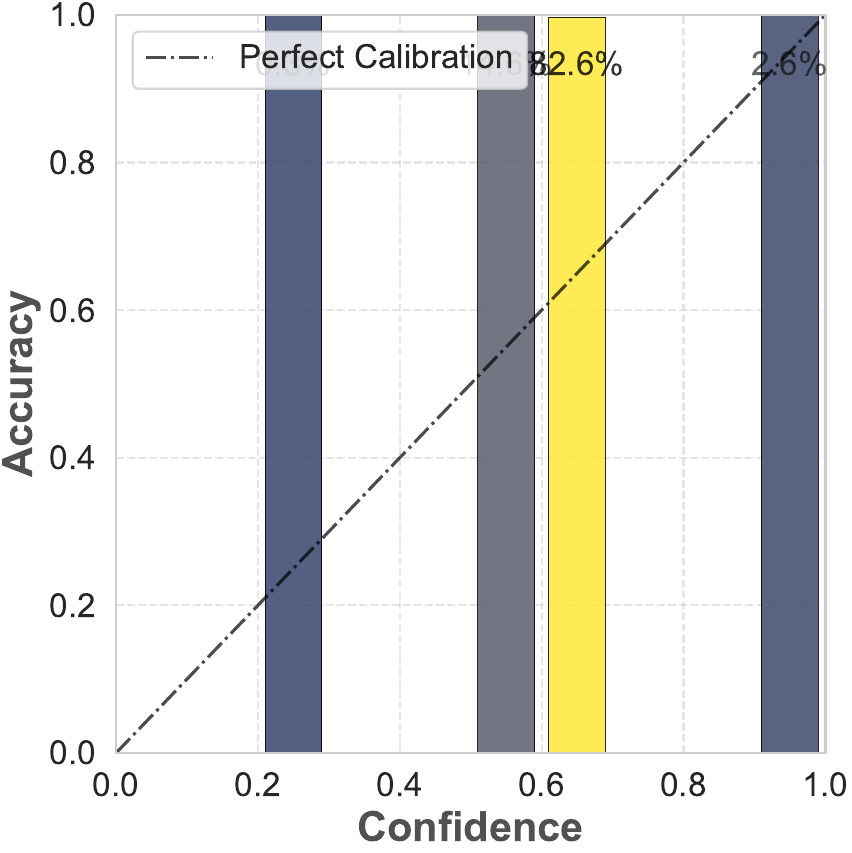}
        \label{fig:verbalized_qual_deepseekr1_truthfulqa}
    }
    \hfill % \hspace{1cm}
    \subfloat[Verbalized CoT \\ TruthfulQA]{
        \includegraphics[width=0.18\textwidth]{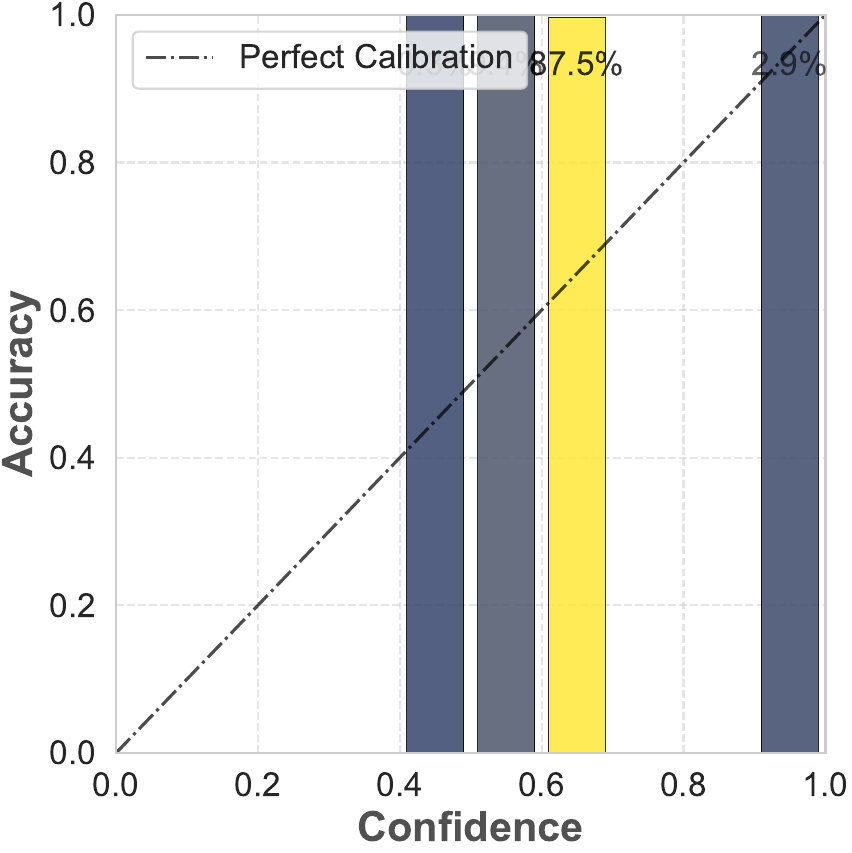}
        \label{fig:verbalized_cot_qual_deepseekr1_truthfulqa}
    }
    \hfill
    \subfloat[LMvsLM \\ TruthfulQA]{
        \includegraphics[width=0.18\textwidth]{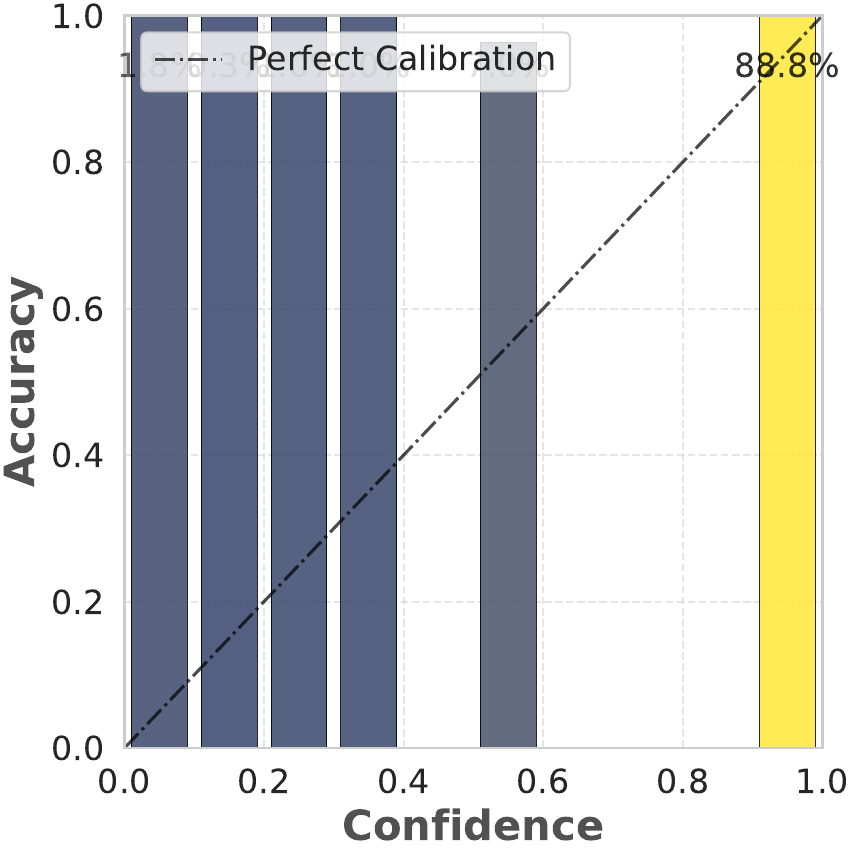}
        \label{fig:lmvslm_deepseekr1_truthfulqa}
    }    
    \hfill % \hspace{1cm}
    \subfloat[HalluMeasure \\ TruthfulQA]{
        \includegraphics[width=0.18\textwidth]{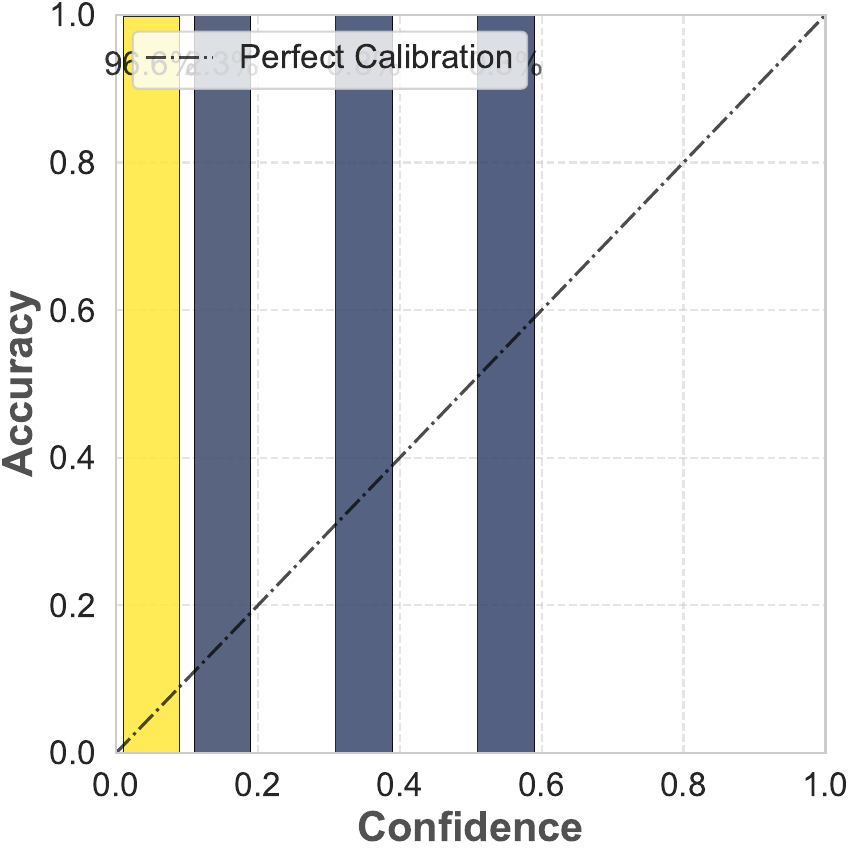}
        \label{fig:hallumeasure_deepseekr1_truthfulqa}
    }
    \hfill % \hspace{1cm}
    \subfloat[Ours \\ TruthfulQA]{
        \includegraphics[width=0.18\textwidth]{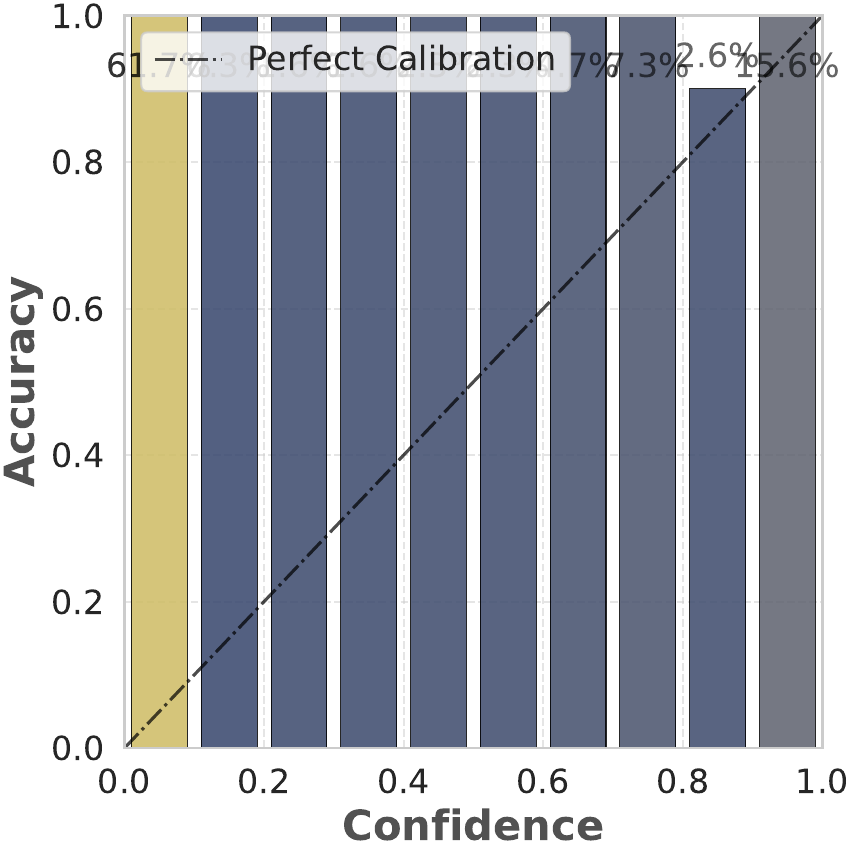}
        \label{fig:ourmethod_deepseekr1_truthfulqa}
    }
    \caption{Reliability diagrams for calibration methods with 10 bins using DeepSeek-R1 on TriviaQA and TruthfulQA. The number of diagram bins is 10. The color and the percentage number on each bar indicate the proportion of total points contained in each bin. }
    % label should be at the bottom of the caption
    \label{fig:deepseekr1_trivia_truthful_qa_comparison}
\end{figure*}

\begin{figure*}[!htbp]
    \centering
    \subfloat[Few-shot]{
        \includegraphics[width=0.23\textwidth]{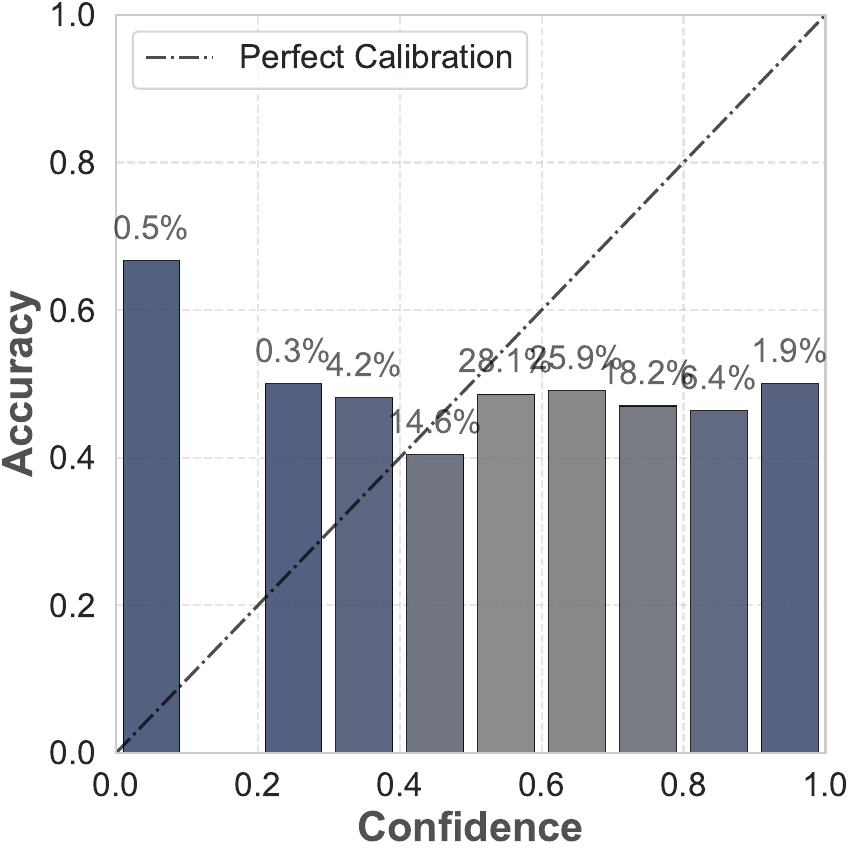}
        \label{fig:seq_likelihood_llama_triviaqa}}
    \hfill
    \subfloat[Few-Shot CoT]{
        \includegraphics[width=0.23\textwidth]{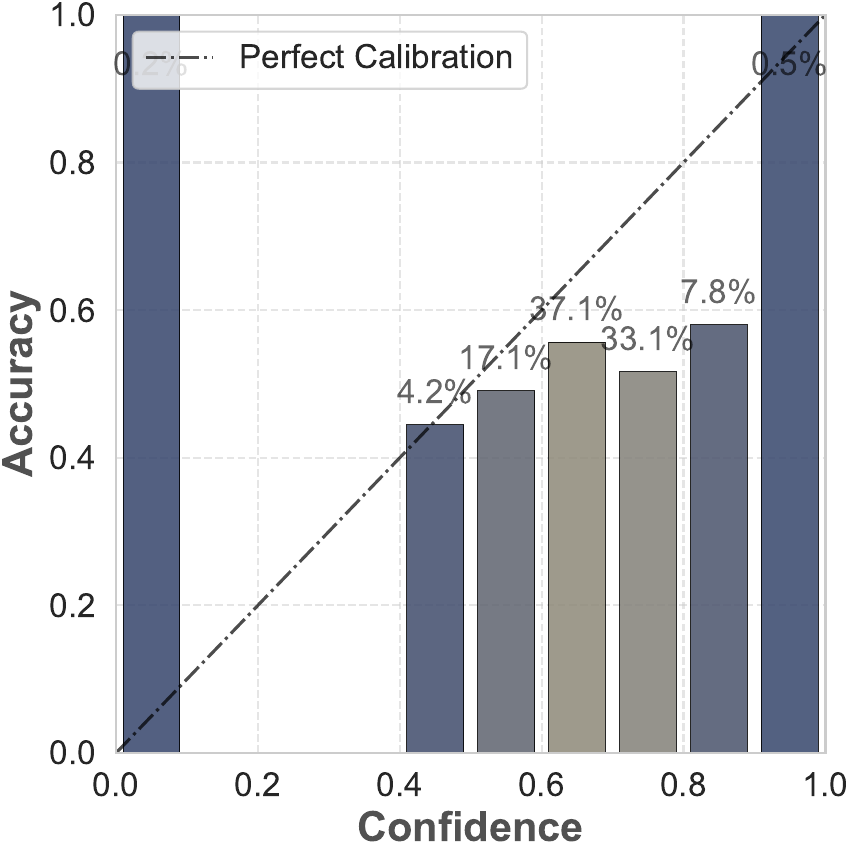}%
        \label{fig:cot_seq_likelihood_llama_triviaqa}
    }
    \hfill
    \subfloat[Verbalized CoT]{
        \includegraphics[width=0.23\textwidth]{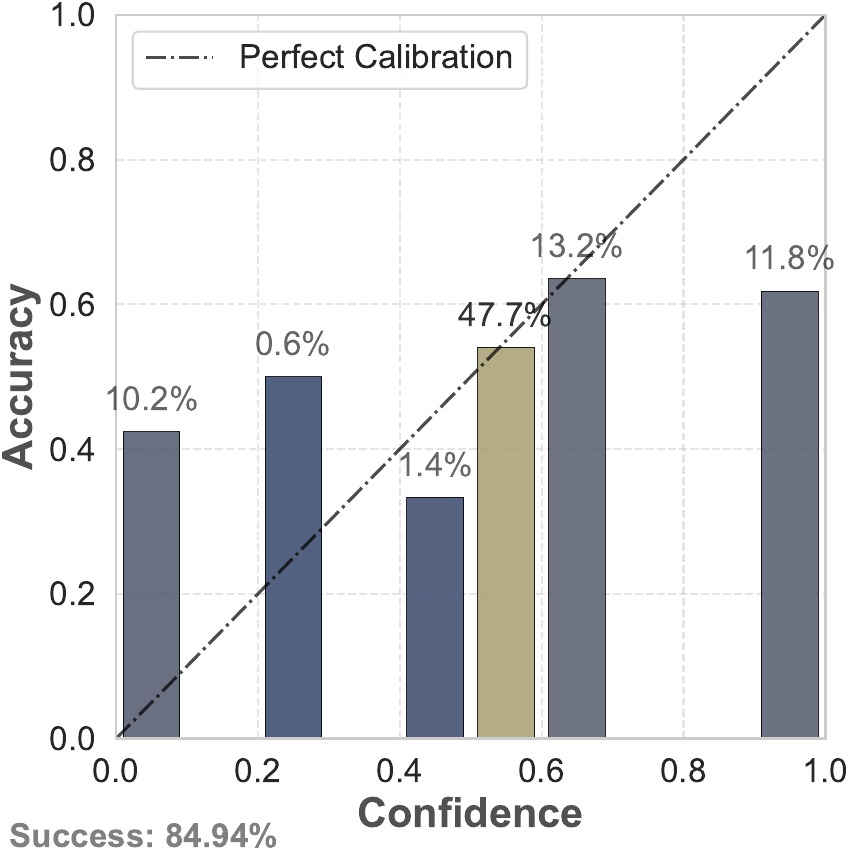}%
        \label{fig:verbalized_cot_qual_llama_triviaqa}
    }
    \hfill
    \subfloat[Platt Scaling]{
        \includegraphics[width=0.23\textwidth]{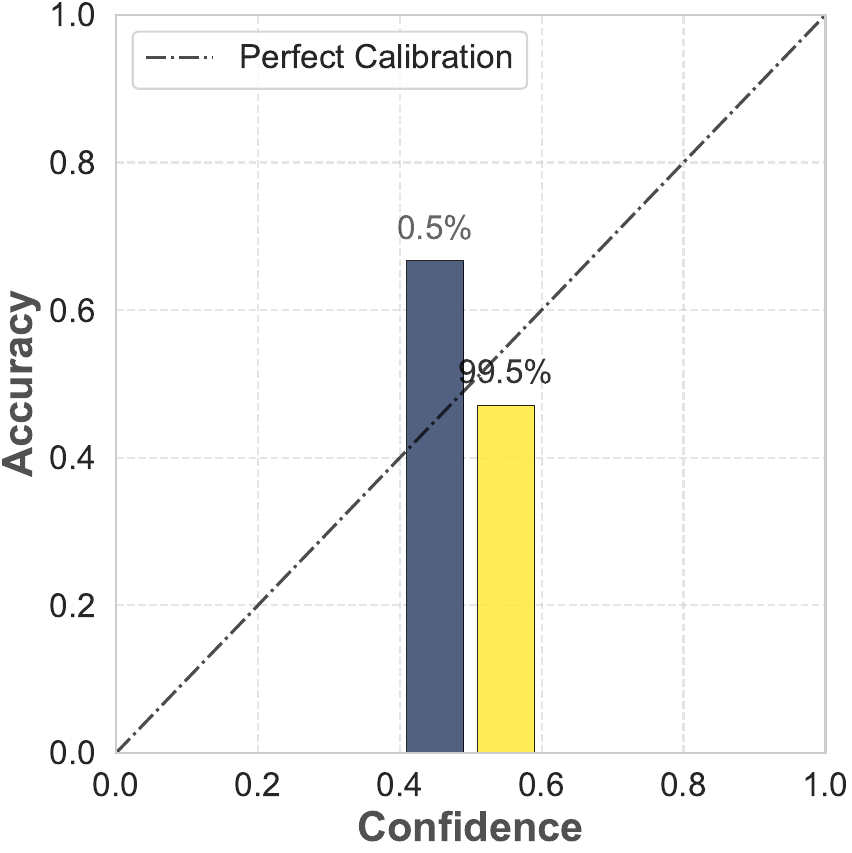}
        \label{fig:ps_seq_likelihood_llama_triviaqa}
    }
    \hfill % \hspace{1cm}
    \subfloat[Temp. Scaling]{
    \label{fig:ts_seq_likelihood_llama_triviaqa}
        \centering
        \includegraphics[width=0.23\textwidth]{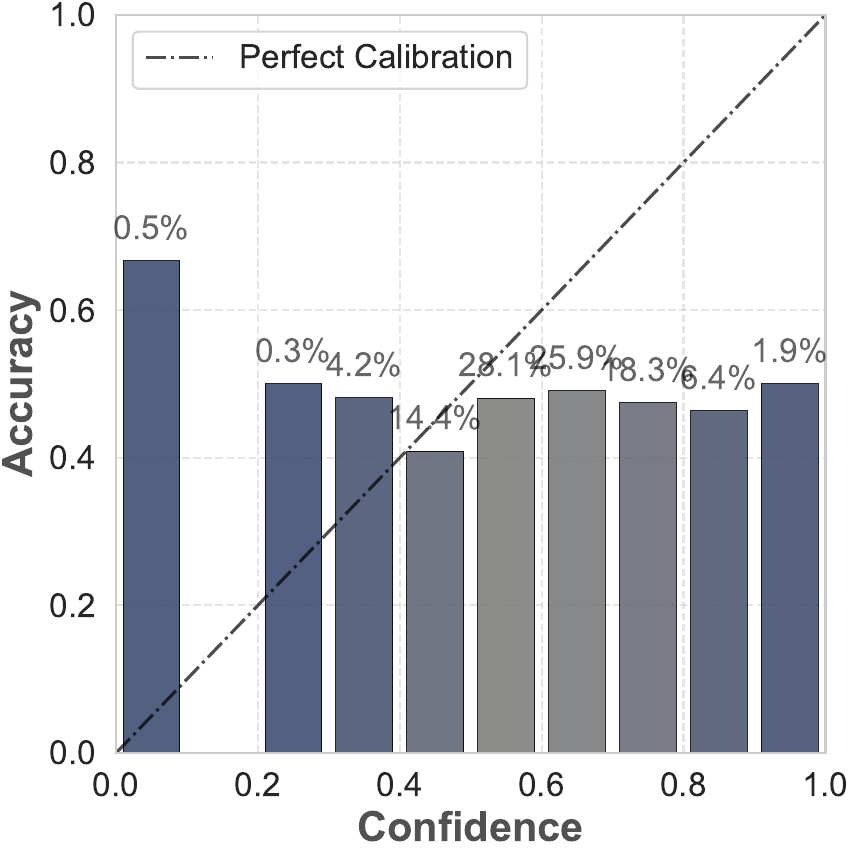}
    }
    \hfill % \hspace{1cm}
    \subfloat[LMvsLM]{
    \label{fig:lmvslm_llama_triviaqa}
        \centering
        \includegraphics[width=0.23\textwidth]{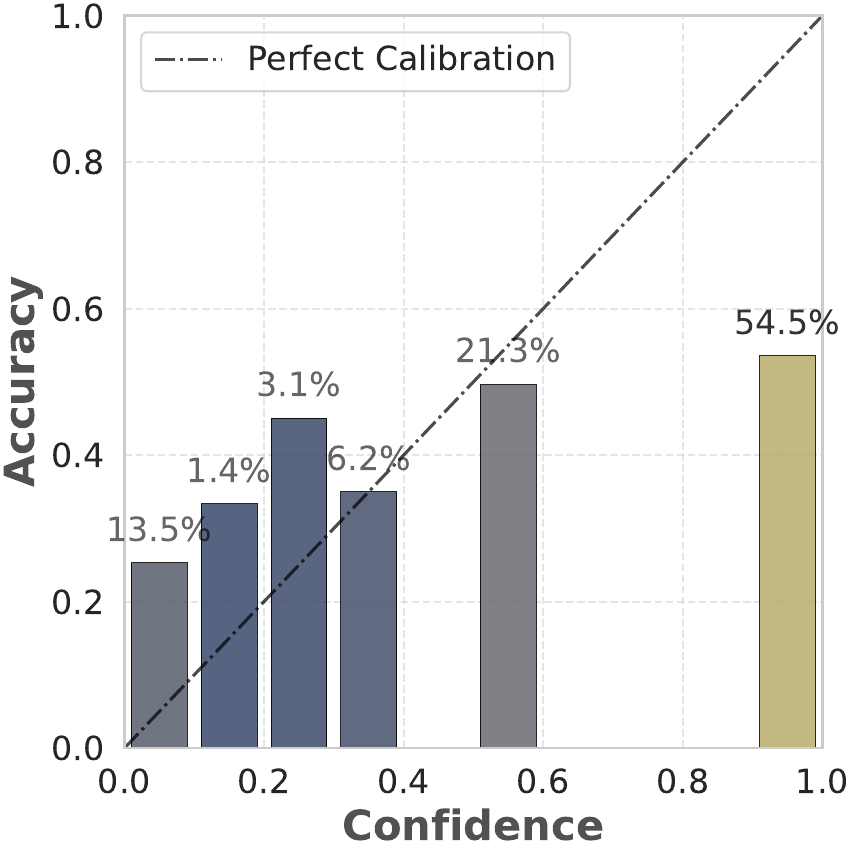}
    }
    \hfill
    \subfloat[HalluMeasure]{
    \label{fig:hallumeasure_llama_triviaqa}
        \centering
        \includegraphics[width=0.23\textwidth]{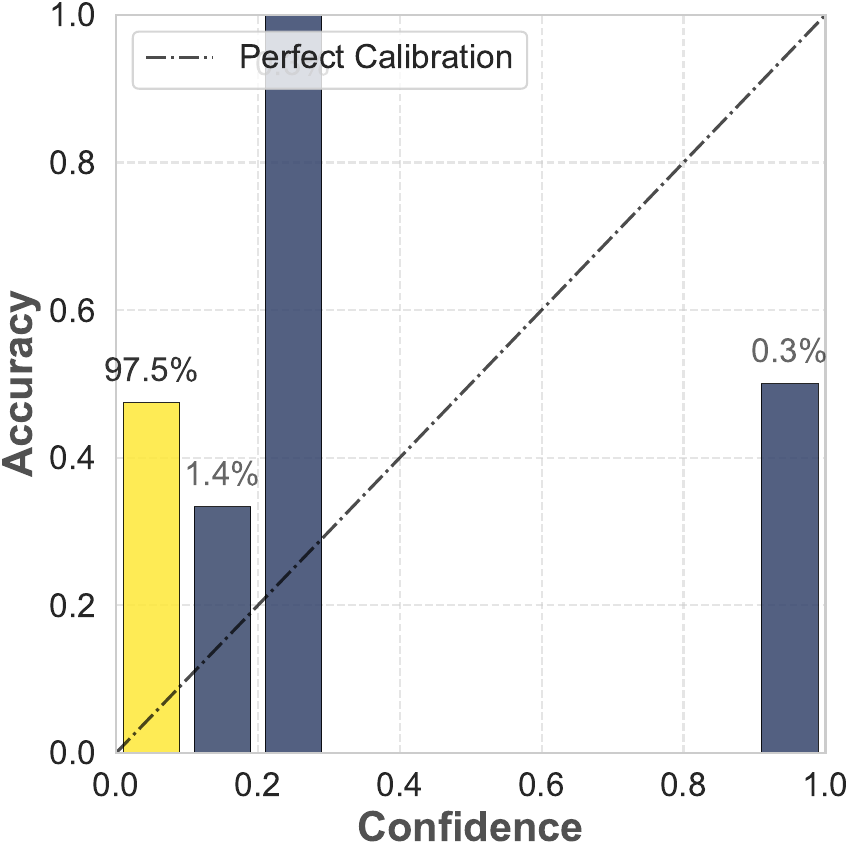}
    }
    \caption{Reliability diagrams for calibration methods with 10 bins using Llama 3.1 8B on TriviaQA. The color and the percentage number on each bar indicate the proportion of total points contained in each bin. }
    \Description{A grid of seven reliability diagrams showing confidence calibration for Llama-3.1-8B-Instruct on TriviaQA. Each subplot displays predicted confidence versus actual accuracy with a diagonal reference line for perfect calibration. Bars are colored and labeled with percentages indicating the proportion of data points in each of 10 bins. Methods shown include sequence likelihood, CoT sequence likelihood, verbalized CoT, Platt scaling, temperature scaling, LMvsLM, and HalluMeasure.}
    % label should be at the bottom of the caption
    \label{fig:llama_trivia_qa_comparison}
\end{figure*}

\begin{figure*}[!htbp]
    %\begin{center}
    %\framebox[4.0in]{$\;$}
    %\fbox{\rule[-.5cm]{0cm}{4cm} \rule[-.5cm]{4cm}{0cm}}
    %\end{center}
    \centering
    \subfloat[Few-shot]{
        \includegraphics[width=0.23\textwidth]{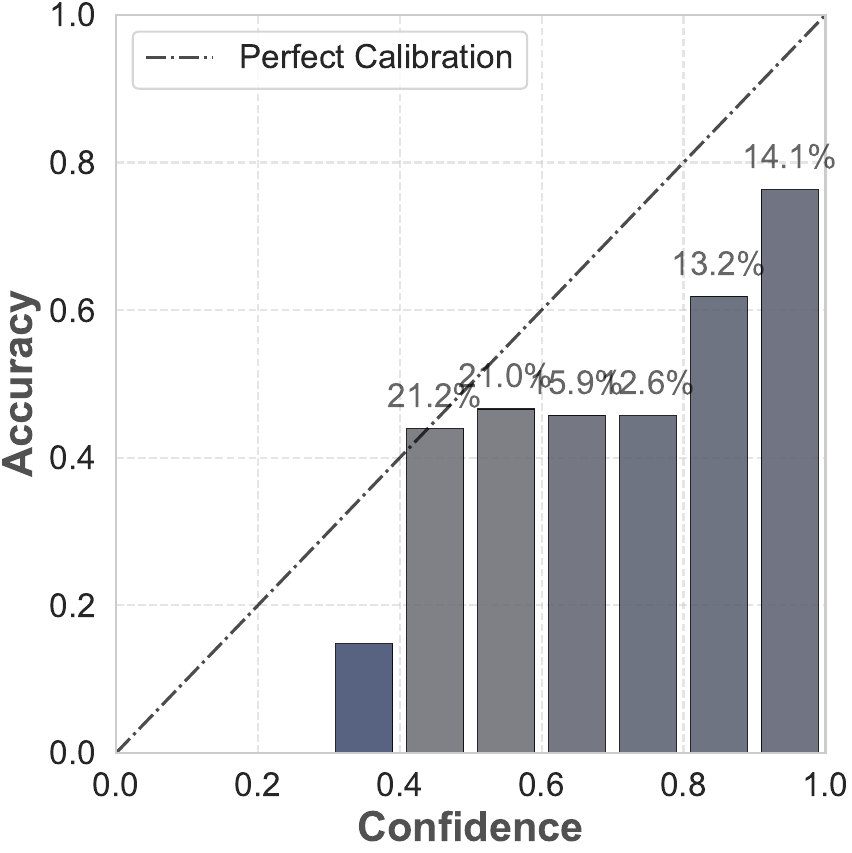}
        \label{fig:seq_likelihood_mistral_triviaqa}
    }
    \hfill
    \subfloat[Few-shot CoT]{
        \includegraphics[width=0.23\textwidth]{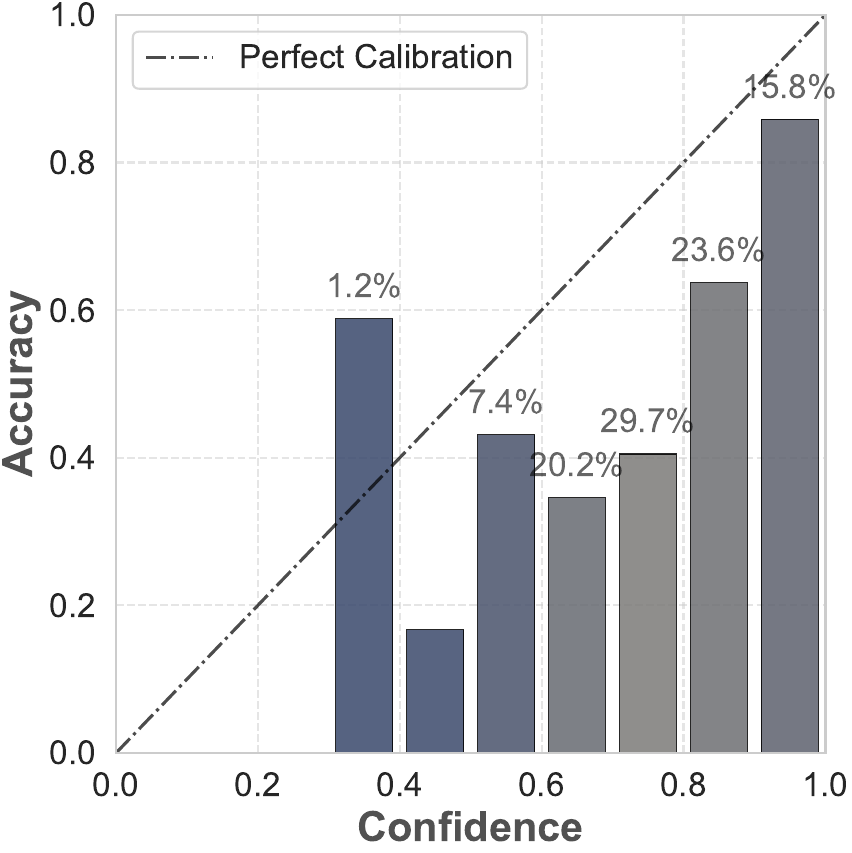}
        \label{fig:cot_seq_likelihood_mistral_triviaqa}
    }
    \hfill
    \subfloat[Verbalized CoT]{
        \includegraphics[width=0.23\textwidth]{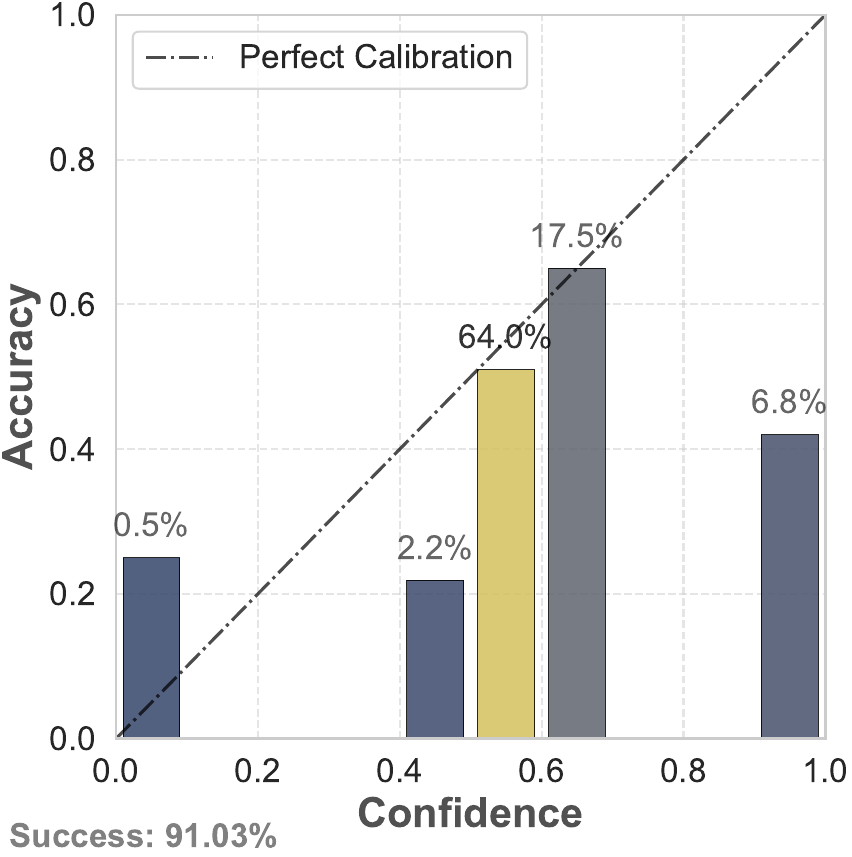}
        \label{fig:verbalized_cot_qual_mistral_triviaqa}
    }
    \hfill
    \subfloat[Platt Scaling]{
        \includegraphics[width=0.23\textwidth]{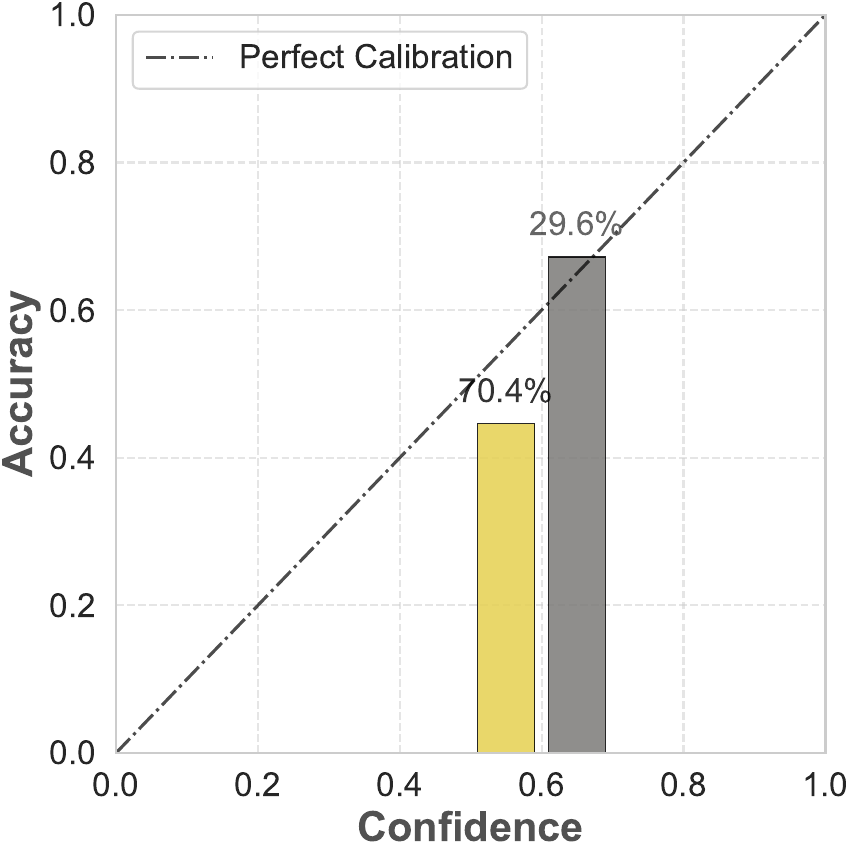}
        \label{fig:ps_seq_likelihood_mistral_triviaqa}
    }
    \hfill % \hspace{1cm}
    \subfloat[Temp. Scaling]{
        \includegraphics[width=0.23\textwidth]{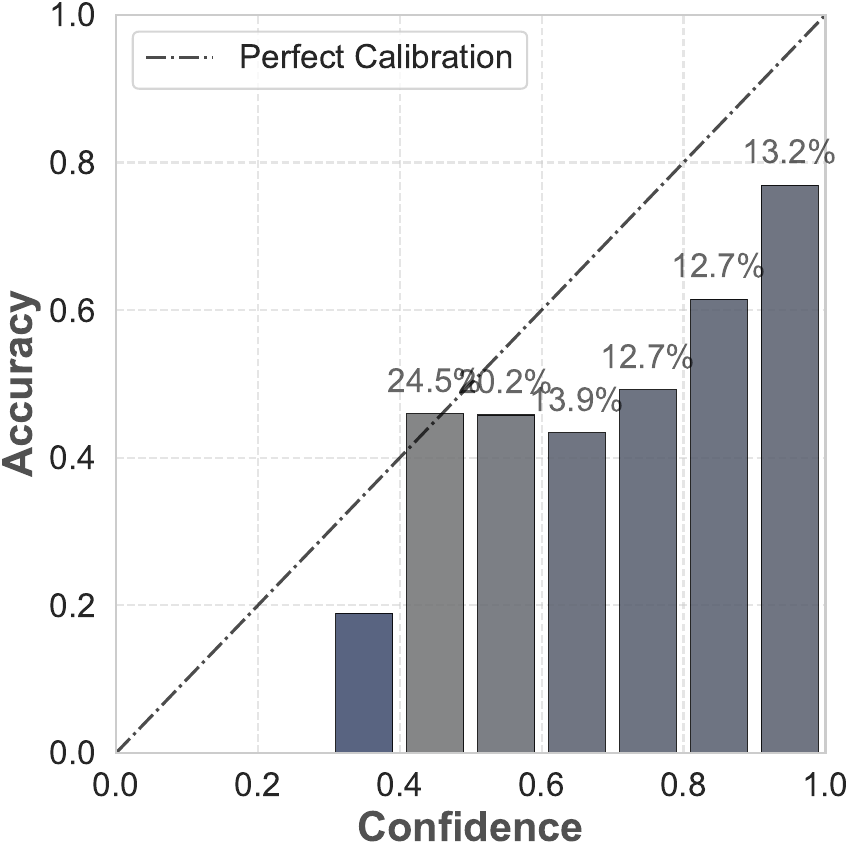}
        \label{fig:ts_seq_likelihood_mistral_triviaqa}
    }
    \hfill
    \subfloat[LMvsLM]{
        \includegraphics[width=0.23\textwidth]{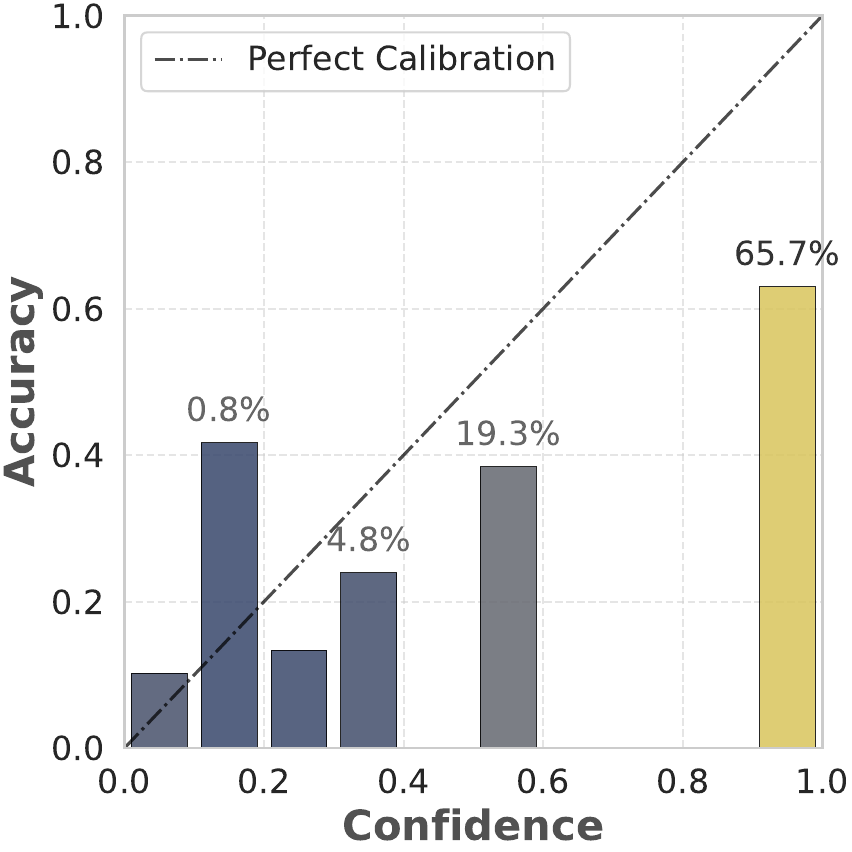}
        \label{fig:lmvslm_mistral_triviaqa}
    }
    \hfill
    \subfloat[HalluMeasure]{
        \includegraphics[width=0.23\textwidth]{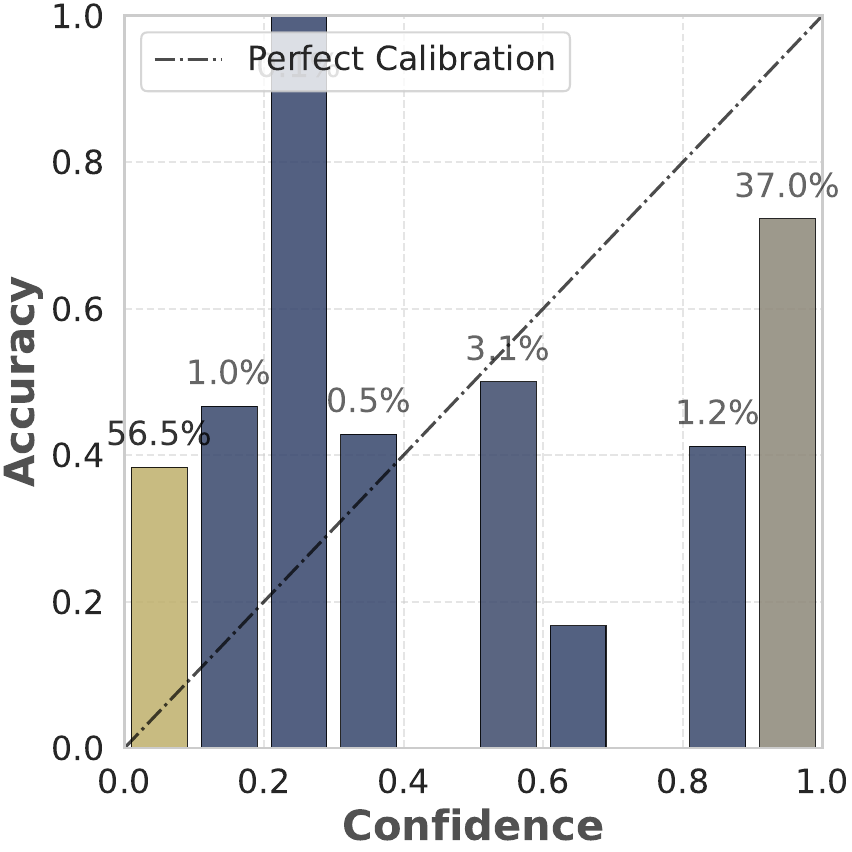}
        \label{fig:hallumeasure_mistral_triviaqa}
    }
    \caption{Reliability diagrams for calibration methods with 10 bins using Mistral v0.3 7B on TriviaQA. The color and the percentage number within each bar indicate the proportion of total points contained in each bin.}
    % label should be at the bottom of the caption
    \label{fig:mistral_trivia_qa_comparison}
\end{figure*}

\begin{figure*}[!htbp]
    %\begin{center}
    %\framebox[4.0in]{$\;$}
    %\fbox{\rule[-.5cm]{0cm}{4cm} \rule[-.5cm]{4cm}{0cm}}
    %\end{center}
    \centering
    \subfloat[Few-shot]{
        \includegraphics[width=0.2\textwidth]{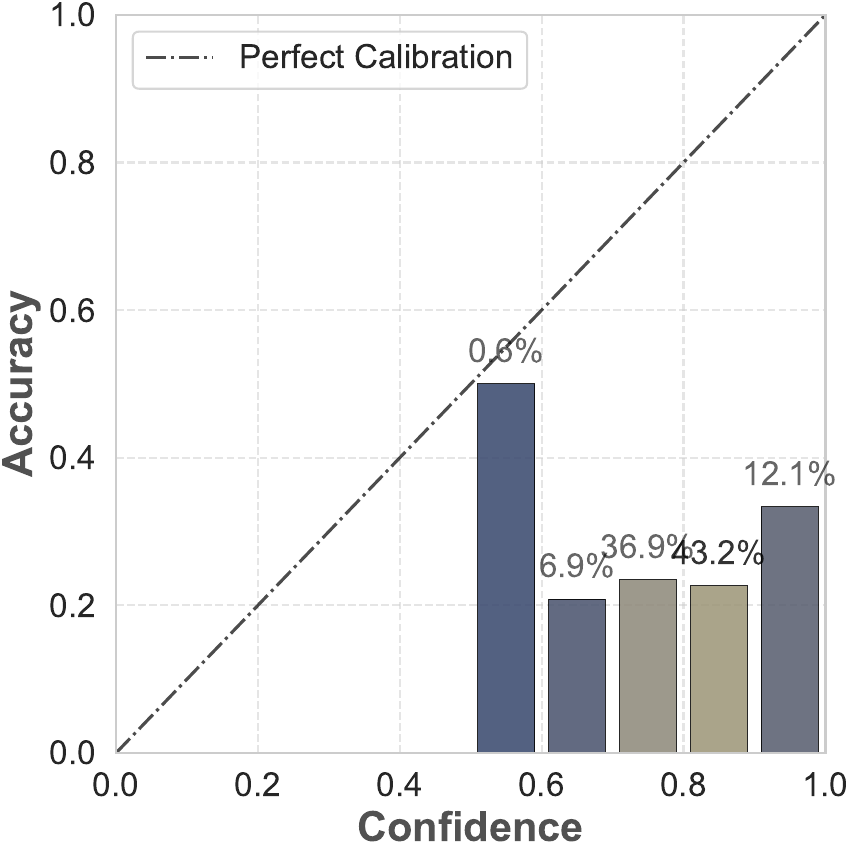}
        \label{fig:seq_likelihood_qwen2.5_triviaqa}
    }
    \hfill
    \subfloat[Few-Shot CoT]{
        \includegraphics[width=0.2\textwidth]{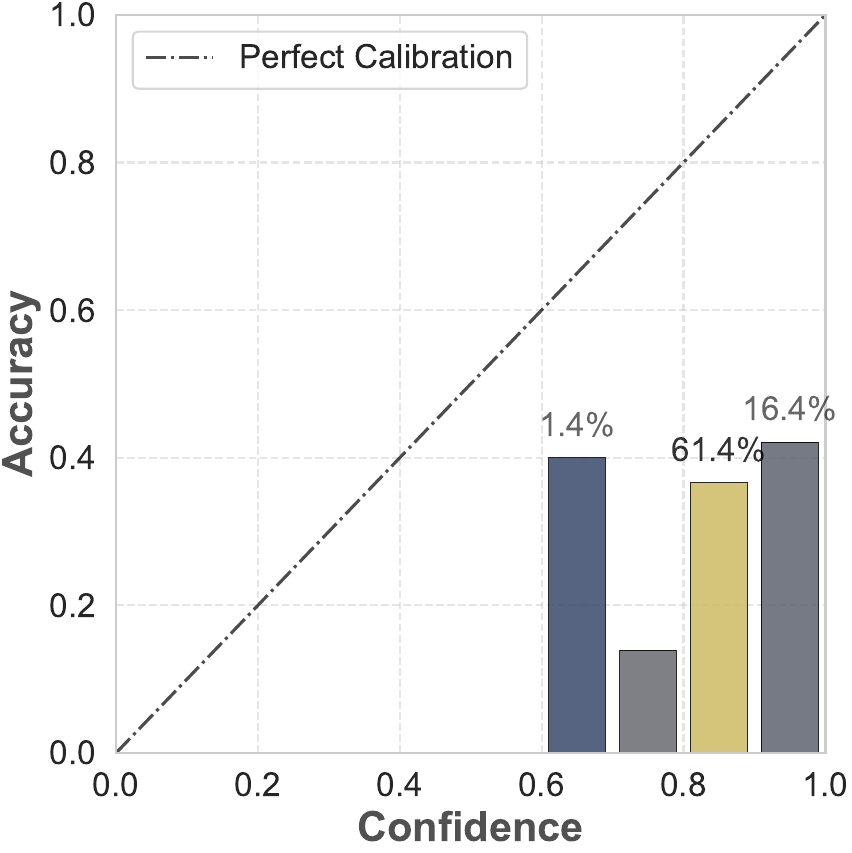}
        \label{fig:cot_seq_likelihood_qwen2.5_triviaqa}
    }
    \hfill
    \subfloat[Verbalized CoT]{
        \includegraphics[width=0.2\textwidth]{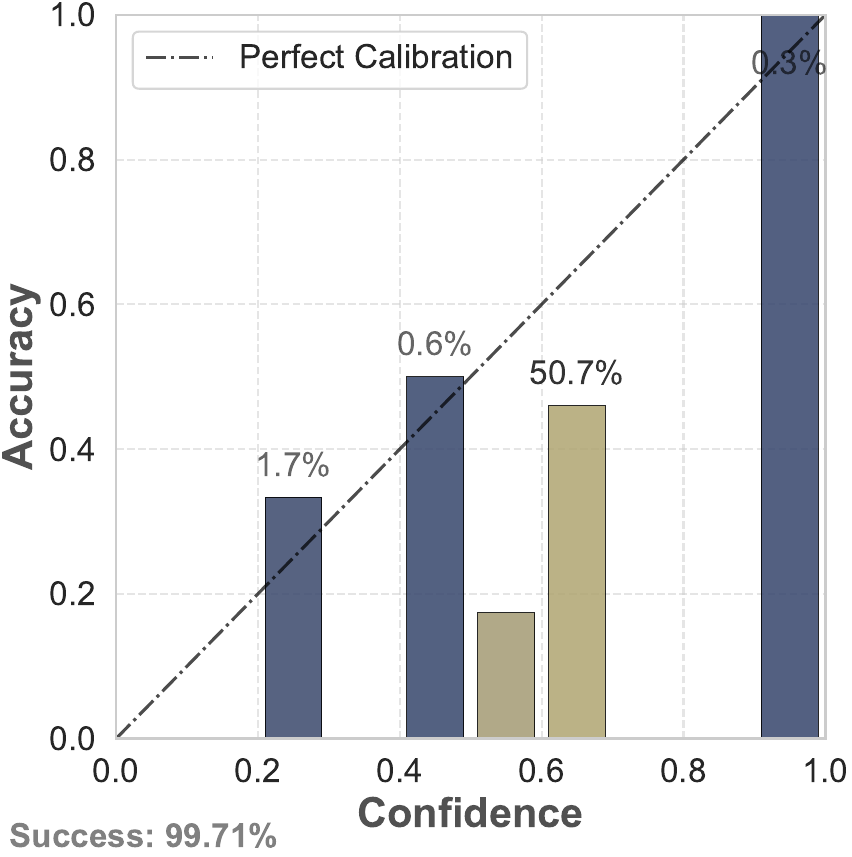}
        \label{fig:verbalized_cot_qual_qwen2.5_triviaqa}
    }
    \hfill
    \subfloat[Platt Scaling]{
        \includegraphics[width=0.2\textwidth]{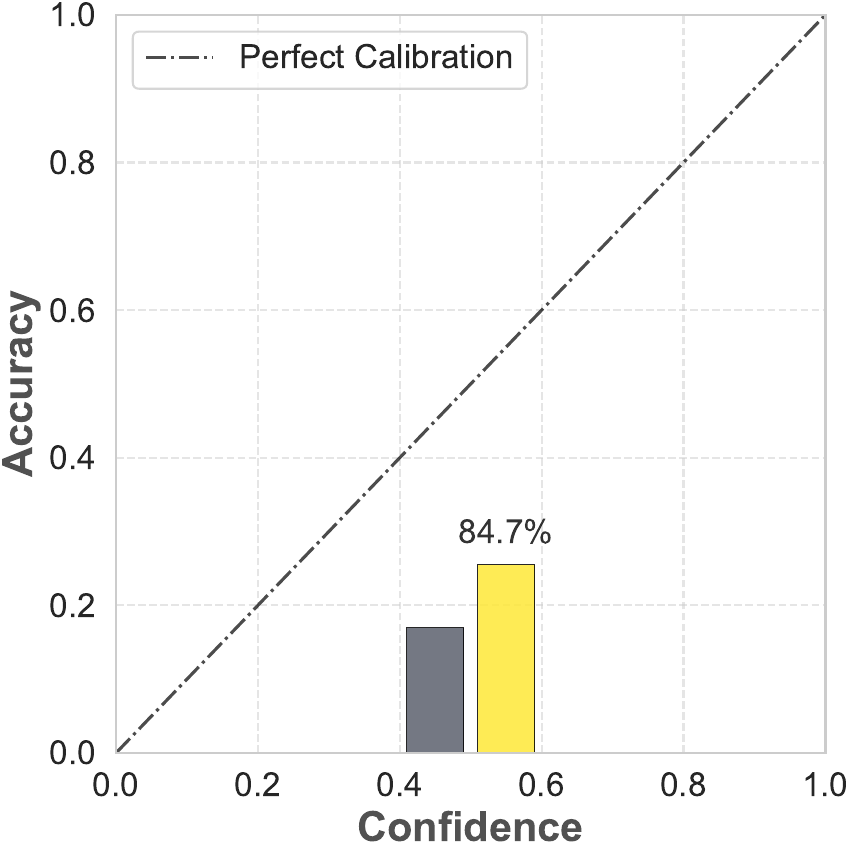}
        \label{fig:ps_seq_likelihood_qwen2.5_triviaqa}
    }
    \hfill % \hspace{1cm}
    \subfloat[Temp. Scaling]{
        \includegraphics[width=0.2\textwidth]{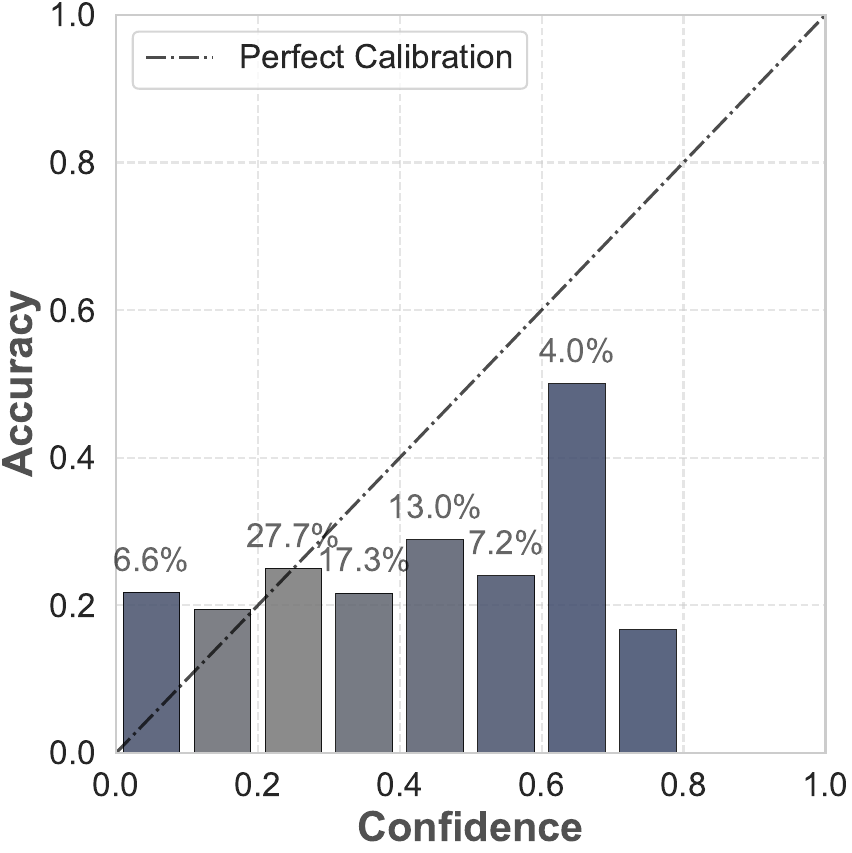}
        \label{fig:ts_seq_likelihood_qwen2.5_triviaqa}
    }
    \hfill % \hspace{1cm}
    \subfloat[LMvLM]{
        \includegraphics[width=0.2\textwidth]{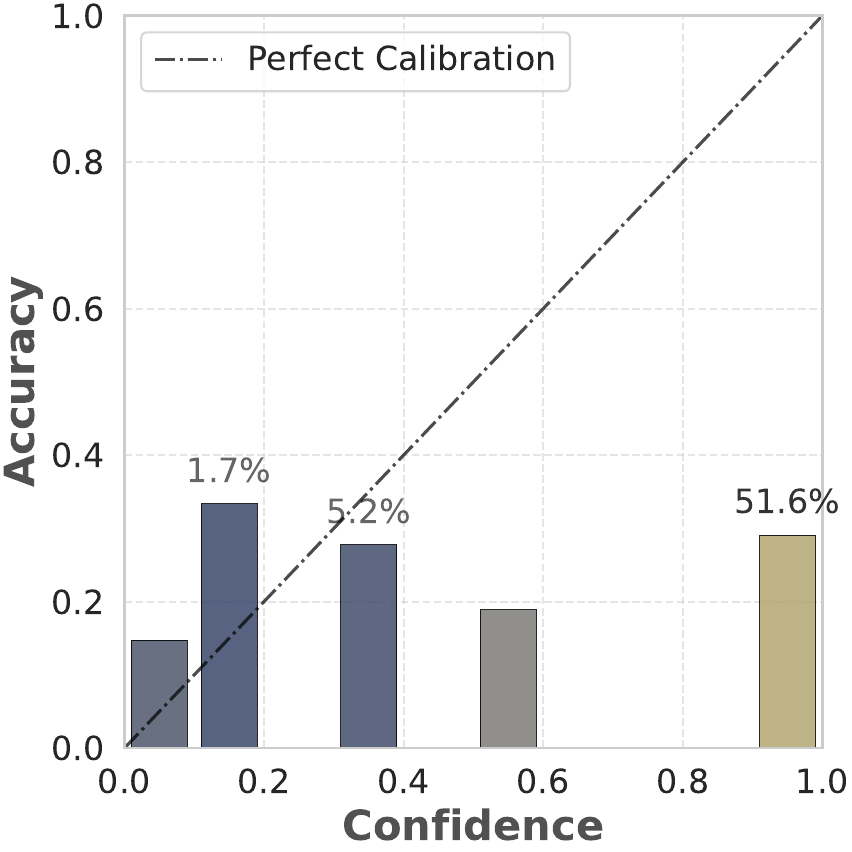}
        \label{fig:lmvlm_qwen2.5_triviaqa}
    }
    \hfill % \hspace{1cm}
    \subfloat[HalluMeasure]{
        \includegraphics[width=0.2\textwidth]{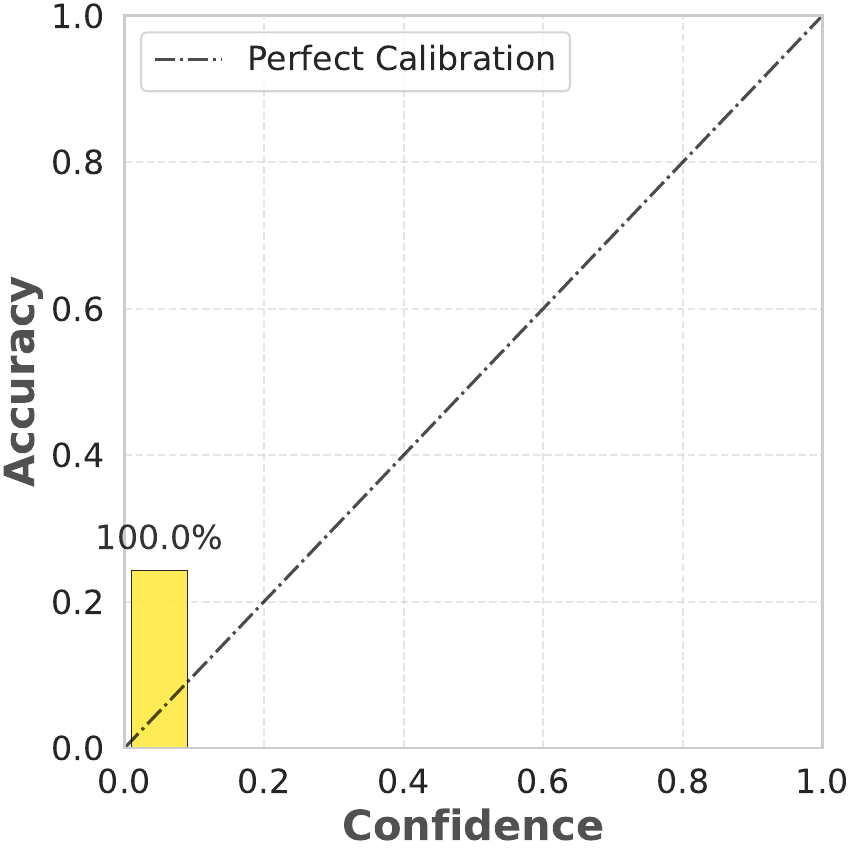}
        \label{fig:hallumeasure_qwen2.5_triviaqa}
    }
    
    \caption{Reliability diagrams for calibration methods using Qwen2.5 7B on TriviaQA. The number of diagram bins is 10. The color and the percentage number within each bar indicate the proportion of total points contained in each bin. }
    % label should be at the bottom of the caption
    \label{fig:qwen2.5_trivia_qa_comparison}
\end{figure*}

\noindent\textbf{TruthfulQA.}
\begin{figure*}[!htbp]
    %\begin{center}
    %\framebox[4.0in]{$\;$}
    %\fbox{\rule[-.5cm]{0cm}{4cm} \rule[-.5cm]{4cm}{0cm}}
    %\end{center}
    \centering
    \subfloat[Few-shot]{
        \includegraphics[width=0.2\textwidth]{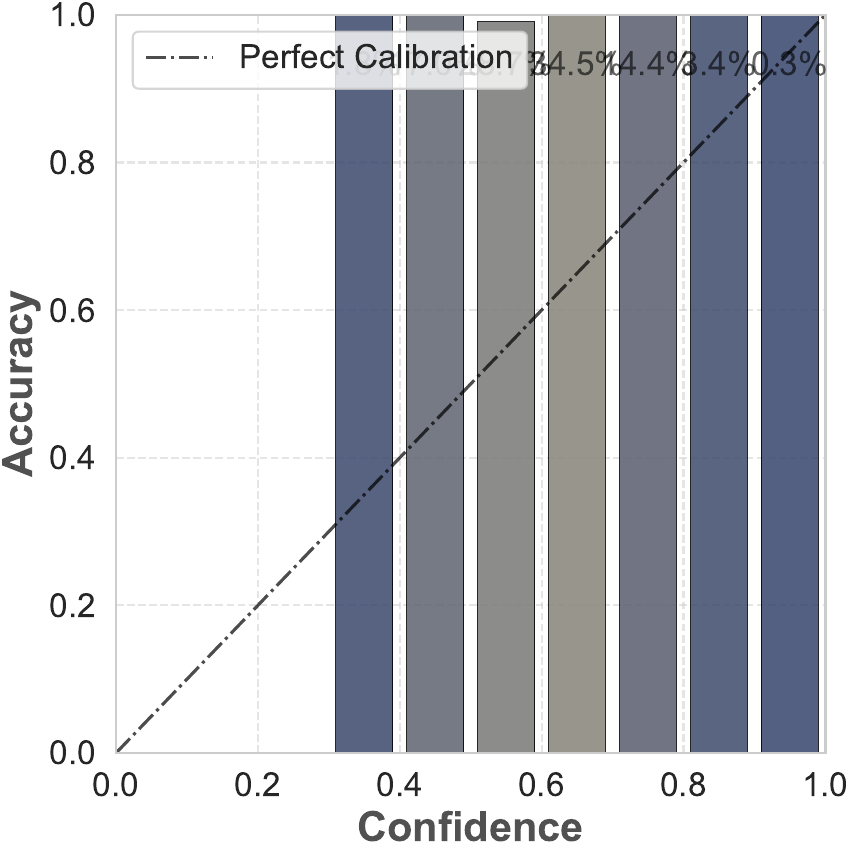}
        \label{fig:seq_likelihood_llama_truthful_qa}}
    \hfill
    \subfloat[Few-Shot CoT]{
        \includegraphics[width=0.2\textwidth]{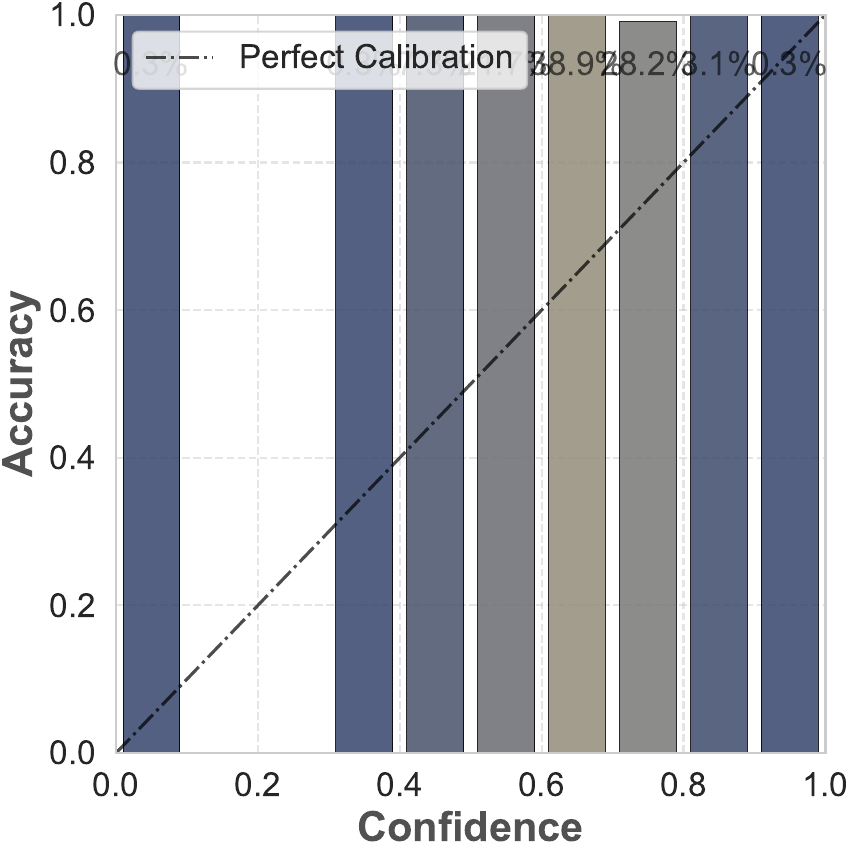}%
        \label{fig:cot_seq_likelihood_llama_truthful_qa}
    }
    \hfill
    \subfloat[Verbalized CoT]{
        \includegraphics[width=0.2\textwidth]{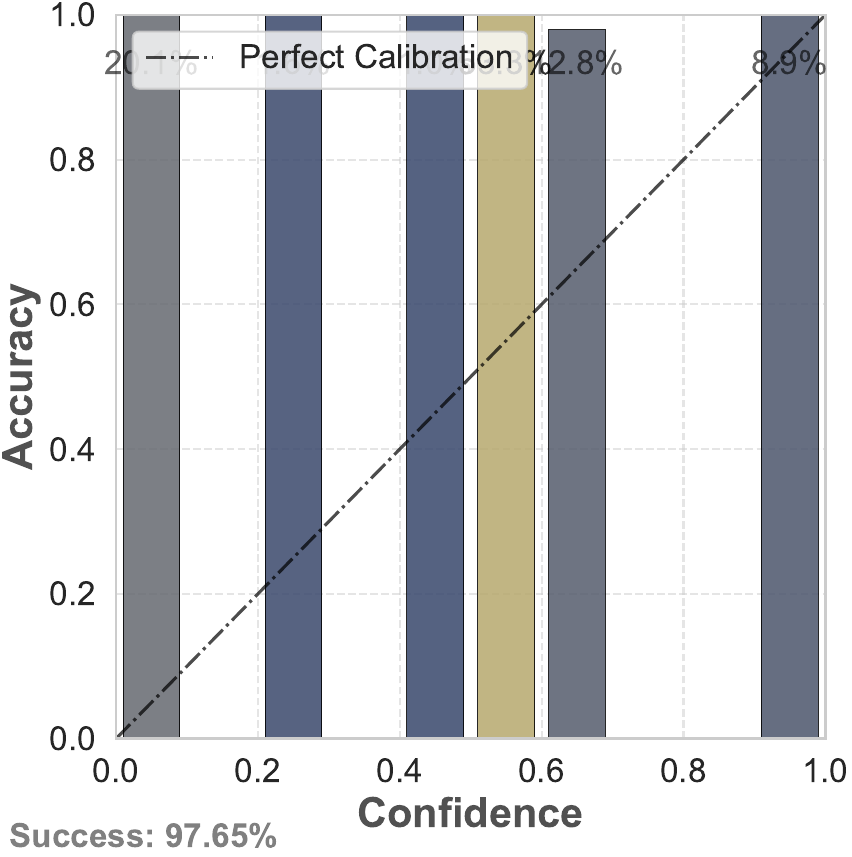}%
        \label{fig:verbalized_cot_qual_llama_truthful_qa}
    }
    \hfill
    \subfloat[Platt Scaling]{
        \includegraphics[width=0.2\textwidth]{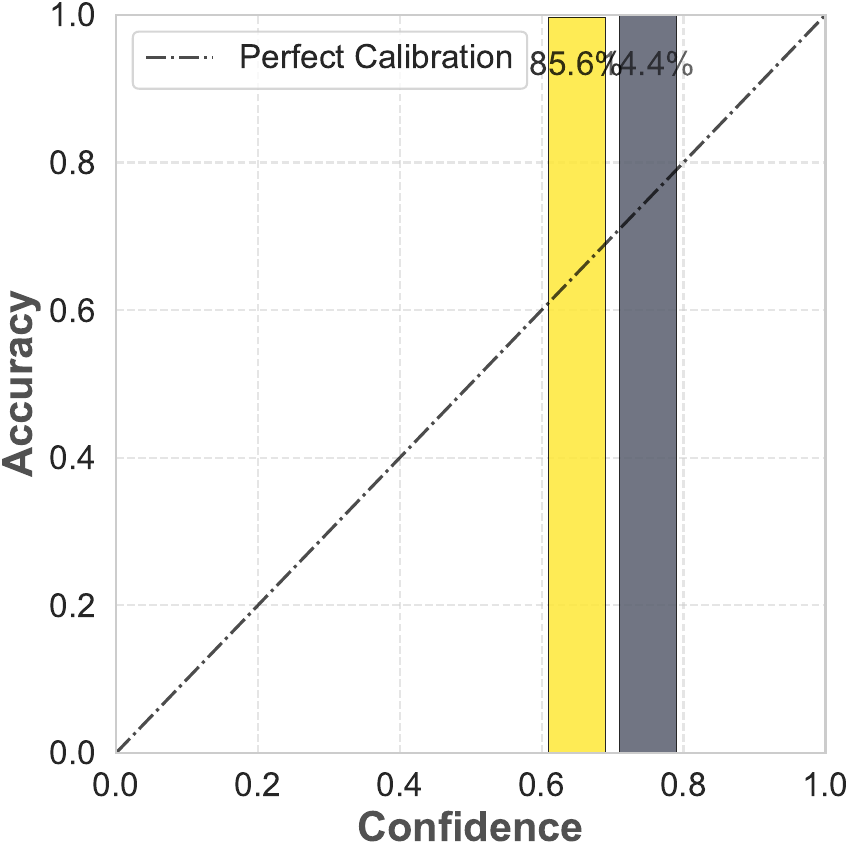}
        \label{fig:ps_seq_likelihood_llama_truthful_qa}
    }
    \hfill
    \subfloat[Temp. Scaling]{
        \includegraphics[width=0.2\textwidth]{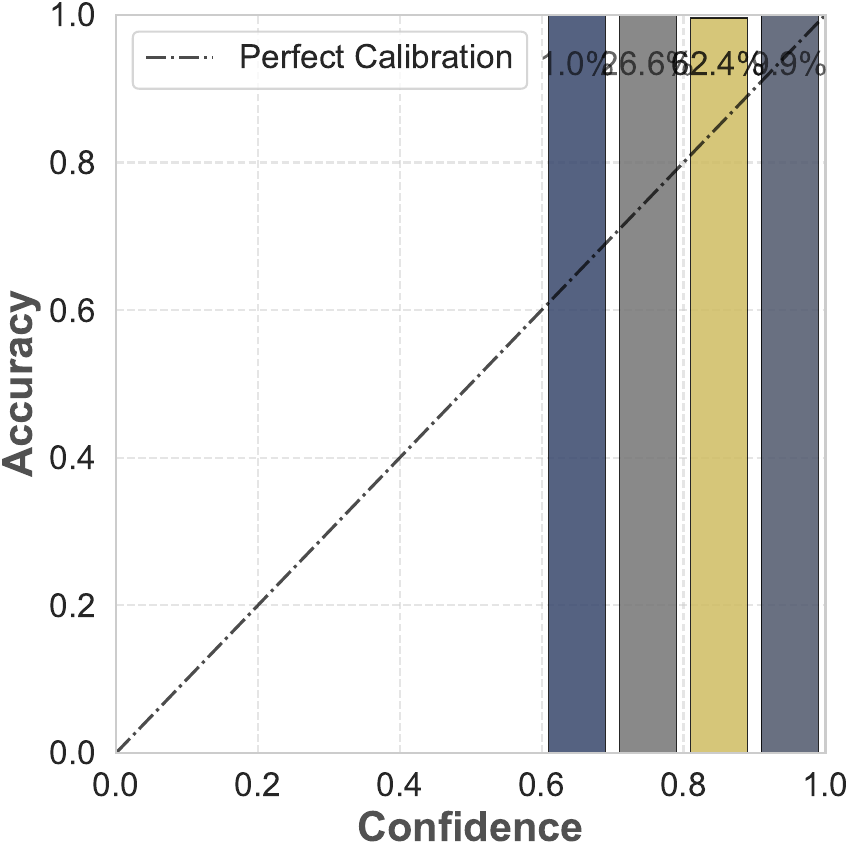}
        \label{fig:ts_seq_likelihood_llama_truthful_qa}
    }
    \hfill
    \subfloat[LMvsLM]{
        \includegraphics[width=0.2\textwidth]{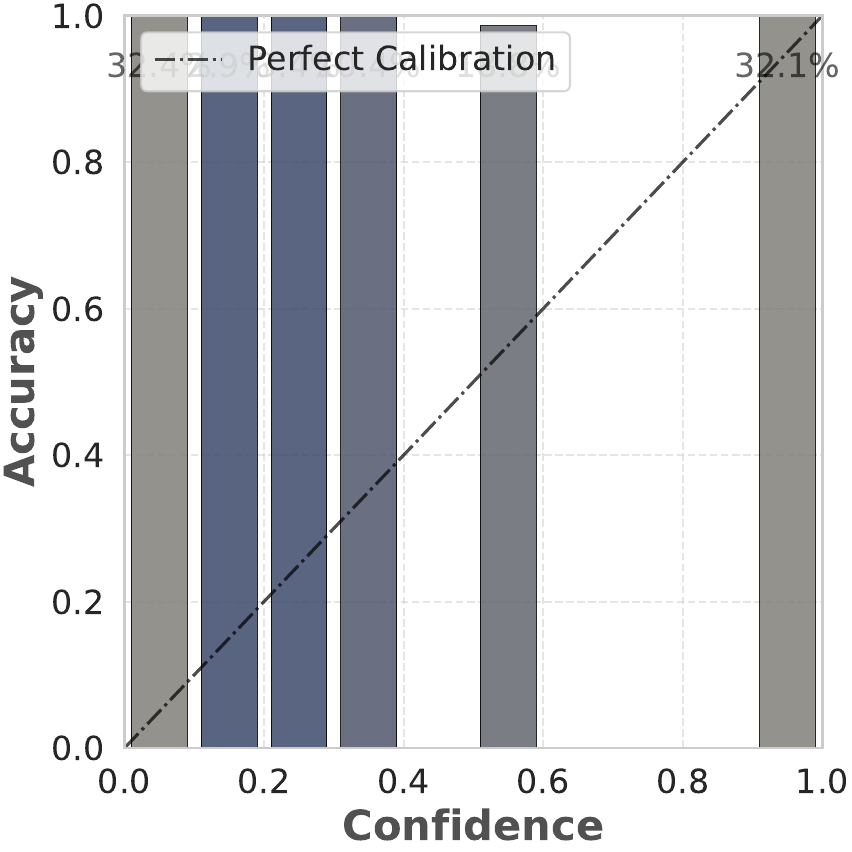}
        \label{fig:lmvslm_llama_truthful_qa}
    }
    \hfill
    \subfloat[HalluMeasure]{
        \includegraphics[width=0.2\textwidth]{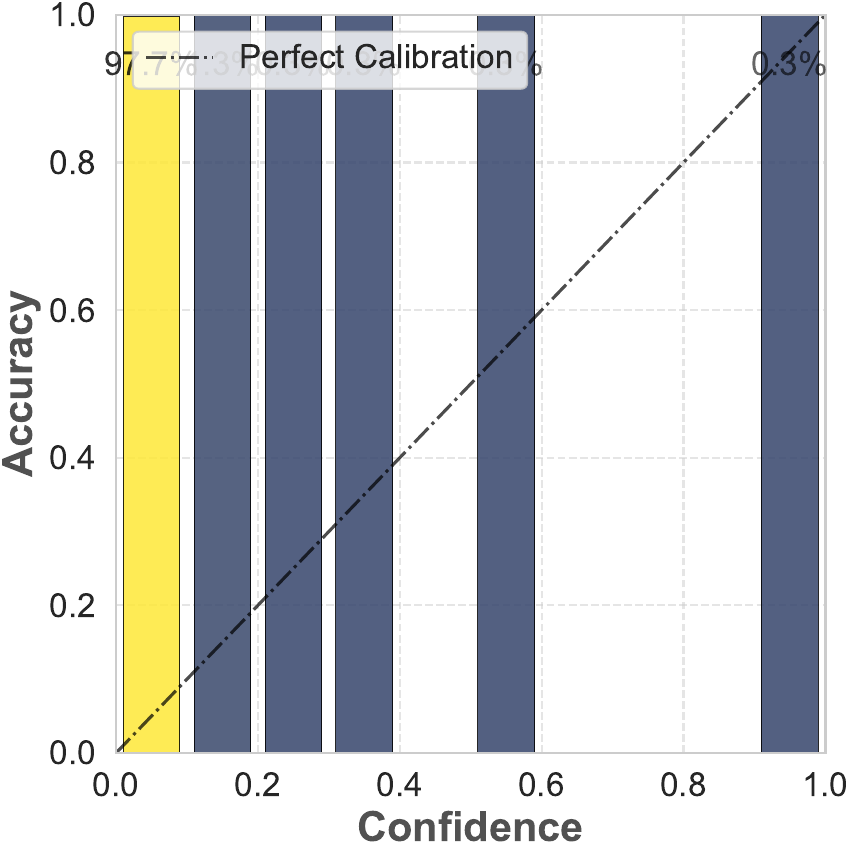}
        \label{fig:hallumeasure_llama_truthful_qa}
    }
    \caption{Reliability diagrams for different calibration methods with 10 bins using Llama 3.1 8B on TruthfulQA. The number of diagram bins is 10. The color and the percentage number on each bar indicate the proportion of total points contained in each bin. }
    % label should be at the bottom of the caption
    \label{fig:llama_truthful_qa_comparison}
\end{figure*}

\begin{figure*}[!htbp]
    %\begin{center}
    %\framebox[4.0in]{$\;$}
    %\fbox{\rule[-.5cm]{0cm}{4cm} \rule[-.5cm]{4cm}{0cm}}
    %\end{center}
    \centering
    \subfloat[Few-shot]{
        \includegraphics[width=0.2\textwidth]{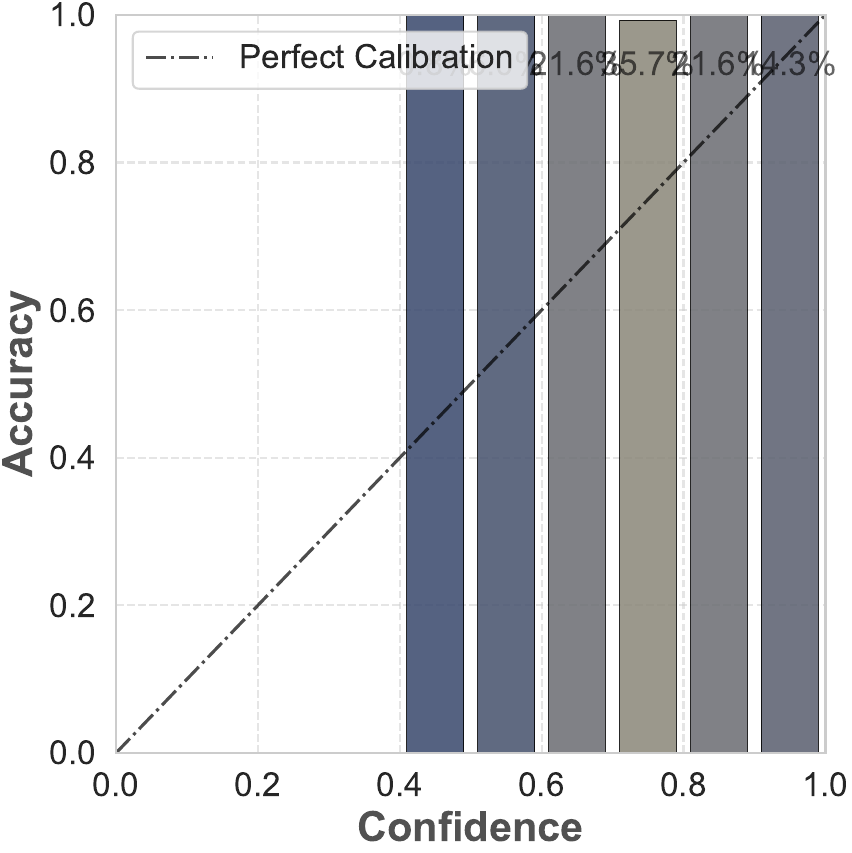}
        \label{fig:seq_likelihood_mistral_truthful_qa}}
    \hfill
    \subfloat[Few-Shot CoT]{
        \includegraphics[width=0.2\textwidth]{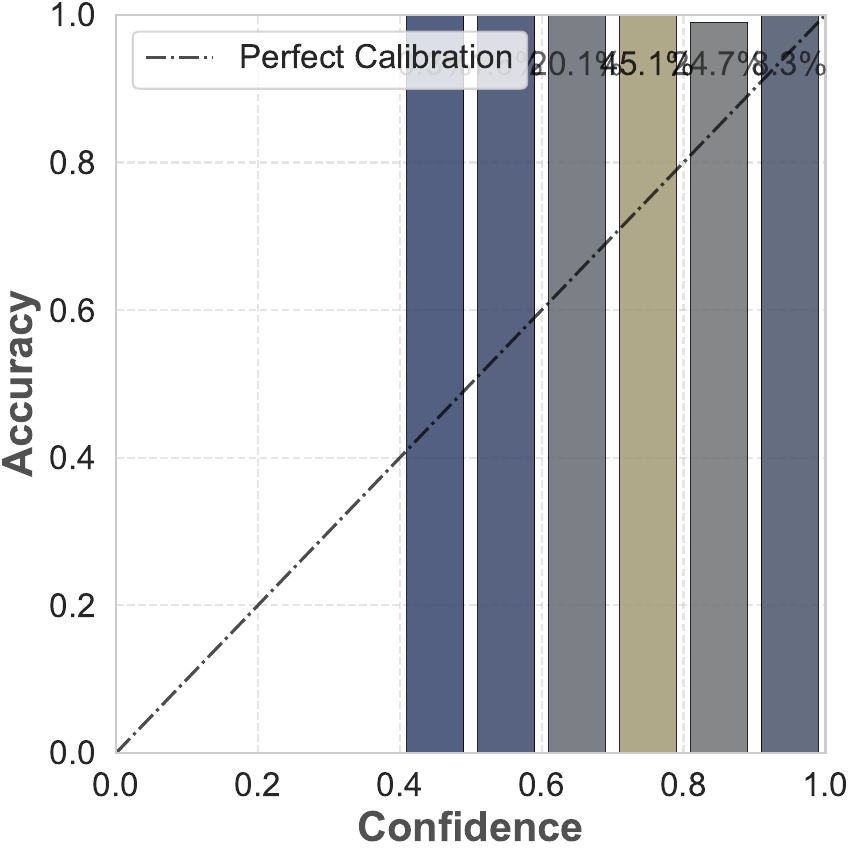}%
        \label{fig:cot_seq_likelihood_mistral_truthful_qa}
    }
    \hfill
    \subfloat[Verbalized CoT]{
        \includegraphics[width=0.2\textwidth]{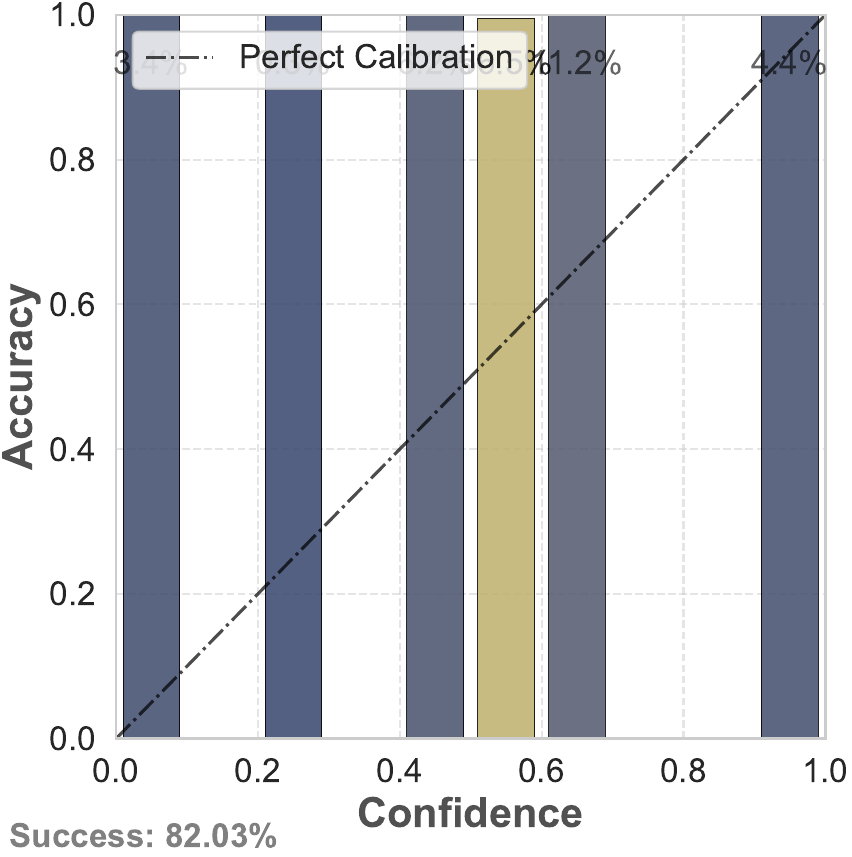}%
        \label{fig:verbalized_cot_qual_mistral_truthful_qa}
    }
    \hfill
    \subfloat[Platt Scaling]{
        \includegraphics[width=0.2\textwidth]{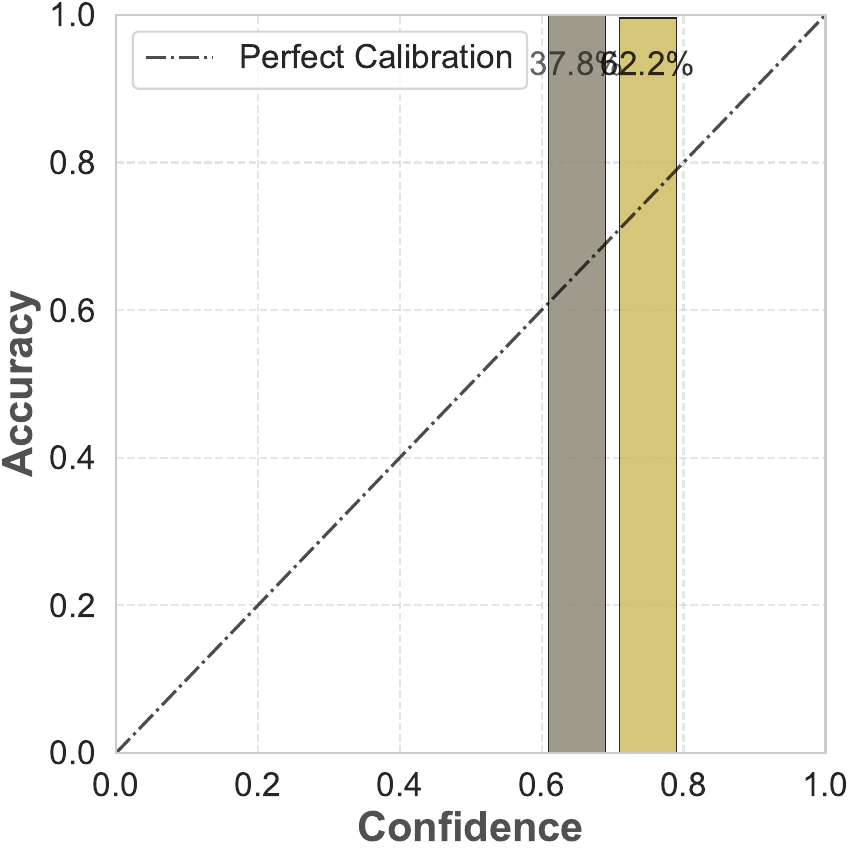}
        \label{fig:ps_seq_likelihood_mistral_truthful_qa}
    }
    \hfill
    \subfloat[Temp. Scaling]{
        \includegraphics[width=0.2\textwidth]{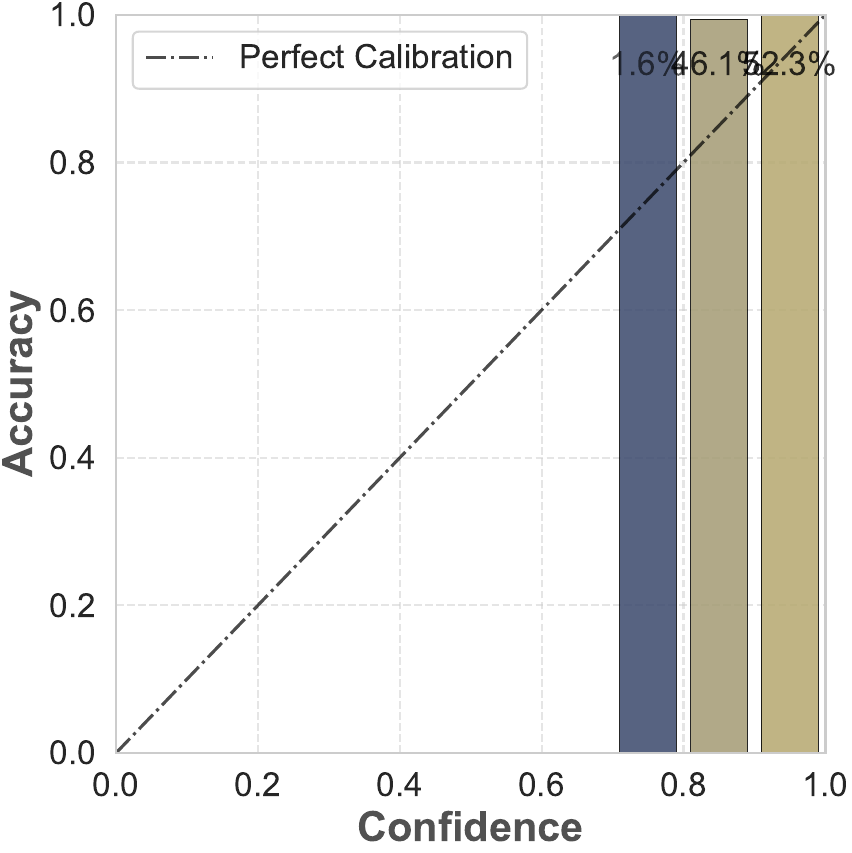}
        \label{fig:ts_seq_likelihood_mistral_truthful_qa}
    }
    \hfill
    \subfloat[LMvsLM]{
        \includegraphics[width=0.2\textwidth]{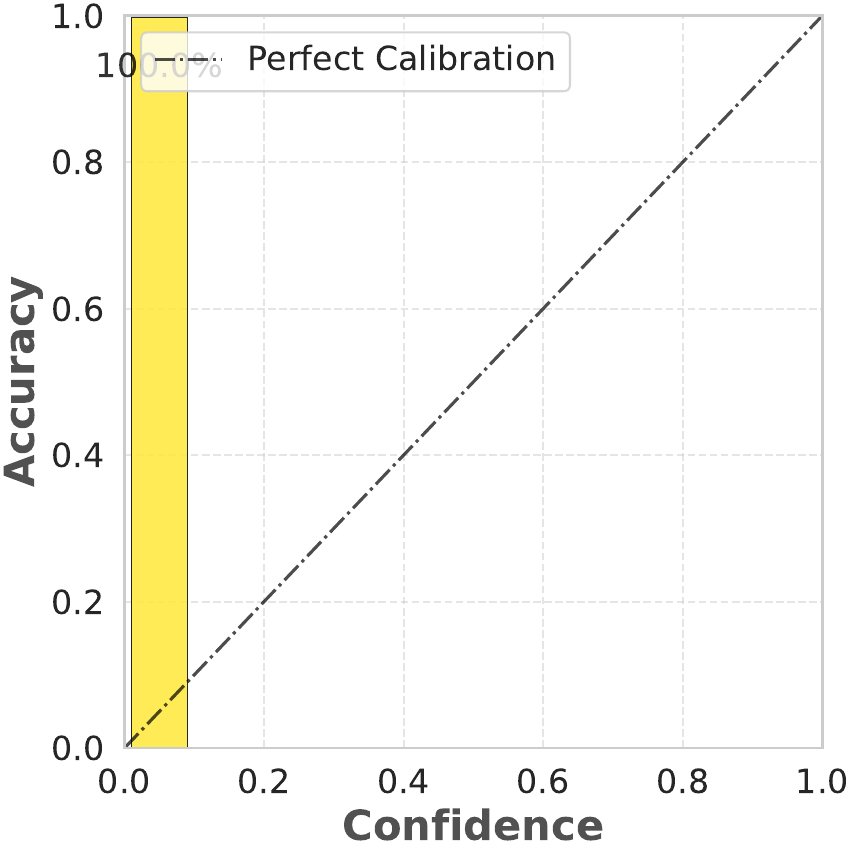}
        \label{fig:lmvslm_mistral_truthful_qa}
    }
    \hfill
    \subfloat[HalluMeasure]{
        \includegraphics[width=0.2\textwidth]{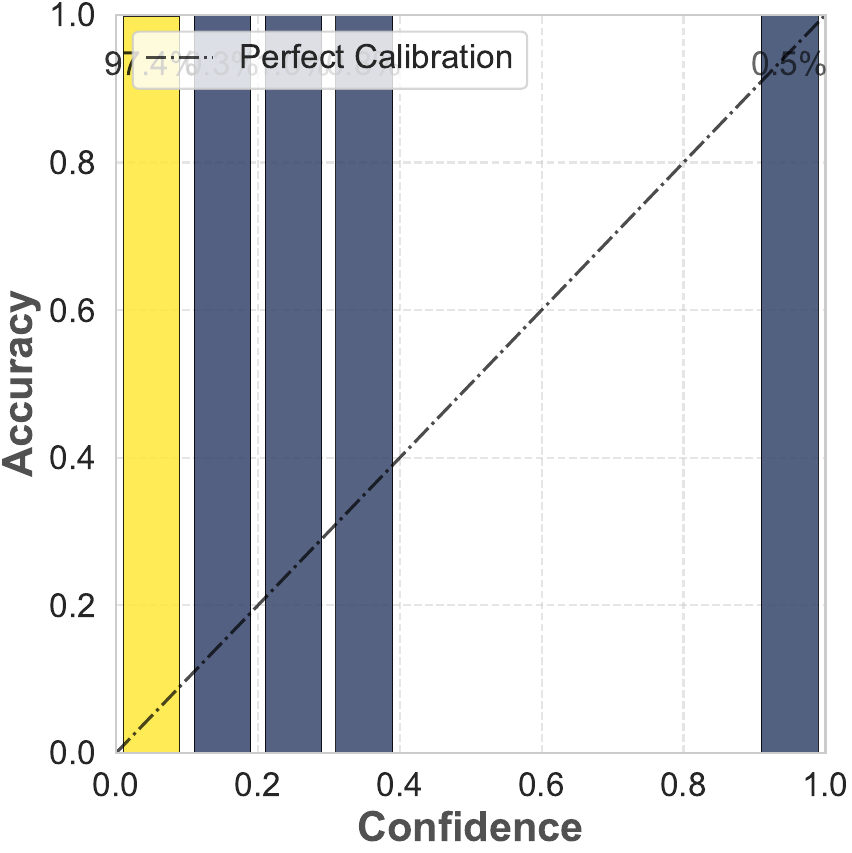}
        \label{fig:hallumeasure_mistral_truthful_qa}
    }
    \caption{Reliability diagrams for different calibration methods with 10 bins using Mistral v0.3 7B on TruthfulQA. The number of diagram bins is 10. The color and the percentage number on each bar indicate the proportion of total points contained in each bin. }
    % label should be at the bottom of the caption
    \label{fig:mistral_truthful_qa_comparison}
\end{figure*}

\begin{figure*}[!htbp]
    %\begin{center}
    %\framebox[4.0in]{$\;$}
    %\fbox{\rule[-.5cm]{0cm}{4cm} \rule[-.5cm]{4cm}{0cm}}
    %\end{center}
    \centering
    \subfloat[Few-shot]{
        \includegraphics[width=0.2\textwidth]{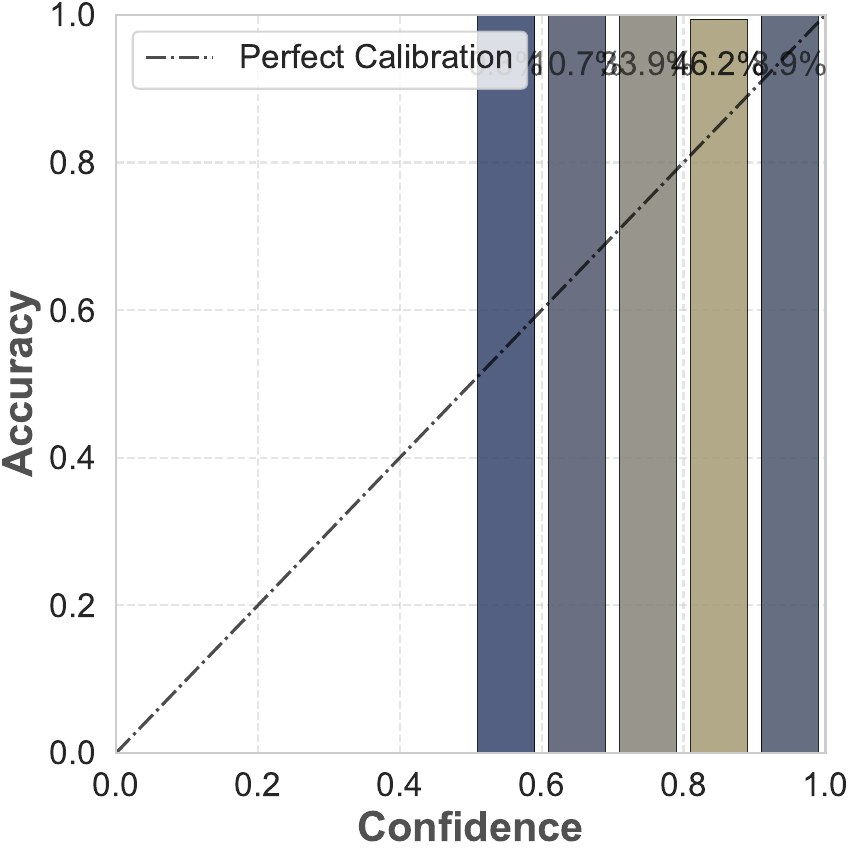}
        \label{fig:seq_likelihood_qwen2.5_truthful_qa}
    }
    \hfill
    \subfloat[Few-Shot CoT]{
        \includegraphics[width=0.2\textwidth]{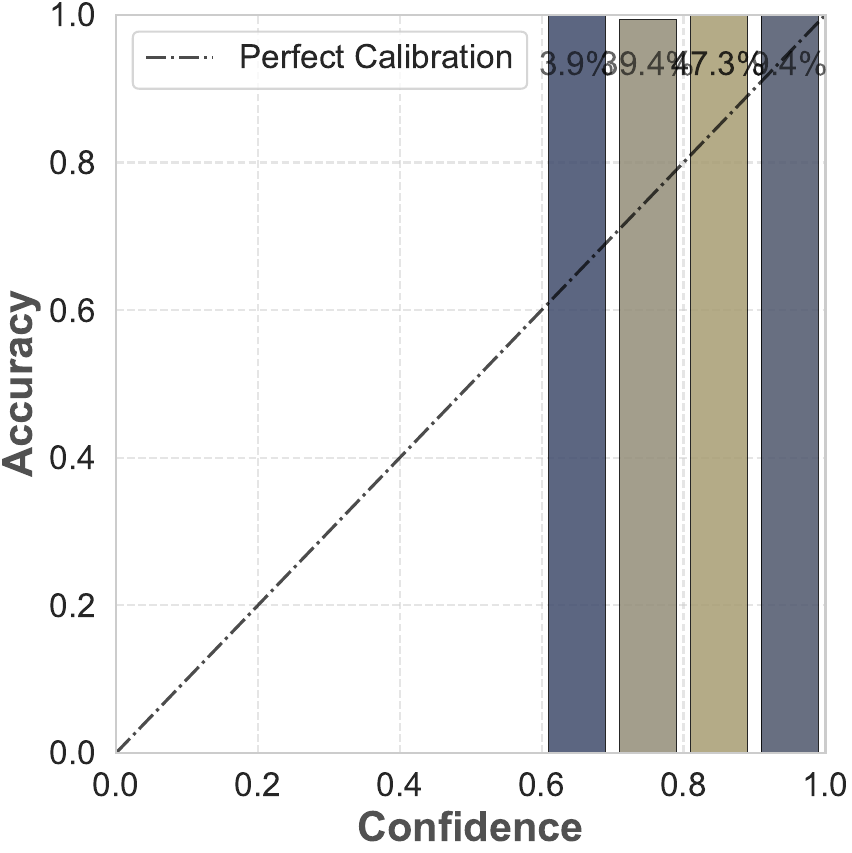}
        \label{fig:cot_seq_likelihood_qwen2.5_truthful_qa}
    }
    \hfill
    \subfloat[Verbalized CoT]{
        \includegraphics[width=0.2\textwidth]{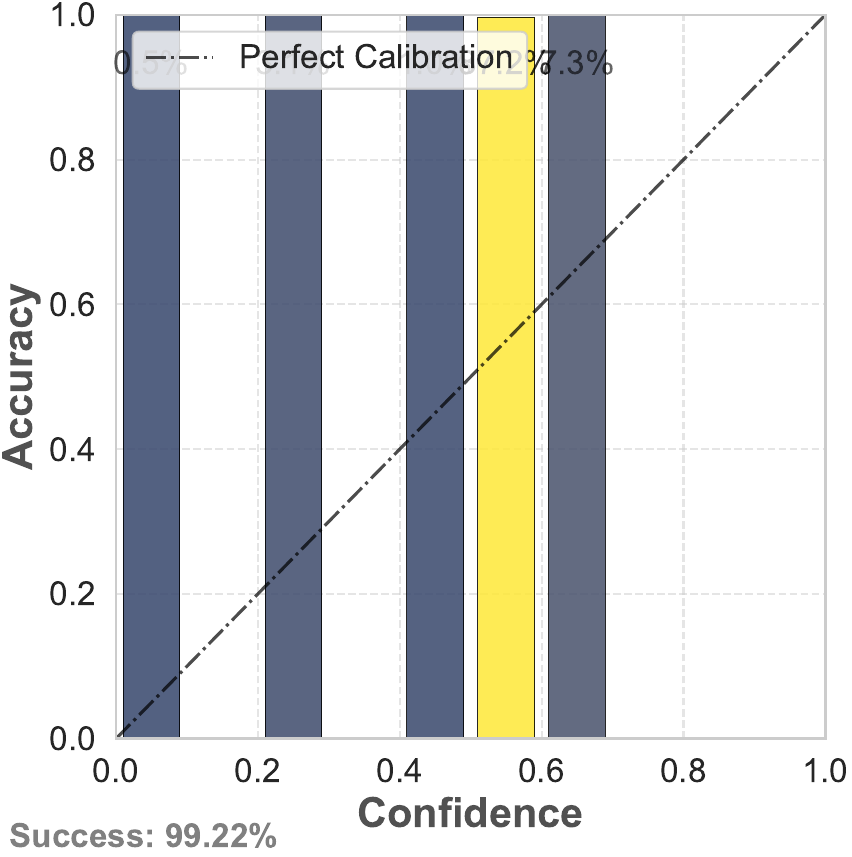}
        \label{fig:verbalized_cot_qual_qwen2.5_truthful_qa}
    }
    \hfill
    \subfloat[Platt Scaling]{
        \includegraphics[width=0.2\textwidth]{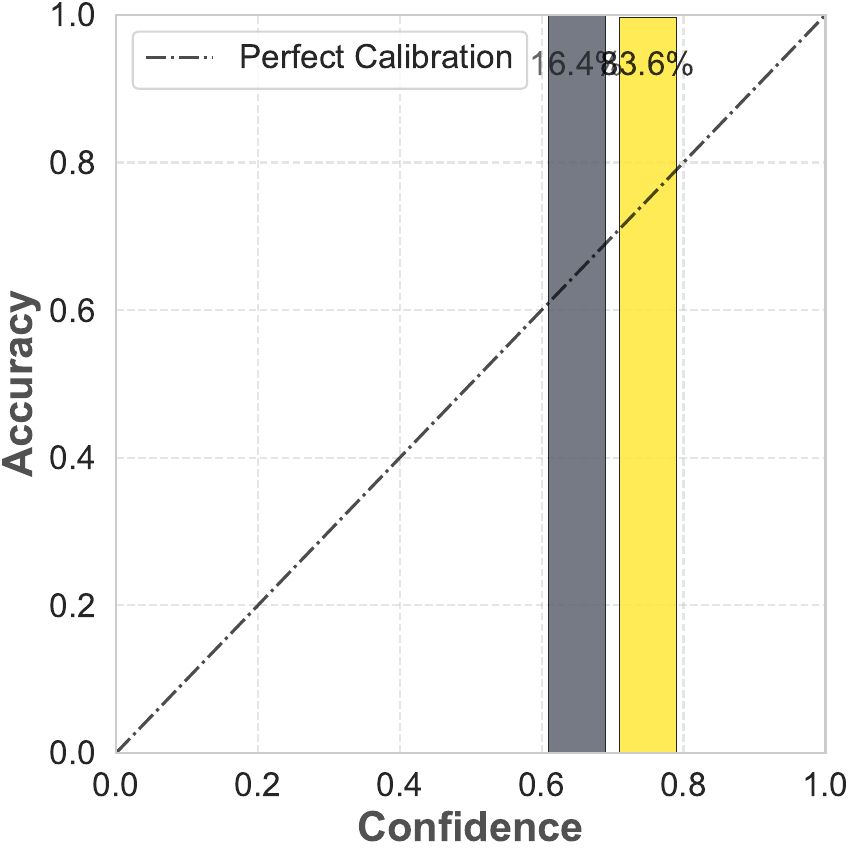}
        \label{fig:ps_seq_likelihood_qwen2.5_truthful_qa}
    }
    \hfill
    \subfloat[Temp. Scaling]{
        \includegraphics[width=0.2\textwidth]{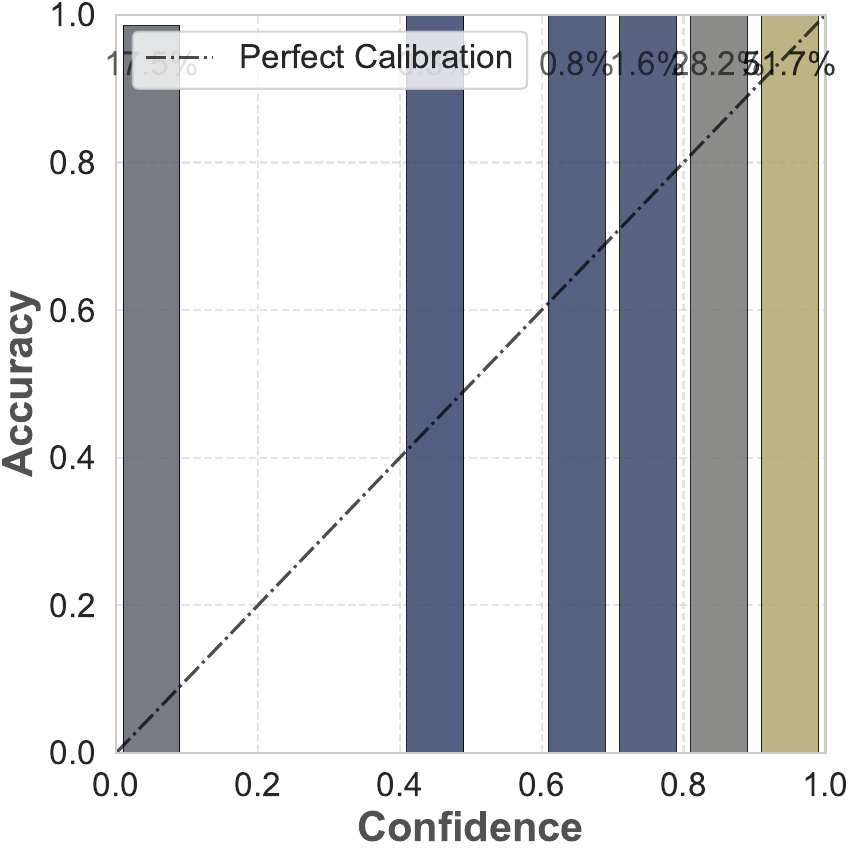}
        \label{fig:ts_seq_likelihood_qwen2.5_truthful_qa}
    }
    \hfill
    \subfloat[LMvsLM]{
        \includegraphics[width=0.2\textwidth]{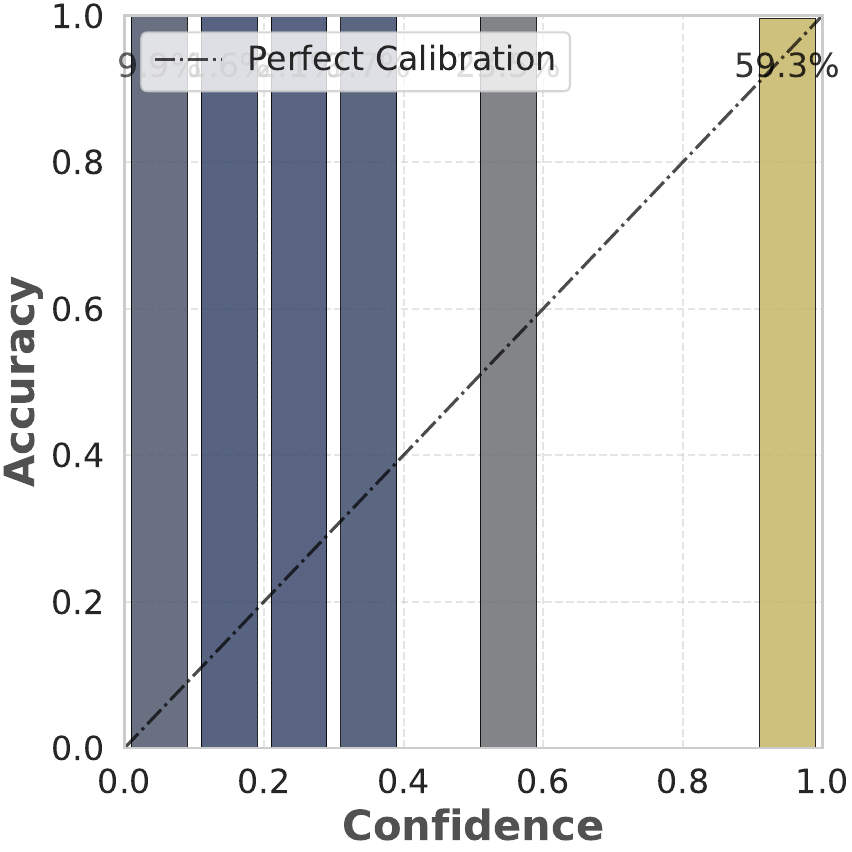}
        \label{fig:lmvslm_qwen2.5_truthful_qa}
    }
    \hfill
    \subfloat[HalluMeasure]{
        \includegraphics[width=0.2\textwidth]{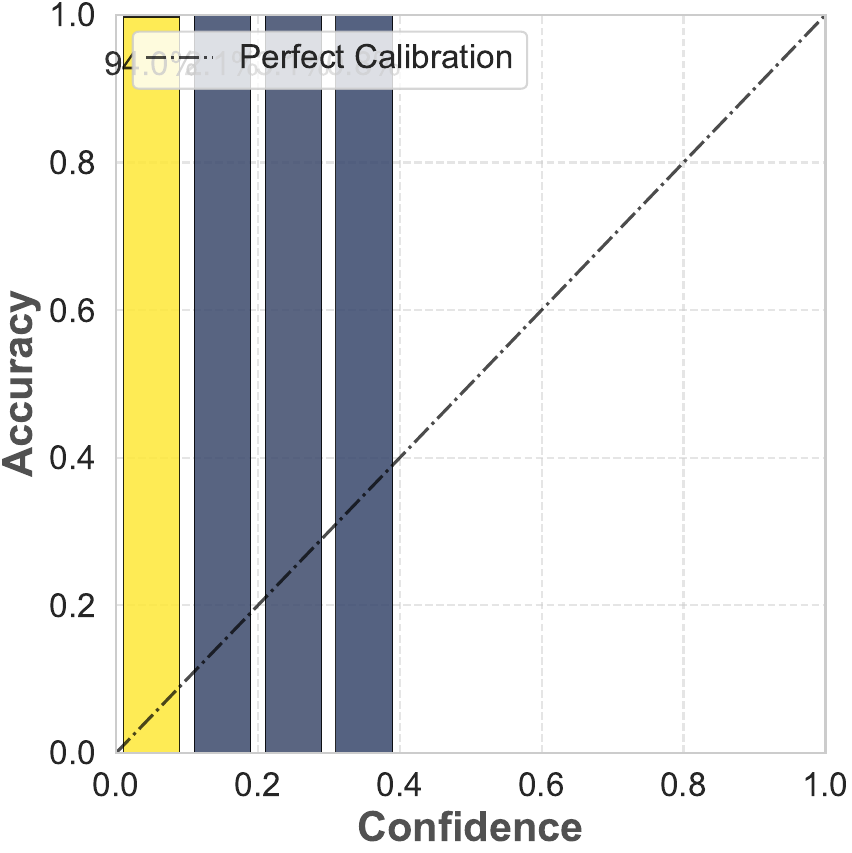}
        \label{fig:hallumeasure_qwen2.5_truthful_qa}
    }
    \caption{Reliability diagrams for calibration methods using Qwen2.5 7B on TruthfulQA. The number of diagram bins is 10. The color and the percentage number within each bar indicate the proportion of total points contained in each bin.}
    % label should be at the bottom of the caption
    \label{fig:qwen2.5_truthful_qa_comparison}
\end{figure*}

\begin{figure*}[!htbp]
    %\begin{center}
    %\framebox[4.0in]{$\;$}
    %\fbox{\rule[-.5cm]{0cm}{4cm} \rule[-.5cm]{4cm}{0cm}}
    %\end{center}
    \centering
    \subfloat[Few-shot]{
        \includegraphics[width=0.2\textwidth]{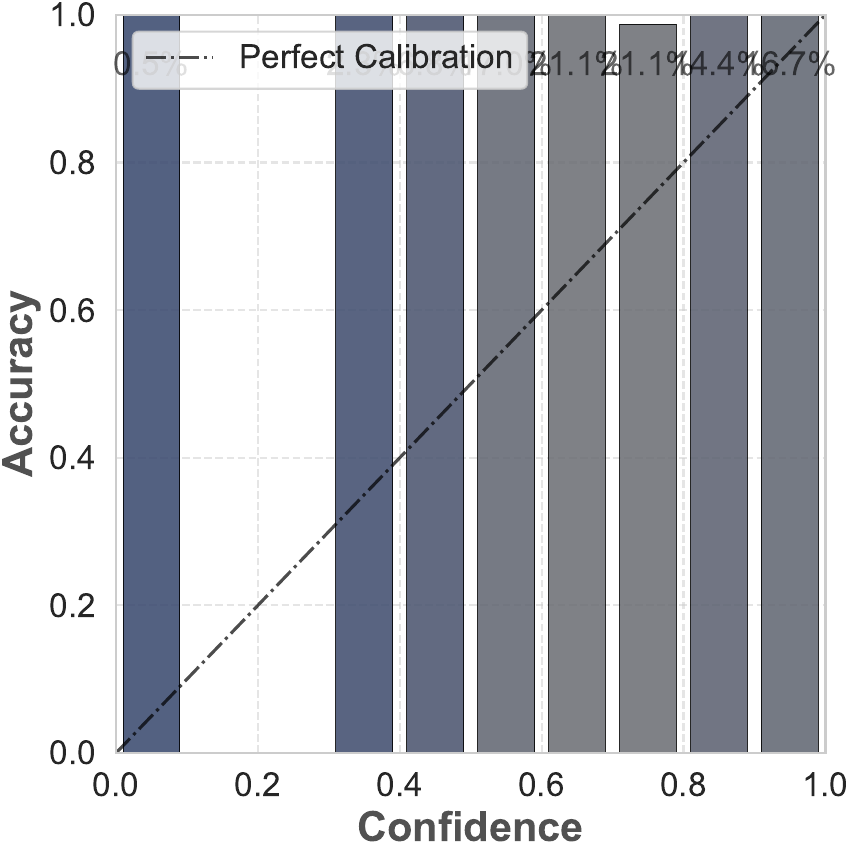}
        \label{fig:seq_likelihood_gpt4o_truthful_qa}
    }
    \hfill
    \subfloat[Few-Shot CoT]{
        \includegraphics[width=0.2\textwidth]{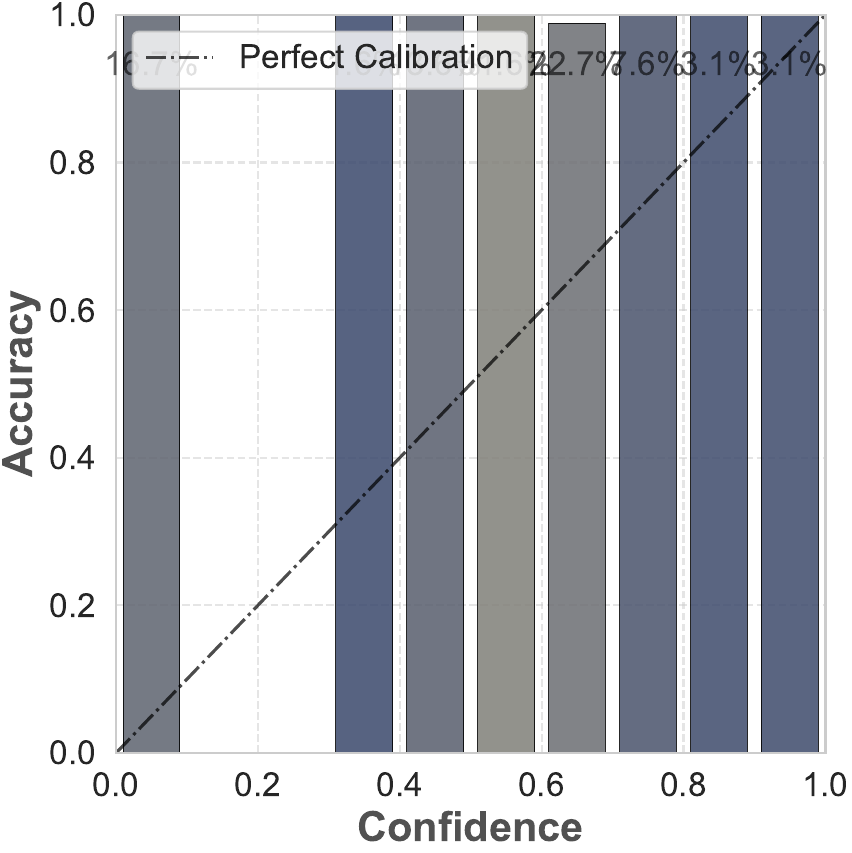}
        \label{fig:cot_seq_likelihood_gpt4o_truthful_qa}
    }
    \hfill
    \subfloat[Verbal. CoT]{
        \includegraphics[width=0.2\textwidth]{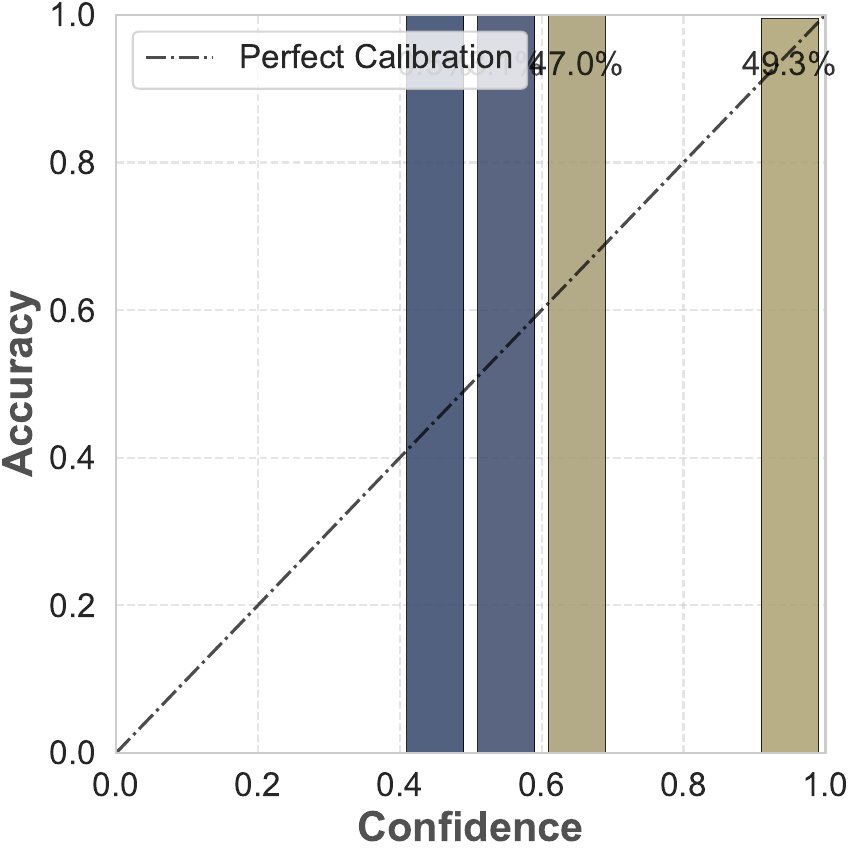}
        \label{fig:verbalized_cot_qual_gpt4o_truthful_qa}
    }
    \hfill
    \subfloat[Platt Scaling]{
        \includegraphics[width=0.2\textwidth]{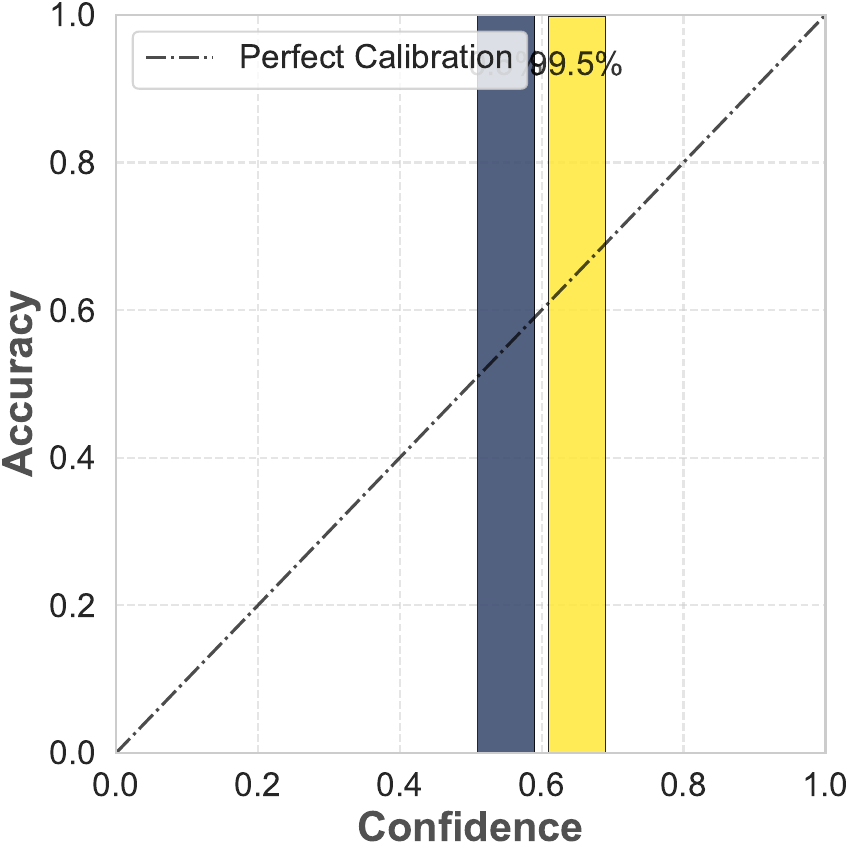}
        \label{fig:ps_seq_likelihood_gpt4o_truthful_qa}
    }
    \hfill
    \subfloat[Temp. Scaling]{
        \includegraphics[width=0.2\textwidth]{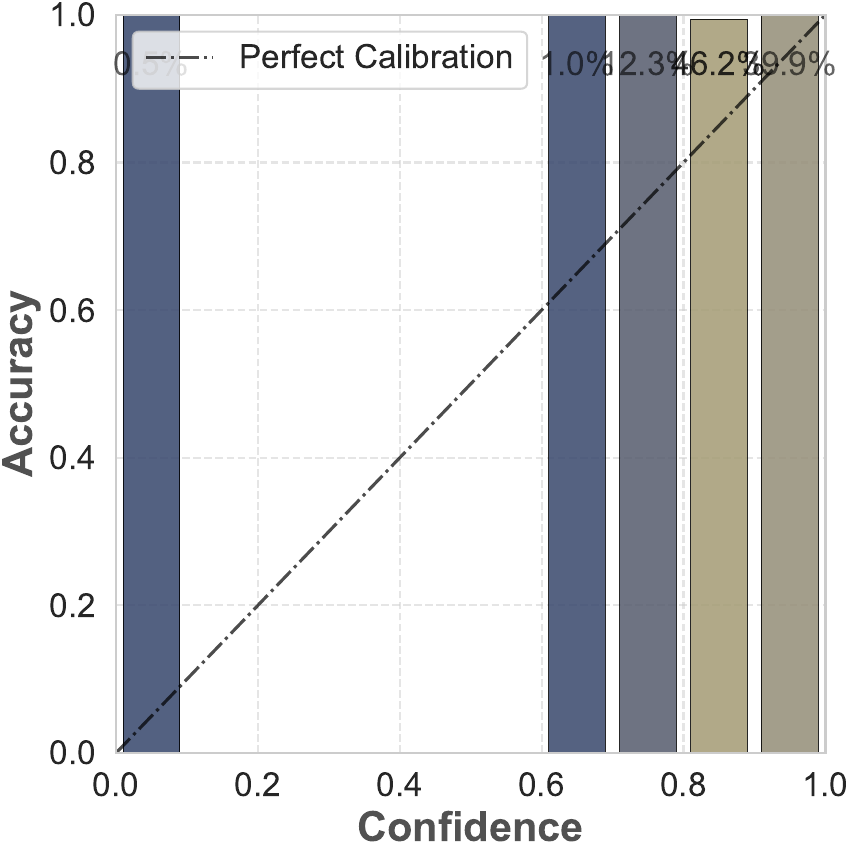}
        \label{fig:ts_seq_likelihood_gpt4o_truthful_qa}
    }
    \hfill % \hspace{1cm}
    \subfloat[LMvsLM]{
        \includegraphics[width=0.2\textwidth]{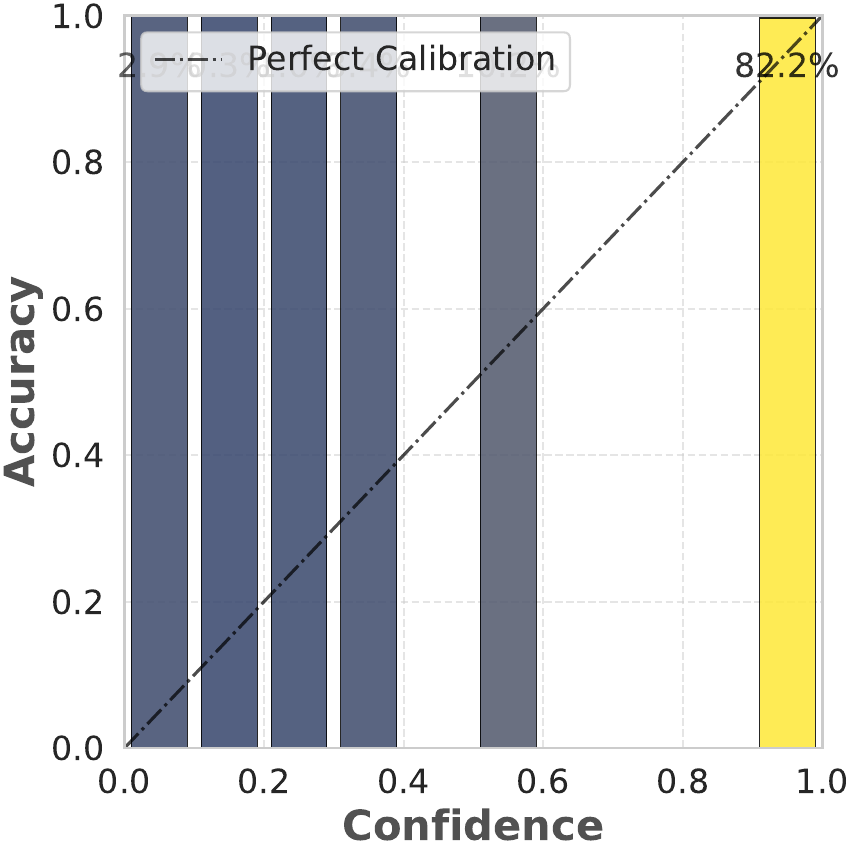}
        \label{fig:lmvslm_gpt-4o_truthfulqa}
    }
    \hfill % \hspace{1cm}
    \subfloat[HalluMeasure]{
        \includegraphics[width=0.2\textwidth]{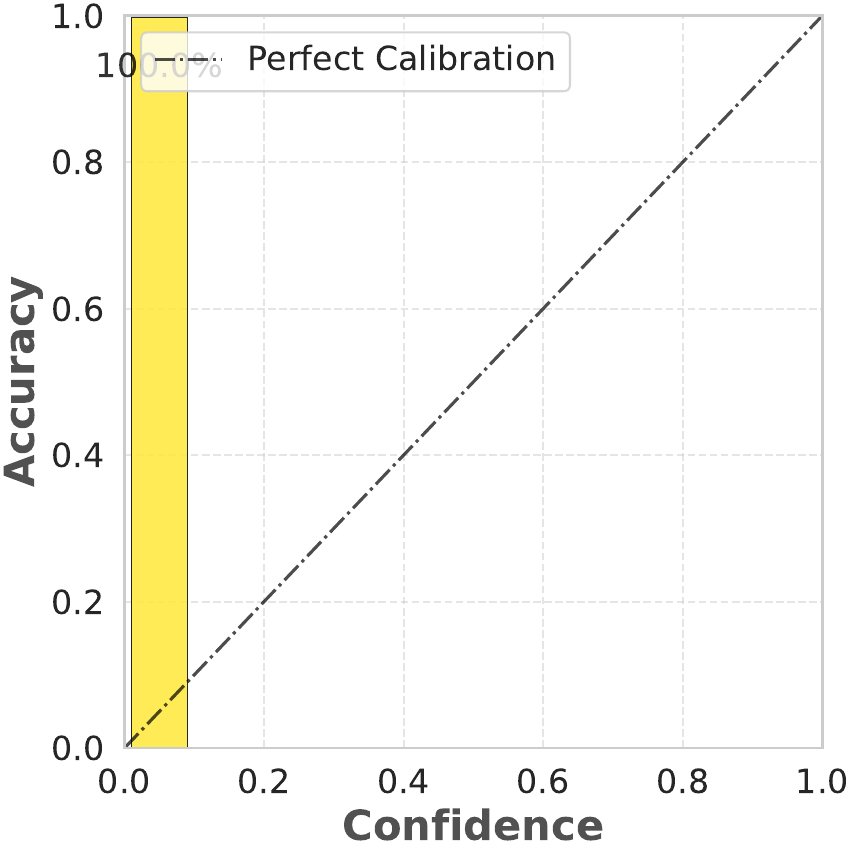}
        \label{fig:hallumeasure_gpt-4o_truthfulqa}
    }
    \hfill % \hspace{1cm}
    \subfloat[Ours]{
        \includegraphics[width=0.2\textwidth]{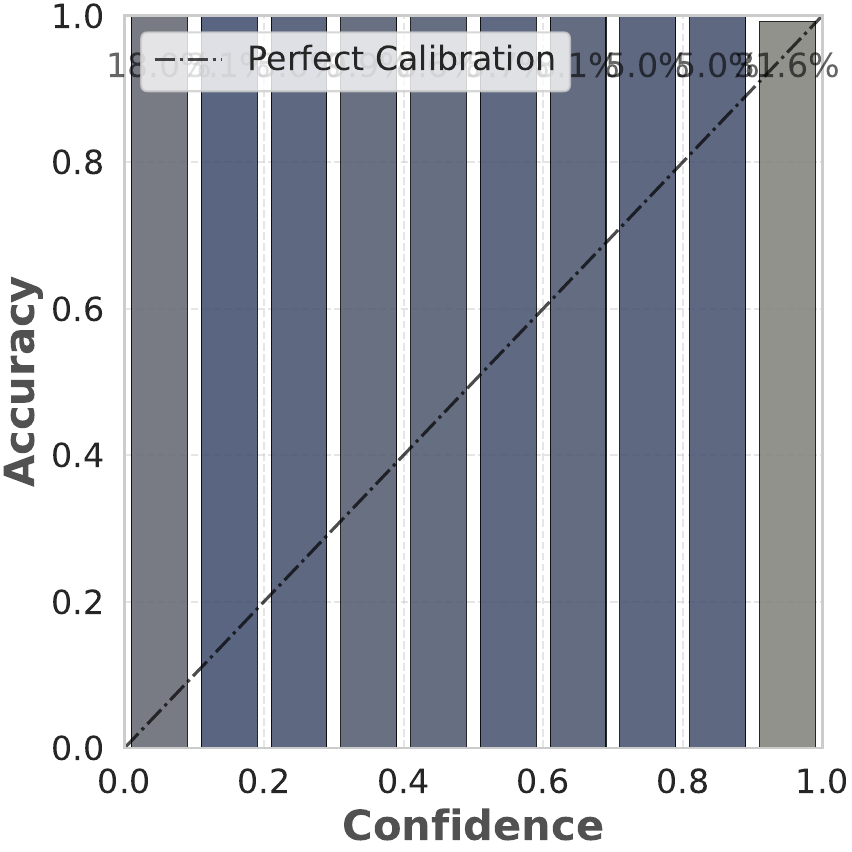}
        \label{fig:ourmethod_gpt-4o_truthfulqa}
    }
    \caption{Reliability diagrams for calibration methods using GPT-4o on TruthfulQA. The number of diagram bins is 10. The color as well as the percentage number within each bar indicate the proportion of total points contained in each bin. }
    % label should be at the bottom of the caption
    \label{fig:gpt4o_truthful_qa_comparison}
\end{figure*}

\section{Discussion: When the Method Helps Decision Making}
\label{appendix:discussion}

\noindent\textbf{Best-case scenarios.} (i)~\emph{Mixed correctness}: when
responses contain both correct and incorrect claims, claim-level scores
support targeted intervention. (ii)~\emph{Closed-box deployment}:
verbalized confidence at the claim level avoids the logit-access
constraint of TS/PS-on-likelihoods and Fadeeva et al.\
\cite{fadeeva2024factcheckingoutpllm}. (iii)~\emph{Moderate complexity}:
multi-claim responses where decomposition adds value beyond a single
score.

\noindent\textbf{Failure modes.} (i)~\emph{Consensus failure}: all
sampled responses agree on an incorrect claim (TruthfulQA on GPT-4o).
(ii)~\emph{Verifier weakness}: errors in claim extraction or verification
propagate. (iii)~\emph{Unverifiable claims}: future predictions or
creative content where no external check exists.

\noindent\textbf{Practical recommendations.}
\textbf{Cost-benefit:} apply claim-level decomposition selectively when
fine-grained confidence matters; for simple queries response-level may
suffice. \textbf{Verifier selection:} verifier quality dominates; consider
hybrid LLM+retrieval verifiers. \textbf{Selective application:} use fast
response-level screening to flag suspicious responses, then claim-level
analysis only for those. \textbf{Threshold calibration:} safety-critical
domains require conservative thresholds; exploratory uses can tolerate
lower confidence.

\noindent\textbf{Future directions.} Adaptive sampling (early-stop $N$
when samples agree); hierarchical confidence over interdependent claims;
domain-specific calibration (medical/legal/scientific); uncertainty-aware
training that incorporates claim-level signals into objectives.

\end{document}